\documentclass[journal]{IEEEtran}

\ifCLASSINFOpdf
\else
\fi

\usepackage{times}
\usepackage{epsfig}
\usepackage{graphicx}
\usepackage{amsmath}
\usepackage{amssymb}
\usepackage{mathrsfs}
\usepackage{multirow}
\usepackage{blkarray}
\usepackage{float}
\usepackage{array}
\usepackage{epstopdf}
\usepackage{subcaption}
\usepackage[T1]{fontenc}
\usepackage{algorithm}
\usepackage{algpseudocode}
\usepackage{color,soul}

\begin{document}

\title{An ADMM Approach to Masked Signal Decomposition Using Subspace Representation}

\author{Shervin~Minaee,~\IEEEmembership{Student Member,~IEEE,}
        and~Yao~Wang,~\IEEEmembership{Fellow,~IEEE}
        \\ New York University, Electrical Engineering Department
}

\maketitle

\begin{abstract}
Signal decomposition is a classical problem in signal processing, which aims to separate an observed signal into two or more components each with its own property. 
Usually each component is described by its own subspace or dictionary.
Extensive research has been done for the case where the components are  additive, but in real world applications, the components are often non-additive. For example, an image may consist of a foreground object overlaid on a background, where each pixel either belongs to the foreground or the background. 
In such a situation, to separate signal components, we need to find a binary mask which shows the location of each component.
Therefore it requires to solve a binary optimization problem.
Since most of the binary optimization problems are intractable, we relax this problem to the approximated continuous problem, and solve it by alternating optimization technique.
We show the application of the proposed algorithm for three applications:  separation of text  from background in images, separation of moving objects from a background undergoing global camera motion in videos, separation of sinusoidal and spike components in one dimensional signals. We demonstrate in each case that considering the non-additive nature of the problem can lead to significant improvement.
\end{abstract}


\IEEEpeerreviewmaketitle

\section{Introduction}
Signal decomposition is an important problem in signal processing and has a wide range of applications.
Image segmentation and audio source separation are some of the applications of signal decomposition.
For example in image segmentation the goal is to decompose the image into several components such that each one represents the same content or semantic information.
Different algorithms have been proposed in the past, where each one of them looked at signal decomposition from a different perspective.
Perhaps, Fourier transform \cite{fourier} is one of the earliest work on signal decomposition where the goal is to decompose a signal into different frequencies.
Using the same methodology, many algorithms have been proposed for high and low frequency components.
Wavelet and multi-resolution decomposition are also another big group of methods which are designed for signal decomposition in both time and frequency domain \cite{wave1}-\cite{wave2}. 
In the more recent works, there have been many works on sparsity based signal decomposition. 
In \cite{bofili}, the authors proposed a sparse representation based method for blind source sparation.
In \cite{starck}, Starck et al proposed an image decomposition approach using both sparsity and variational approach.
The same approach has been used for morphological component analysis by the Elad et al \cite{elad}.
In the more recent works, there have been many works on low-rank decomposition, where in the simplest case the goal is to decompose a signal into two components, one being low rank, another being sparse.
Usually the nuclear and $\ell_1$ norms \cite{fazel} are used to promote low-rankness and sparsity respectively. 
To name some of the promising works along this direction,
in \cite{candes}, Candes et al proposed a low-rank decomposition for matrix completion.
In \cite{peng}, Peng et al proposed a sparse and low-rank decomposition approach with application for robust image alignment.
Recently, RPCA has been widely used for various applications  in image and video processing \cite{back_sub}.

Most of the prior approaches for signal decomposition consider additive model, i.e. the signal components are added in a mathematical sense to generate the overall signal. 
In the case of two components, this can be described by:
\begin{equation}
x= x_1+ x_2,
\end{equation}
Here $x$ denotes a vector in $R^N$. Assuming we have some prior knowledge about each component, we can form an optimization problem as in Eq. (2) to solve the signal decomposition problem. Here $\phi_k(\cdot)$ is the regularization term that encodes the prior knowledge about the corresponding signal component.
\begin{equation}
\begin{aligned}
& \hspace{0.1cm} \underset{x_k}{\text{min}}
 \   \sum_{k=1}^2 \phi_k(x_k), \ \ \  \ \text{s.t.} \ \ \sum_{k=1}^2 x_k= x.
\end{aligned}
\end{equation}
In this work, we investigate a different class of signal decomposition, where the signal components are overlaid on top of each other, rather than simply added. 
In other word, at each signal element only one of the signal components contributes to the observed signal x.
We can formulate this as:
\begin{equation}
x= \sum_{k=1}^2 w_k \circ x_k   \ \ \ \ \text{s.t.}  \ \ w_k \in \{0,1\}^N, \  \sum_{k=1}^2 w_k= \textbf{1},
\end{equation}
where $\circ$ denotes the element-wise product \cite{hadamard}, and $w_k$'s are the binary masks, where at each element one and only one of the $w_k$'s is 1, and the rest are zero. 
The constraint $\sum_{k=1}^2 w_k= \textbf{1}$, results in $w_1= \textbf{1}- w_2$.
In our work, we assume that each component can be represented with a known subspace/dictionary.
In this case, the main goal is to find the binary mask $w_k$ which denotes the support of each signal components.
This happens in many signal decomposition applications, such as image segmentation, text extraction from images, face detection with occlusion, and cyclic alternating pattern using masking signals \cite{cycling}.
\\To show an example application of this problem, consider the left image in Figure 1, where some text is overlaid on top of a textured background using a binary mask as shown in the right figure. 
As it can be seen from this image, the pixel values at the text locations are not affected by the background image. 
\begin{figure}[h]
\begin{center}
 \hspace{-0.1cm}  \includegraphics [scale=0.16] {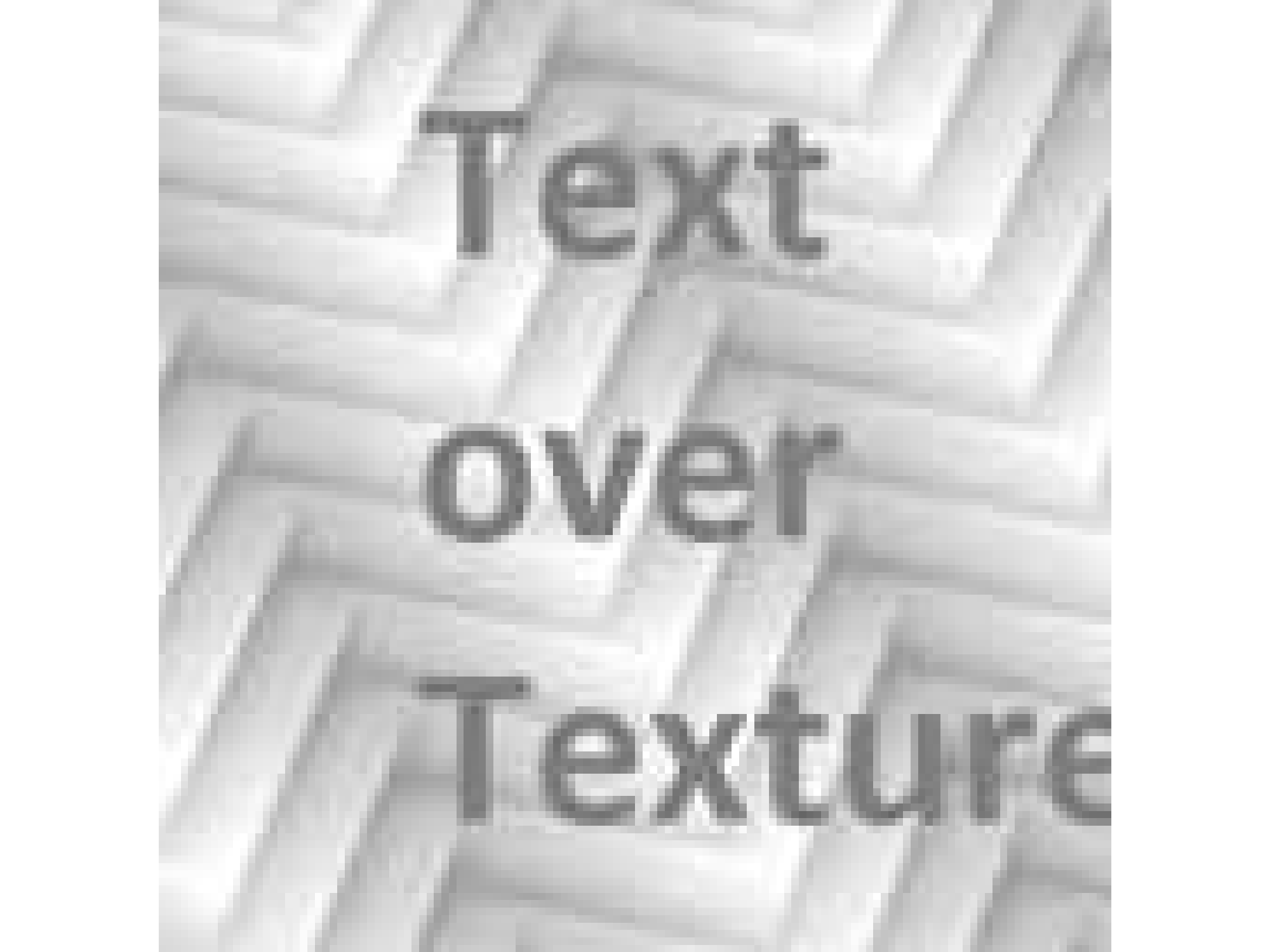}
 \hspace{-0.1cm}  \includegraphics [scale=0.16] {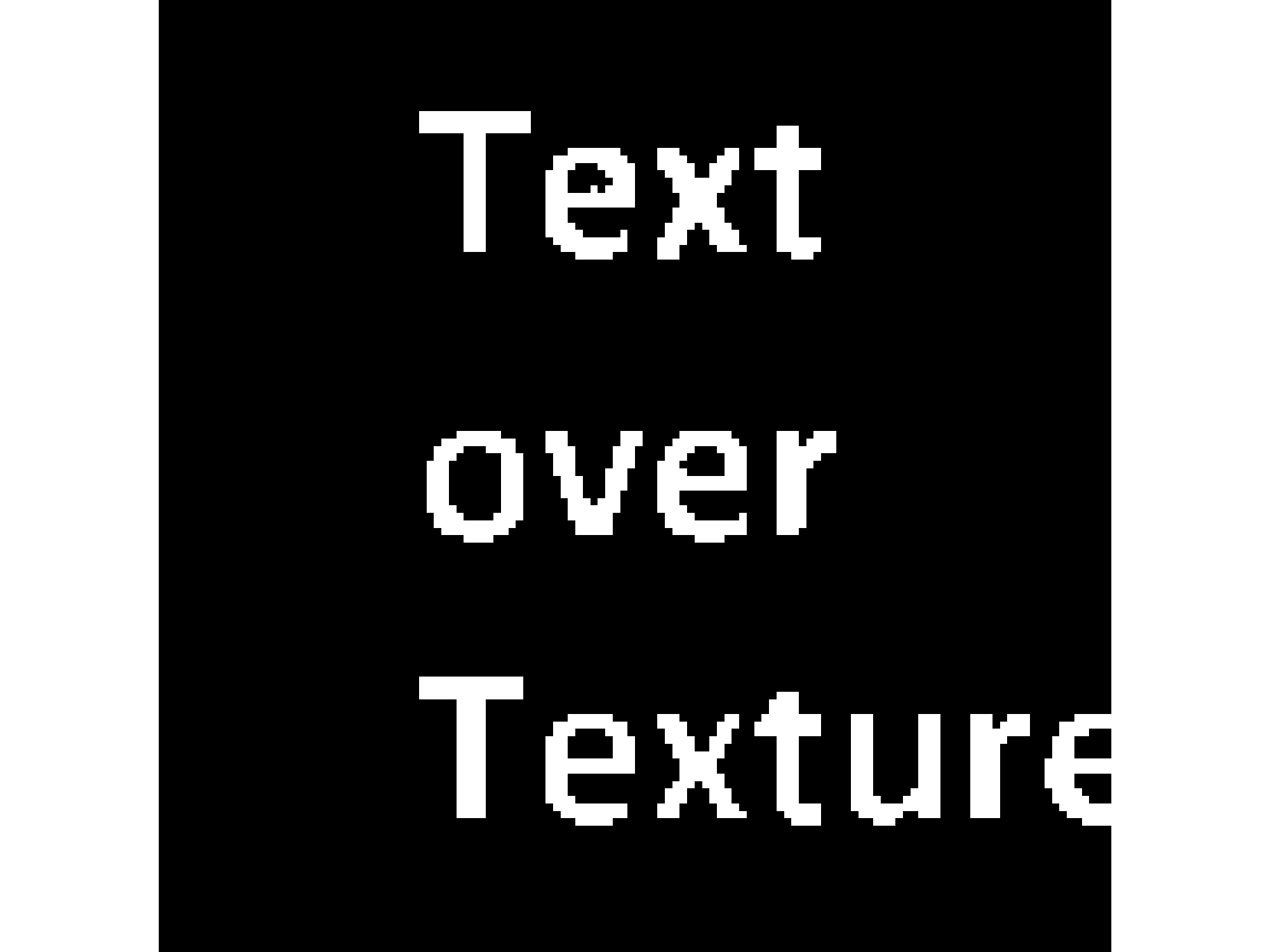} 
\end{center}
 \caption{The overlaid image and corresponding foreground mask}
\end{figure}

Different algorithms are used to separate the text from background, such as clustering-based algorithms, sparse-decomposition based methods, and morphological operations \cite{clus}-\cite{sp_text}. 
Figure 2 denotes a comparison between the segmentation results using hierarchical k-means approach \cite{clus2}, sparse decomposition \cite{mytv}, and the proposed algorithm in this work.
As we can see, each of the previous approaches have their own difficulties. 
For example, clustering-based scheme would have difficulty separating text from background in the case where the text has a similar color to the background. 
The sparse-decomposition based model misses some part of the text, while detecting some part of the background as text.
The proposed algorithm performed very well in separating the text from the background texture.
Note that the main difference between the sparse decomposition method \cite{mytv} and the  proposed algorithm is that the former assuming the background and foreground components are additive, while  the proposed method explicitly take into account that the foreground is overlaid on top of the background.
We would like to note that we are not claiming that the proposed method is the best for text segmentation, but rather we  try to show that text segmentation is a good application of the proposed method.
\begin{figure}[h]
        \centering
        \hspace{-0.2cm}
        \begin{subfigure}[b]{0.18\textwidth}
                \includegraphics[width=\textwidth]{texture8_orig-eps-converted-to.pdf}
            \hspace{-3.5cm} 
        \end{subfigure}%
        ~ 
        \begin{subfigure}[b]{0.18\textwidth}
                \includegraphics[width=\textwidth]{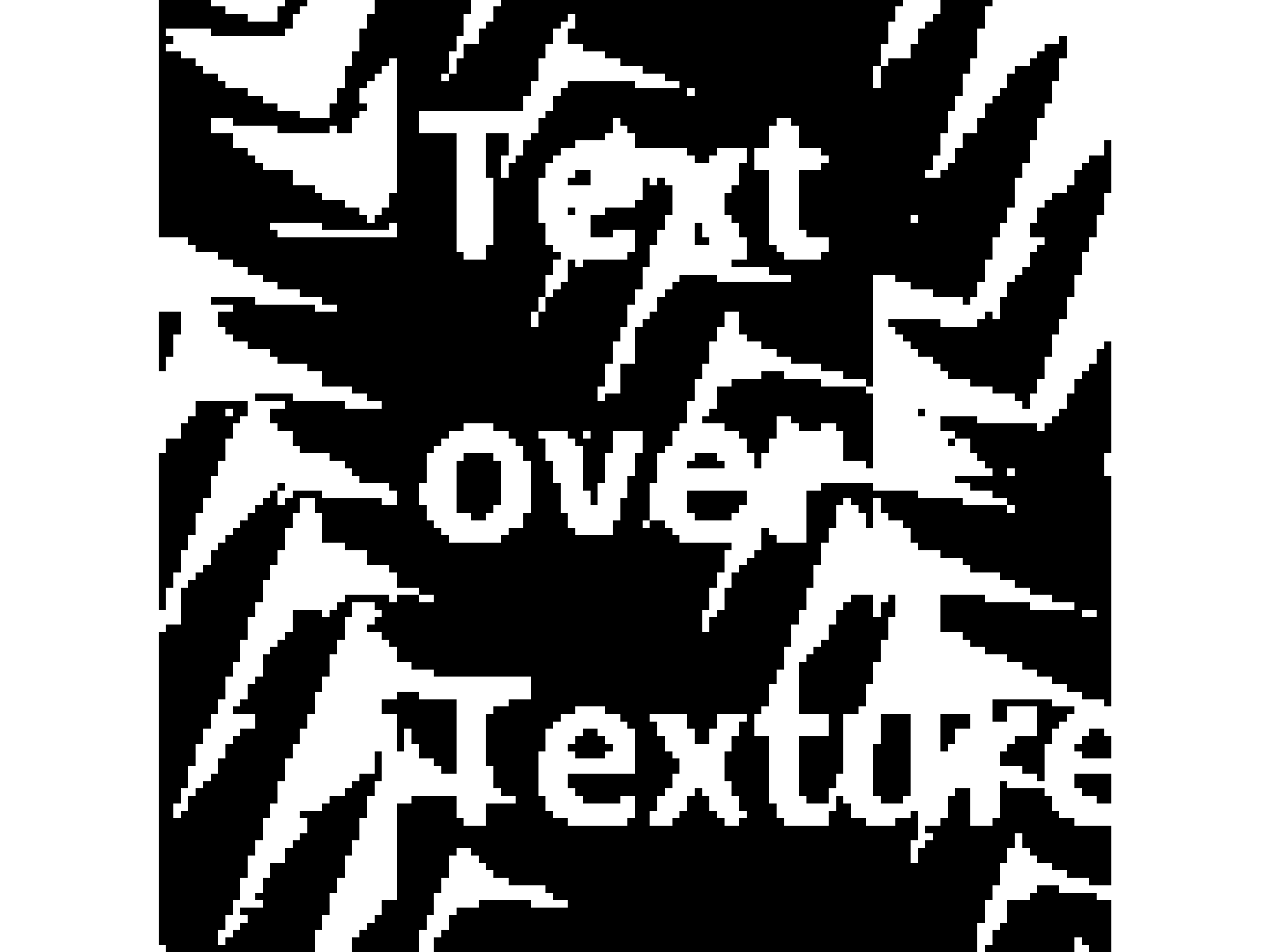}
              \hspace{-4.8cm}
        \end{subfigure}
         \\[1ex]
                 \centering
        \hspace{-1.4cm}

        \begin{subfigure}[b]{0.18\textwidth}
                \includegraphics[width=\textwidth]{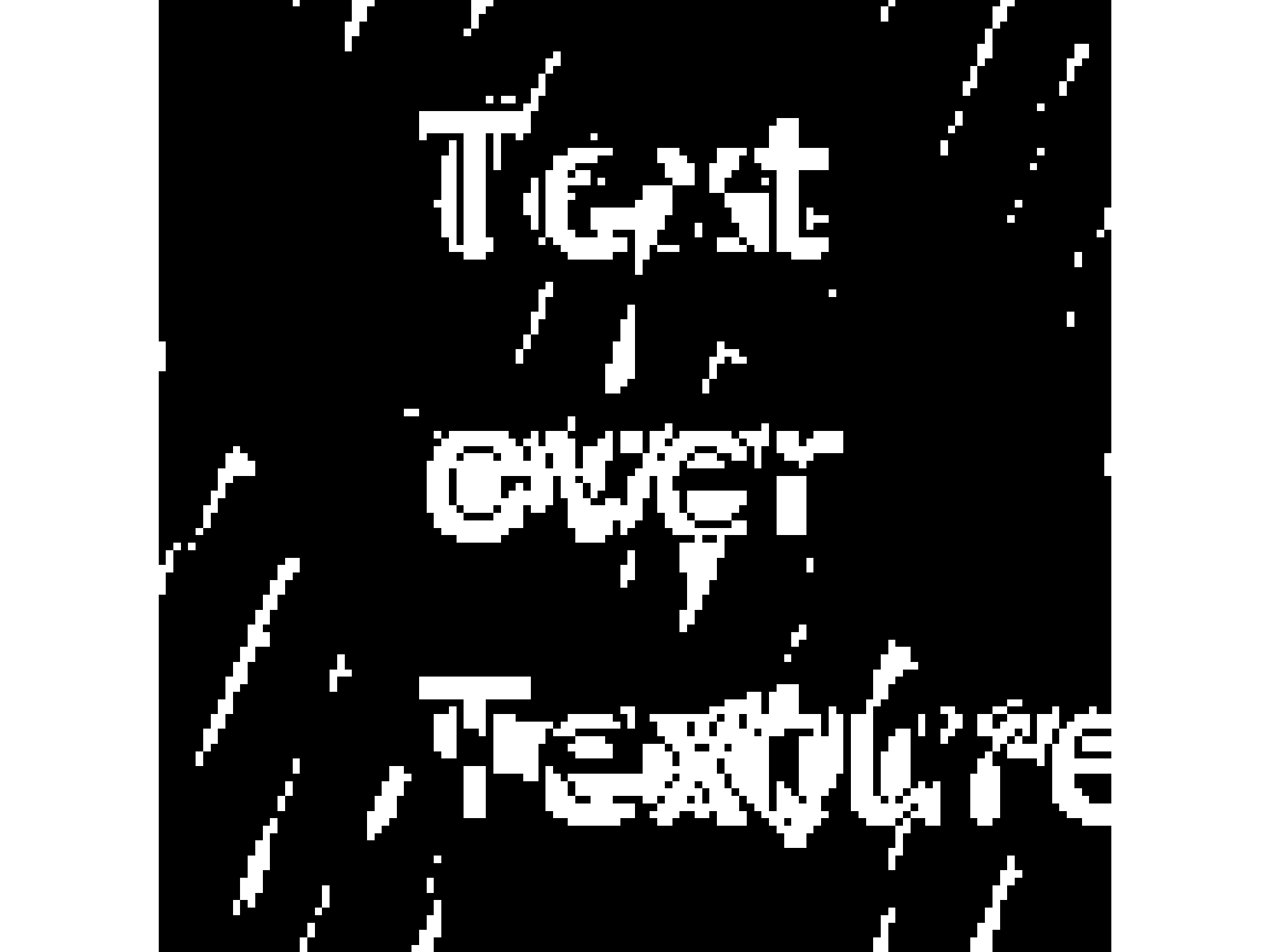}
            \hspace{-3cm} 
        \end{subfigure}%
        ~ 
        \begin{subfigure}[b]{0.18\textwidth}
                \includegraphics[width=\textwidth]{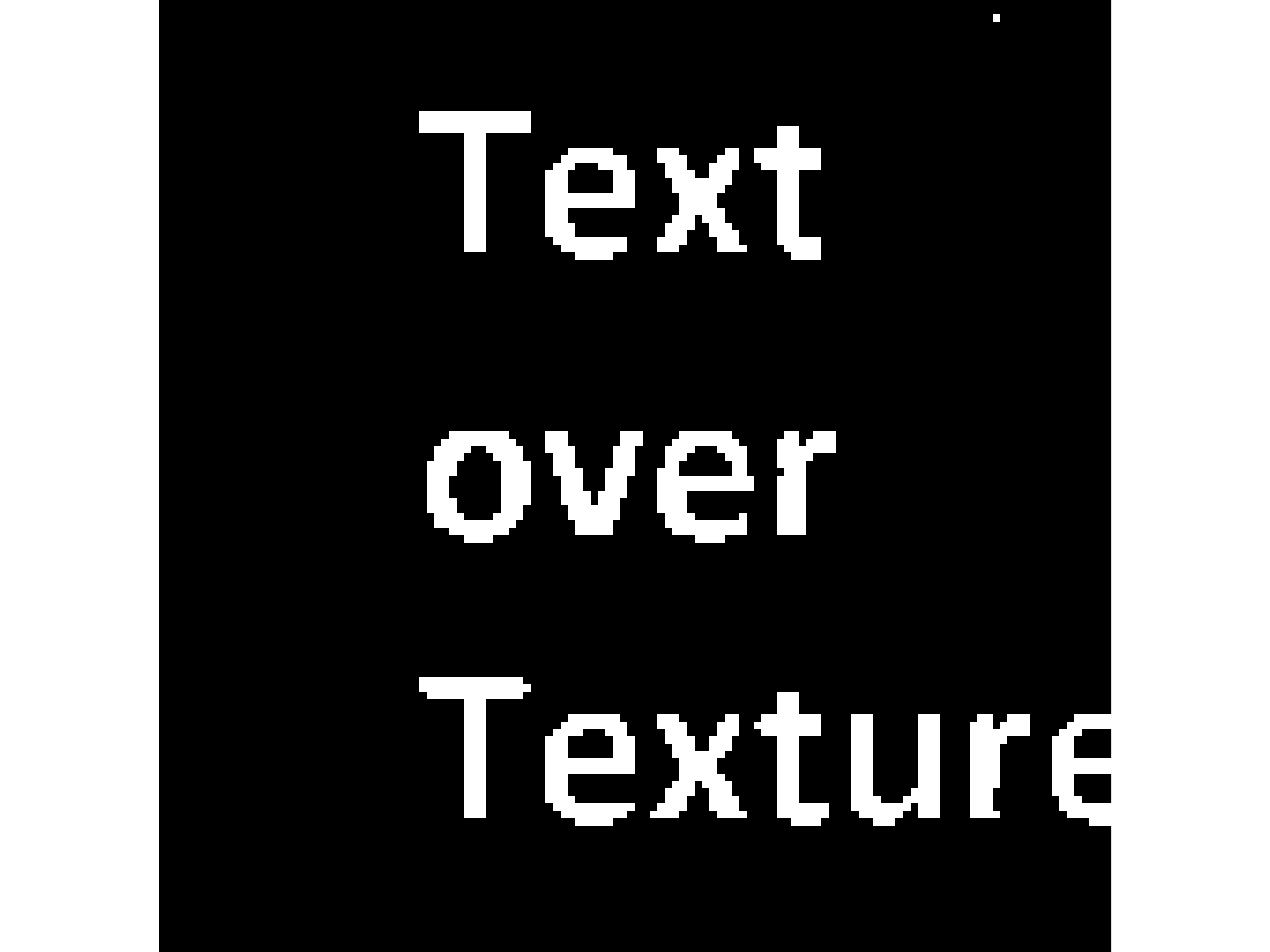}
              \hspace{-4.8cm}
        \end{subfigure}
         \\[1ex]
        \caption{ The top-right, bottom-left and bottom-right images denote the segmentation map by hierarchical clustering, sparse decomposition, and the proposed algorithm respectively.}
\end{figure}

The structure of the rest of this paper is as follows: Section II presents the problem formulation. 
Section III describes the augmented Lagrange multiplier to solve the binary mask estimation.
In Section IV, we show the application of this approach for motion segmentation.
The experimental results, and the applications are provided in Section V, and the paper is concluded in Section VI.

\section{Problem Formulation}
As discussed earlier, we consider a masked signal decomposition model with two components, described by $x= \sum_{k=1}^2 w_k \circ x_k$,
where $\circ$ denotes the element-wise product, and $w_k$'s are the binary masks (showing the support of each component), such that $w_2= \textbf{1}-w_1$.
If these components have different characteristics, it would be possible to separate them to some extent. 
One possible way is to form an optimization problem as below:
\begin{equation}
\begin{aligned}
& \hspace{1.4cm} \underset{w_k, x_k}{\text{min}}
 \ \phi_1(x_1, w_1)+\phi_2(x_2, w_2) \\
& \ \text{s.t.}
\ \ w_k \in \{0,1\}^n, \ w_2= \textbf{1}-w_1 ,\  \sum_{k=1}^2 w_k \circ x_k= x,
\end{aligned}
\end{equation}
where $\phi_k$ encodes our prior knowledge about each component and its corresponding mask.

One prior knowledge that we assume is that each component $x_k$ has a sparse representation using some proper dictionary/bases $P_k$.
To make the formulation easier to explain, we denote the  binary masks corresponding to the first and second component as $\textbf{1}-w$ and $w$ respectively. 
In this case we can drop the index subscript, and get:
\begin{equation}
x= (\textbf{1}-w) \circ {(P_1\alpha_1)}+ w \circ  (P_2\alpha_2) \ ,
\end{equation}
where $w \in \{0,1\}^N$, and $P_k$ is $N\times M_k$ matrix, where each column denotes one of the basis functions from the corresponding subspace/dictionary, and $M_k$ is the number of basis functions for component $k$.

Note that, in an alternative notation, (5) can be written as:
\begin{equation}
x= (I-W) {P_1\alpha_1}+ W P_2\alpha_2 \ ,
\end{equation}
where $W= \text{diag}(w)$ is a diagonal matrix with the vector $w$ on its main diagonal.
If a diagonal element is 1, the corresponding element belongs to component 2, otherwise to component 1.

The decomposition problem in Eq. (6) is a highly ill-posed problem. 
Therefore we need to impose some prior on each component, and also on $w$ to be able to perform this decomposition.
We assume that each component has a sparse representation with respect to its own subspace, but not with respect to the other one.
We also assume that the second component is sparse and connected. 
This would be the case, for example,  if the second component corresponds to text overlaid over a background image; or a moving object over a stationary background in a video frame.

To promote sparsity of the second component, we add the $\ell_0$ norm of $w$ to the cost function (note that $w$ corresponds to the support of the second component).
To promote connectivity, we can either add the group sparsity or total variation of $w$ to the cost function.
Here we use total variation.
The main reason is that for group sparsity, it is not very clear what is the best way to define groups, as the foreground pixels could be connected in any arbitrary direction, whereas total variation can deal with this arbitrary connectivity more easily.\\

We can incorporate all these priors in an optimization problem as shown below:
\begin{small}
\begin{equation}
\begin{aligned}
& \hspace{-0.2cm}\underset{w, \alpha_1, \alpha_2}{\text{min}}
 \  \frac{1}{2} \| x- (\textbf{1}-w) \circ P_1\alpha_\textbf{1}-w \circ P_2\alpha_2  \|_2^2+ \lambda_1 \| w \|_0+ \lambda_2 \text{TV}(w)   \\
& \ \text{s.t.}
\ \ \ \ \ \ w \in \{0,1\}^N  , \ \| \alpha_1 \|_0 \leq K_1 , \ \| \alpha_2 \|_0 \leq K_2.
\end{aligned}
\end{equation}
\end{small}

Total variation of 1D signals $w=[w_1, w_2,  ..., w_N]^T$ is straightforward, and can be defined as:
\begin{equation}
\begin{aligned}
TV(w)=  \sum_{n=1}^{N-1} |w_{n+1}-w_{n}|= \| D_1w\|_1 ,
\end{aligned}
\end{equation}
where $D_1$ is a $(N-1)\times N$ matrix as below:
\begin{equation*}
D_1= 
\begin{bmatrix}
    -1       & 1 & 0 & \dots & 0 \\
    0       & -1 & 1 & \dots & 0 \\
    \vdots & \vdots & \vdots & \ddots & \vdots \\
    0       & 0 & \dots &  -1 & 1
\end{bmatrix}
\end{equation*}

For 2D signals $w= [w_{i,j}, i=1,...,I, j=1,...,J]$, we can  either use the isotropic or the anisotropic version of 2D total variation \cite{TV}. To make our optimization problem simpler, we have used the anisotropic version in this algorithm, which is defined as:
\begin{equation}
\begin{aligned}
TV(w)=  \sum_{i,j} |w_{i+1,j}-w_{i,j}|+|w_{i,j+1}-w_{i,j}|.
\end{aligned}
\end{equation}
After converting the 2D blocks into 1D vector, we can denote this total variation as below:
\begin{equation}
\begin{aligned}
TV(w)= \| D_xw \|_1+\| D_yw \|_1= \|Dw\|_1 \ ,
\end{aligned}
\end{equation}
where $D_x$ and $D_y$ are the horizontal and vertical difference operator matrices, and $D=[ D_x',D_y']'$.\\

The problem in (7) involves multiple variables, and is not tractable, both because of the  $\| w \|_0$ term in the cost function and also the binary nature of $w $. We relax these conditions to be able to solve this problem in an alternating optimization approach. 
We replace the $\| w \|_0$ in the cost function with $\| w \|_1$, and also relax the $w \in \{0,1\}^N$ condition to $w \in [0,1]^N$ (which is known as linear relaxation in the mixed integer programming).
Then we will get the following optimization problem:

\begin{small}
\begin{equation}
\begin{aligned}
& \hspace{-0.2cm}\underset{w, \alpha_1, \alpha_2}{\text{min}}
  \frac{1}{2} \| x- (\textbf{1}-w)\circ  P_1\alpha_\textbf{1}-w \circ  P_2\alpha_2  \|_2^2+ \lambda_1 \| w \|_1+ \lambda_2 \| Dw \|_1   \\
& \ \text{s.t.}
\ \ \ \ \ \ w \in [0,1]^N  , \ \| \alpha_1 \|_0 \leq K_1 , \ \| \alpha_2 \|_0 \leq K_2.
\end{aligned}
\end{equation}
\end{small}

This problem can be solved with different approaches, such as majorization minimization, alternating direction method, and random sampling approach \cite{lag1}-\cite{ransac2}.
Here we present an algorithm based on alternating direction method of Lagrange multipliers (ADMM) \cite{lag1} to solve this problem. 

\section{The Proposed Optimization Framework}
ADMM is a  popular  algorithm  which  combines  superior  convergence
properties of method of multiplier and decomposability of dual
ascent. 
To solve the optimization problem in Eq. (12) with Augmented Lagrangian algorithm, we first introduce two auxiliary random variables as:

\begin{small}
\begin{equation}
\begin{aligned}
& \underset{w, \alpha_1, \alpha_2}{\text{min}}
 \  \ \frac{1}{2} \| x- (\textbf{1}-w)\circ  P_1\alpha_\textbf{1}-w \circ  P_2\alpha_2  \|_2^2+ \lambda_1 \| y \|_1+ \lambda_2 \| z \|   \\
& \ \text{s.t.}
\ \ \ \ \ \ w \in [0,1]^N, \ y= w,\ z= Dw, \| \alpha_1 \|_0 \leq K_1 ,  \| \alpha_2 \|_0 \leq K_2.
\end{aligned}
\end{equation}
\end{small}

To solve this problem, we form the augmented Lagrangian as below:
\begin{small}
\begin{equation}
\begin{aligned}
L(\alpha_1, \alpha_2, w, y, z, u_1, u_2)=  \frac{1}{2} \| x- (\textbf{1}-w)\circ  P_1\alpha_\textbf{1}-w \circ  P_2\alpha_2  \|_2^2 \hspace{2cm} \\ + \lambda_1 \| y \|_1+ \lambda_2 \| z \|_1+ u_1^t (w-y)+ u_2^t (Dw-z)+ \frac{\rho_1}{2} \|w-y\|_2^2 \hspace{2.2cm} \\
+ \frac{\rho_2}{2} \|Dw-z\|_2^2  \hspace{8.54cm}  \\
\text{s.t.}
\ \ \ w \in [0,1]^N,  \| \alpha_1 \|_0 \leq K_1 ,  \| \alpha_2 \|_0 \leq K_2, \hspace{4.76cm}
\end{aligned}
\end{equation}
\end{small}
where $u_1$ and $u_2$ denote the dual variables.
Now we can solve this problem by minimizing the Augmented Lagrangian w.r.t. to primal variables ($\alpha_1$, $\alpha_2$, $w$, $y$ and $z$) and using dual ascent for dual variables ($u_1$, $u_2$). 
For updating the variables $\alpha_1$ and $\alpha_2$, we first ignore the constraints and take the derivative of $L$ w.r.t. them and set it to zero. Then we project the solution on the constraint $\| \alpha_i \|_0 \leq K_i $ by keeping the $K_i$ largest components (in absolute value sense).
Since the cost function is symmetric in $\alpha_1$ and $\alpha_2$, we only show the solution for $\alpha_2$ here.
The solution for $\alpha_1$ is very similar.

\begin{small}
\begin{equation*}
\begin{aligned}
  \alpha_2=  \underset{ \alpha_2}{\text{\ \ argmin}}
  L(\alpha_1, \alpha_2, w, y, z, u_1, u_2) = \hspace{2.75cm}
  \\ \underset{ \alpha_2}{\text{\ \ argmin}} 
 \  \| x- (\textbf{1}-w)\circ  P_1\alpha_\textbf{1}-w \circ  P_2\alpha_2  \|_2^2= \hspace{2.2cm} \\
  \underset{ \alpha_2}{\text{\ \ argmin}} 
 \  \| x- (I-W) P_1\alpha_\textbf{1}-W  P_2\alpha_2  \|_2^2,  \text{ setting } \nabla_{\alpha_2}L=0: \hspace{0.1cm} \\
  P_2^tW^t \big( WP_2\alpha_2+(I-W)P_1\alpha_1-x \big)=0 \hspace{2.62cm} \\
 \Rightarrow \alpha_2= (P_2^tW^tWP_2)^{-1} P_2^tW^t (x-(I-W)P_1\alpha_1). \hspace{1.29cm}
\end{aligned}
\end{equation*}
\end{small}

Now we keep the $K_2$ largest components of the above $\alpha_2$, which is denoted by: $\alpha_2^*= \Pi_{top-K_2}(\alpha_2)$.

We now show the optimization with respect to $w$.
We solve this optimization by first ignoring the constraint, and then projecting the optimal solution of the cost function onto the feasible set ($w \in [0,1]^n$).
\\It basically follows the same methodology, we just need to notice that $\text{diag}(w) P_2\alpha_2$ is the same as
$ \text{diag}(P_2\alpha_2) w $.
Therefore we will get the following optimization for $w$:

\begin{small}
\begin{equation*}
\begin{aligned}
w=  \underset{ w}{\text{\ argmin}}
\ \frac{1}{2} \| x- \text{diag}(P_1\alpha_1)(\textbf{1}-w) - \text{diag}(P_2\alpha_2) w  \|_2^2+ \hspace{0.1cm} \\
u_1^t (w-y)+ u_2^t (Dw-z)+ \frac{\rho_1}{2} \|w-y\|_2^2+ \frac{\rho_2}{2} \|Dw-z\|_2^2.  \hspace{0.14cm}  
\end{aligned}
\end{equation*}
\end{small}

\hspace{-0.45cm} We can rewrite this problem as:
\begin{small}
\begin{equation*}
\begin{aligned}
w= \underset{w}{\text{argmin}}
 \  \ \frac{1}{2} \| h - C w  \|_2^2+ \frac{\rho_1}{2} \|w-y\|_2^2+ \frac{\rho_2}{2} \|Dw-z\|_2^2  \hspace{0.8cm} \\
 + u_1^t (w-y)+ u_2^t (Dw-z), \hspace{5.1cm}
\end{aligned}
\end{equation*}
\end{small}
\hspace{-0.24cm} where $C= \text{diag}(P_2\alpha_2)-\text{diag}(P_1\alpha_1)= \text{diag}(P_2\alpha_2-P_1\alpha_1)$, and $h= x- P_1\alpha_1$.
If we take the derivative w.r.t. $w$ and set it  to zero we will get:

\begin{small}
\begin{equation*}
\begin{aligned}
C^t(Cw-h)+\rho_1 (w-y)+ \rho_2 D^t(Dw-z)+u_1+D^t u_2=0 \Rightarrow \\
 (C^tC+ \rho_2 D^t D+ \rho_1 I)w= C^t h+ \rho_1 y+ \rho_2 D^t z-u_1-D^t u_2 \Rightarrow \\
 w= M_w^{-1} (C^t h+ \rho_1 y+ \rho_2 D^t z-u_1-D^t u_2), \hspace{2.5cm}
\end{aligned}
\end{equation*}
\end{small}
\hspace{-.25cm} where $M_w=(C^tC+ \rho_2 D^t D+ \rho_1 I) $.
After finding $w$ using the above equation, we need to project them on the set $w \in [0,1]^n$, which basically maps any negative number to 0, and any number larger than 1 to 1. Denoting the projection operator by $\Pi_{[0,1]}$, the optimization solution of the $w$ step would be:
\begin{small}
\begin{equation}
\begin{aligned}
 \ w= \Pi_{[0,1]}\big( M_w^{-1} (C^t h+ \rho_1 y+ \rho_2 D^t z-u_1-D^t u_2) ).
\end{aligned} 
\end{equation}
\end{small}
The projection operator $\Pi_{[0,1]}$ is defined as below (which essentially maps any points to the closest point in the feasible set):
\begin{equation*}
  \Pi_{[0,1]}(x)=\begin{cases}
    1, & \text{if $x>1$}.\\
    0, & \text{if $x<0$}.\\
    x, & \text{otherwise}.
  \end{cases}
\end{equation*}

The optimization w.r.t. $y$ and $z$ are quite simple, as they result in a soft-thresholding solution \cite{soft}.
The overall algorithm is summarized in Algorithm 1.

\begin{algorithm}
\footnotesize
  \caption{pseudo-code for variable updates of problem (12)}\label{euclid}
  \begin{algorithmic}[1]
   \Statex Given a block of size NxN, represented by a vector x, and the subspace matrices $P_1$ and $P_2$, and preset values for parameters $\rho_1$, $\rho_2$, $T_{max}$, initialize the loss value with $L^{(0)}= 1$, and: \vspace{0.15cm}
      \For{\texttt{$j$=1:$T_{max}$}} \vspace{0.06cm} 
      \hspace{-0.2cm}  \State $\alpha_1^{j+1}= \Pi_{top-K_1}( (P_1^tW_*^tW_*P_1)^{-1} P_1^tW_*^t (x-WP_2\alpha_2^{j}))$ \vspace{0.15cm}
        \State $\alpha_2^{j+1}= \Pi_{top-K_2}((P_2^tW^tWP_2)^{-1} P_2^tW^t (x-(I-W)P_1\alpha_1^{j+1}))$ \vspace{0.2cm}
        \State $ w^{j+1}= \Pi_{[0,1]}\big( M_w^{-1} (C^t h^{j+1}+ \rho_1 y^{j}+ \rho_2 D^t z^{j}-u_1^{j}- \hspace{2.5cm} D^t u_2^{j})  $   \vspace{0.15cm} 
        \State $ y^{j+1}=  \text{soft}( w^{j+1}+ \frac{u_1^{j}}{\rho_1}, \lambda_1/\rho_1)     $    \vspace{0.15cm}  
        \State $ z^{j+1}= \text{soft}( Dw^{j+1}+\frac{u_2^{j}}{\rho_2}, \lambda_2/\rho_2)    $    \vspace{0.15cm}   
         \State $ u_1^{j+1}= u_1^{j}+ \rho_1 (w^{j+1}-y^{j+1})   $    \vspace{0.15cm}  
         \State $ u_2^{j+1}= u_2^{j}+ \rho_2 (Dw^{j+1}-z^{j+1})  $    \vspace{0.15cm} 
          
         \State $ \textbf{if} \ \ \frac{|L^{(j)}-L^{(j-1)}|}{L^{(j-1)}} \leq 10^{-6}  $    \vspace{0.15cm} 
		\State    \ \ \ \ \ \text{stop iterating}  \vspace{0.15cm} 
		\State \textbf{end if}         
         
      \EndFor
      \Statex 
      \Statex Where  $C= \text{diag}(P_2\alpha_2-P_1\alpha_1)$, $h= x- P_1\alpha_1$, 
      \Statex  $W= \text{diag}(w)$, $W_*= I-W$
    \Statex and $M_w=(C^tC+ \rho_2 D^t D+ \rho_1 I)$
  \end{algorithmic}
\end{algorithm}
In Algorithm 1, $\text{soft}(x,\lambda)$ denotes the soft-thresholding operator \cite{soft}, applied element-wise and defined as:
\begin{small}
\begin{gather*}
\text{Soft}(x,\lambda)= \text{sign}(x) \ \text{max}(|x|-\lambda,0).
\end{gather*}
\end{small}
In terms of computational complexity, assume the image block size is $N \times N$, the number of basis functions in $P_1$ and $P_2$ are $K_1$ and $K_2$ respectively, then the vectors $\alpha_1$, $\alpha_2$ and $x$ have dimensions $K_1 \times 1$, $K_2 \times 1$ and $N^2 \times 1$, and the matrices $P_1$, $P_2$ and $W$ have dimensions of $N^2 \times K_1$, $N^2 \times K_2$ and $N^2 \times N^2$, but $W$ is a diagonal matrix.
Then it is easy to see that the first three updates in Algorithm 1 are the bottleneck for computational complexity, because they involve matrix multiplication and inversion. 
Assuming the computational complexity of inverting an $d \times d$ matrix to be $O(d^3)$ (in reality there are  faster inversion techniques such as Coppersmith-Winograd algorithm which is of $O(d^{2.376}$)), 
it can be shown than the overall computational complexity of the algorithm for a block of $N\times N$ is going to be $O(N^6+(K_1^2+K_2^2)N^2+K_1^3+K_2^3)$ (also, if we assume $P_1$ and $P_2$ are subspace representations, which means $K_1,K_2< N^2$, then the complexity of this algorithm for a block of $N \times N$ is going to be $O(N^6)$). 
\\Now for a given image of size $n \times m$, assuming $n$ and $m$ are much larger than $N$, we will have approximately around $nm/N^2$ blocks, resulting in a an overall complexity of $O(nm (N^4+(K_1+K_2)N^2+\frac{K_1^3+K_2^3}{N^2}) ) $, which for case where $P_1$ and $P_2$ represent a subspace, it will be simplified to in a complexity of $O(nmN^4)$.

\section{Application for Robust Motion Segmentation}
One potential application of the proposed formulation is for moving object detection under global camera motion in a video, or essentially segmentation of a motion field into regions with global motion and object motions, respectively.

Suppose we use the homography mapping (also known as perspective mapping) to model the camera motion, where each pixel in the new frame is related to its position in the previous frame by:

\begin{equation}
\begin{aligned}
x_{new}= \frac{a_1+a_2x+a_3y}{1+a_7x+a_8y} \hspace{0.08cm} \\
y_{new}= \frac{a_4+a_5x+a_6y}{1+a_7x+a_8y}.
\end{aligned} 
\end{equation}
Let $u=x_{new}-x$ and $v=x_{new}-x$, we can rewrite the above equation as:
\begin{equation}
\begin{aligned}
(x+u)(1+a_7x+a_8y)= (a_1+a_2x+a_3y) \hspace{0.08cm} \\
(y+v)(1+a_7x+a_8y)= (a_4+a_5x+a_6y).
\end{aligned} 
\end{equation}

For each pixel $(x,y)$ and its motion vector $(u,v)$, we will get two equations for the homograph parameters $a= [a_1,...,a_8]^T$, that can be written as:
\begin{small}
\begin{equation*}
\begin{bmatrix} 
    1&x&y&0&0&0&-x(x+u)&-y(x+u) \\ 0&0&0&1&x&y&-x(y+v)&-y(y+v)
\end{bmatrix} a= 
\begin{bmatrix} 
    x+u \\ y+v
\end{bmatrix}.
\end{equation*}
\end{small}
\hspace{-0.2cm} Using  the equations at all pixels, we will get a matrix equation as:
\begin{equation*}
\begin{aligned}
Pa=b.
\end{aligned} 
\end{equation*}

Suppose $u(x,y)$ and $v(x,y)$ are derived using a chosen optical flow estimation algorithm. 
Then the goal is to find the global motion parameters and the set of pixels which do not follow the global motion. 
Note that here, $P$ will not be the same for different video frames, as it depends on the optical flow (which could be different for different frames).
Assuming there are some outliers (corresponding to moving objects) in the video, we can use the model $b=(\textbf{1}-w) \circ Pa+w \circ s$, where $w$ denotes the outlier pixels, and $s$ denotes the new location for the outlier pixels.
If the outlier pixels belong to a single object with a consistent local motion, we can model $s$ as $s=P_2 a_2$.
However, in general, the outlier pixels may correspond  to multiple foreground objects with different motions or different subregions of a single object (e.g. different parts of a human body) with different motions.  Therefore, we do not want to model s with a single parameterized motion. Rather we will directly solve for s with a sparsity  constraint.  These considerations lead to the following optimization problem:

\begin{small}
\begin{equation}
\begin{aligned}
& \underset{w, a, s}{\text{min}}
 \  \ \frac{1}{2} \| b- (\textbf{1}-w)\circ  Pa- w \circ  s  \|_2^2+ \lambda_1 \| s \|_1    \\ 
& + \lambda_2 \| w \|_1+\lambda_3 \|Dw \|_1 \hspace{0.3cm} \\
& \ \text{s.t}
\ \ \  w \in [0,1]^n.
\end{aligned}
\end{equation}
\end{small}
Note that $b$, $P$ and $s$ in Eq. (17) have two parts, one corresponding to the horizontal direction, and another part corresponding to vertical direction (denoted with subscripts $x$ and $y$ respectively). Therefore we can re-write this problem as:
\begin{equation}
\begin{aligned}
& \underset{w, a, s}{\text{min}}
 \  \frac{1}{2} \| b_x- (\textbf{1}-w)\circ  P_xa- w \circ  s_x  \|_2^2+ \lambda_1 \| s_x \|_1+ \lambda_1 \| s_y \|_1     \\ 
 & \ \ \ + \frac{1}{2} \| b_y- (\textbf{1}-w)\circ  P_ya- w \circ  s_y  \|_2^2+ \lambda_2 \| w \|_1+ \lambda_3 \| Dw \|_1 \hspace{0.6cm} \\
& \ \ \text{s.t}
\ \ \ \  w \in [0,1]^n.
\end{aligned}
\end{equation}

This problem can be solved with ADMM. After solving this problem we will get the mask for the moving objects.
Note that this approach works for other global motion models such as the affine mapping.
In the extended version of this algorithm, we can directly work on a volume of $\tau$ frames to use the temporal information for mask extraction. In that case, the mask $w$ would be a 3D tensor.

In the experimental result section, we provide the result of motion segmentation using the proposed algorithm.

\section{Experimental Results}
In this section we provide the experimental study on the application of the proposed algorithm for 1D signals decomposition, image segmentation, and also motion segmentation. 
For each application, different sets of parameters are used, which are tuned on a validation data from the same task. 

\subsection{1D signal decomposition}
To illustrate the power of the proposed algorithm for non-additive decomposition, we first experiment with a toy example using 1D signals. We generate two 1D signals, each 256 dimensional, using different subspaces. The first signal is generated from a 10-dimensional sinusoid subspace, and the second component is generated from a 10-dimensional Hadamard subspace.
We then generate a random binary mask with the same size as the signal, and added these two components using the mask as: $x= (\textbf{1}-w)\circ x_1+w\circ x_2$.
The goal is to separate these two components, and estimate the binary mask.
The signal components and binary mask for one example are shown in Figure 3.
\begin{figure}[h]
\begin{center}
 \hspace{-0.3cm} \includegraphics [scale=0.64] {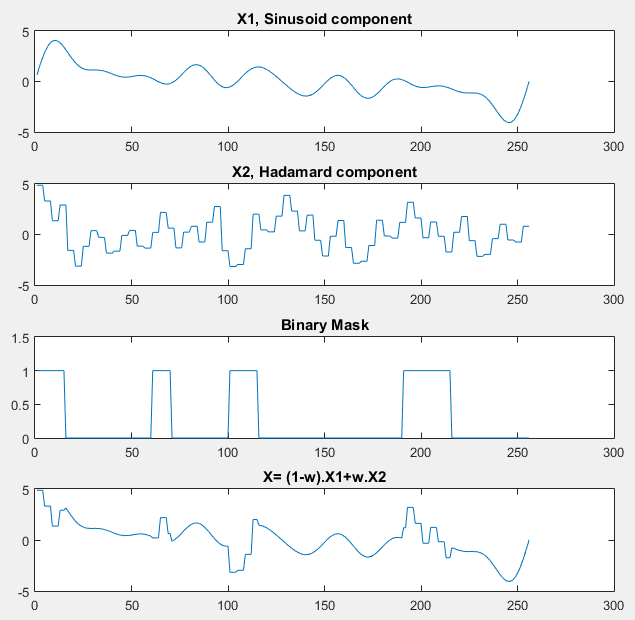} 
\end{center}
 \vspace{-0.1cm} \caption{The binary mask, and signal components}
\end{figure}

We then use the proposed model to estimate each signal component and extract the binary mask, and compare it with the signal decomposition under additive model. By additive model we mean the following optimization problem:
\begin{equation}
\begin{aligned}
& \underset{\alpha_1, \alpha_2}{\text{min}}
 \  \frac{1}{2} \| f-  P_1\alpha_1-  P_2\alpha_2  \|_2^2+ \lambda_1 \| P_2\alpha_2 \|_1+ \lambda_2 TV(P_2\alpha_2)    \\
& \ \text{s.t.}
\ \ \ \ \ \  \| \alpha_1 \|_0 \leq k_1 , \ \| \alpha_2 \|_0 \leq k_2.
\end{aligned}
\end{equation}
We need to mention that for the above additive model, the binary mask is derived by thresholding the values of the second component (we adaptively chose the threshold values such that it yields the best visual results).
The estimated signal components and binary mask by each algorithm, for two examples are shown in Figure 4. 
In our experiment, the weight parameters for the regularization terms in Eq. (12) are chosen to be $\lambda_1= 0.3$ and $\lambda_2= 10$.
The number of iterations for alternating optimization algorithm is chosen to be 20. 
As it can be seen the proposed algorithm achieves much better result than the additive signal model.
This is as expected, because the additive model could try to model some parts of the second component with the first subspace and vice versa.

\begin{figure*}
        \centering
        \begin{subfigure}[b]{0.99\textwidth}
                \includegraphics[width=\textwidth]{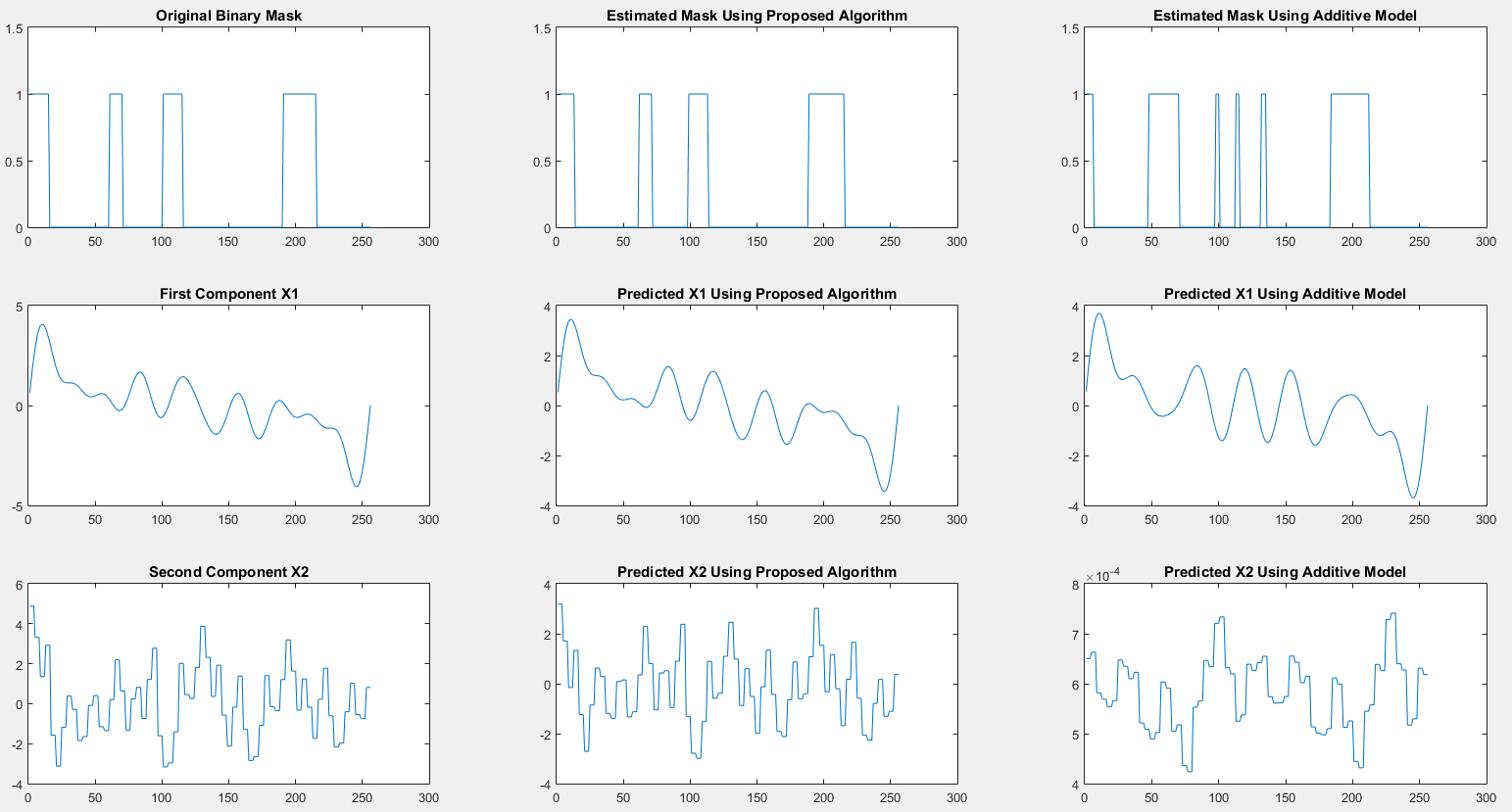}
                                \vspace{.5cm}    
          \hspace{-2.5cm}    
        \end{subfigure}
        
        \begin{subfigure}[b]{0.99\textwidth}
        \includegraphics[width=\textwidth]{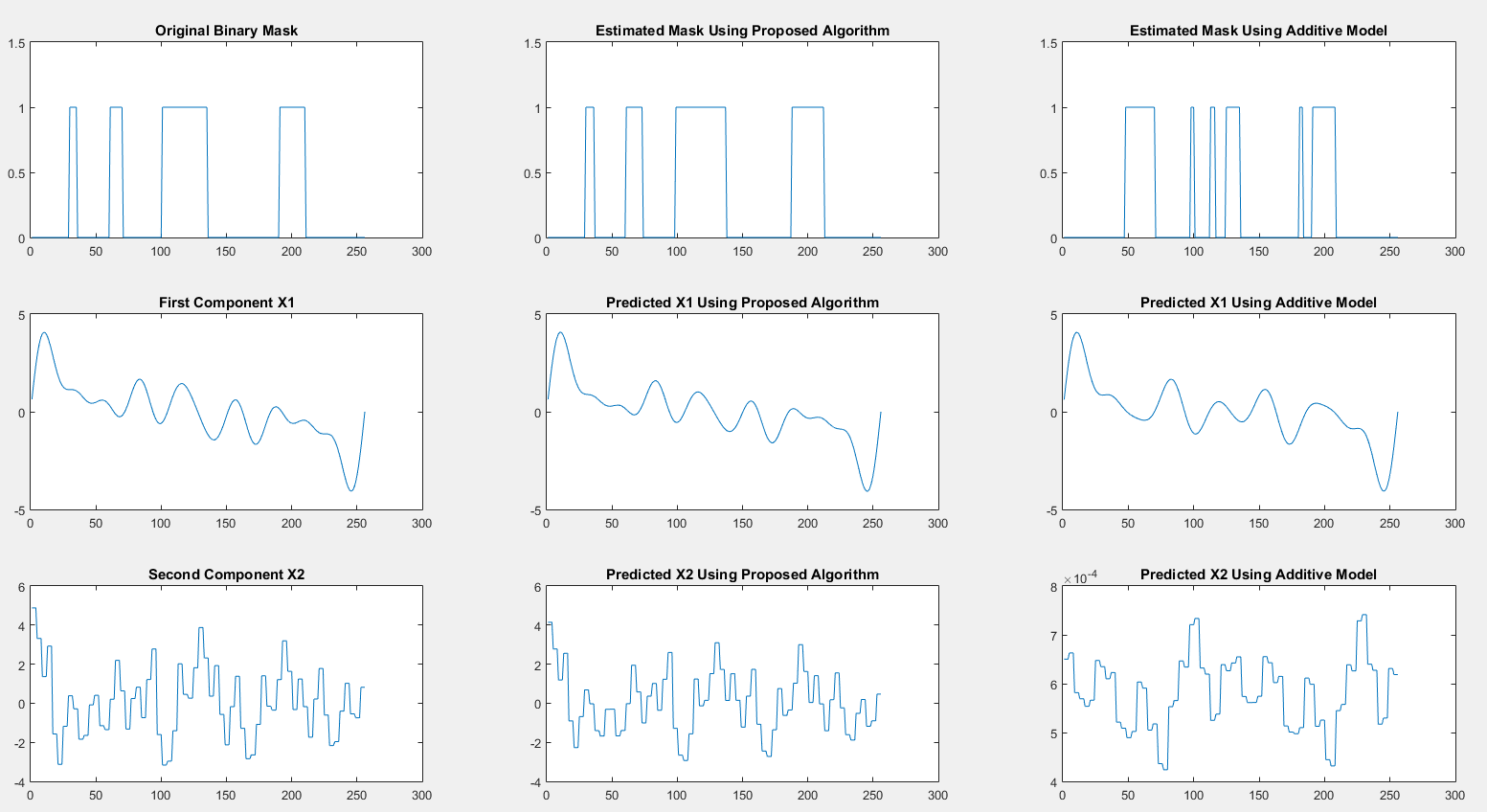}               
        \end{subfigure}
        \caption{The 1D signal decomposition results for two examples. The figures in the first column denotes the original binary mask, first and second signal components respectively. The second and third columns denote the estimated binary mask and signal components using  the proposed algorithm and the additive model signal decomposition, respectively. }
\end{figure*}

\subsection{Application in Text/Graphic Segmentation From Images}
Next we test the potential of the proposed algorithm for the text and graphic segmentation from images.
We perform segmentation on two different sets of images. 
The first one is on a dataset of screen content images, which consist of 332 image blocks of size 64x64, extracted from sample frames of HEVC test sequences for screen content coding \cite{SCC_data}, \cite{SCC_tran}.
The second set of images are generated manually by adding text on top of other images.

We apply our algorithm on blocks of 64x64 pixels. We first convert each block into a vector of  dimension 4096, and then apply the proposed algorithm.
For the smooth background we use low-frequency DCT basis with $k_1=40$, and for the second component we use Hadamard basis with $k_2=8$.
The weight parameters for the regularization terms are chosen to be $\lambda_1= 10$ and $\lambda_2= 0.2$, which are tuned by testing on a separate validation set of more than 50 patches.
The values of $\rho_1$ and $\rho_2$ are set to 1 in Algorithm 1.
The number of iterations for alternating optimization algorithm is chosen to be 10.
We compare the proposed algorithm with four previous algorithms: hierarchical k-means clustering in DjVu \cite{clus2}, SPEC \cite{spec}, least absolute deviation fitting (LAD) \cite{lad}, and sparsity based signal decomposition \cite{mytv}.

\begin{figure*}
        \centering
        \begin{subfigure}[b]{0.22\textwidth}
                \includegraphics[width=\textwidth]{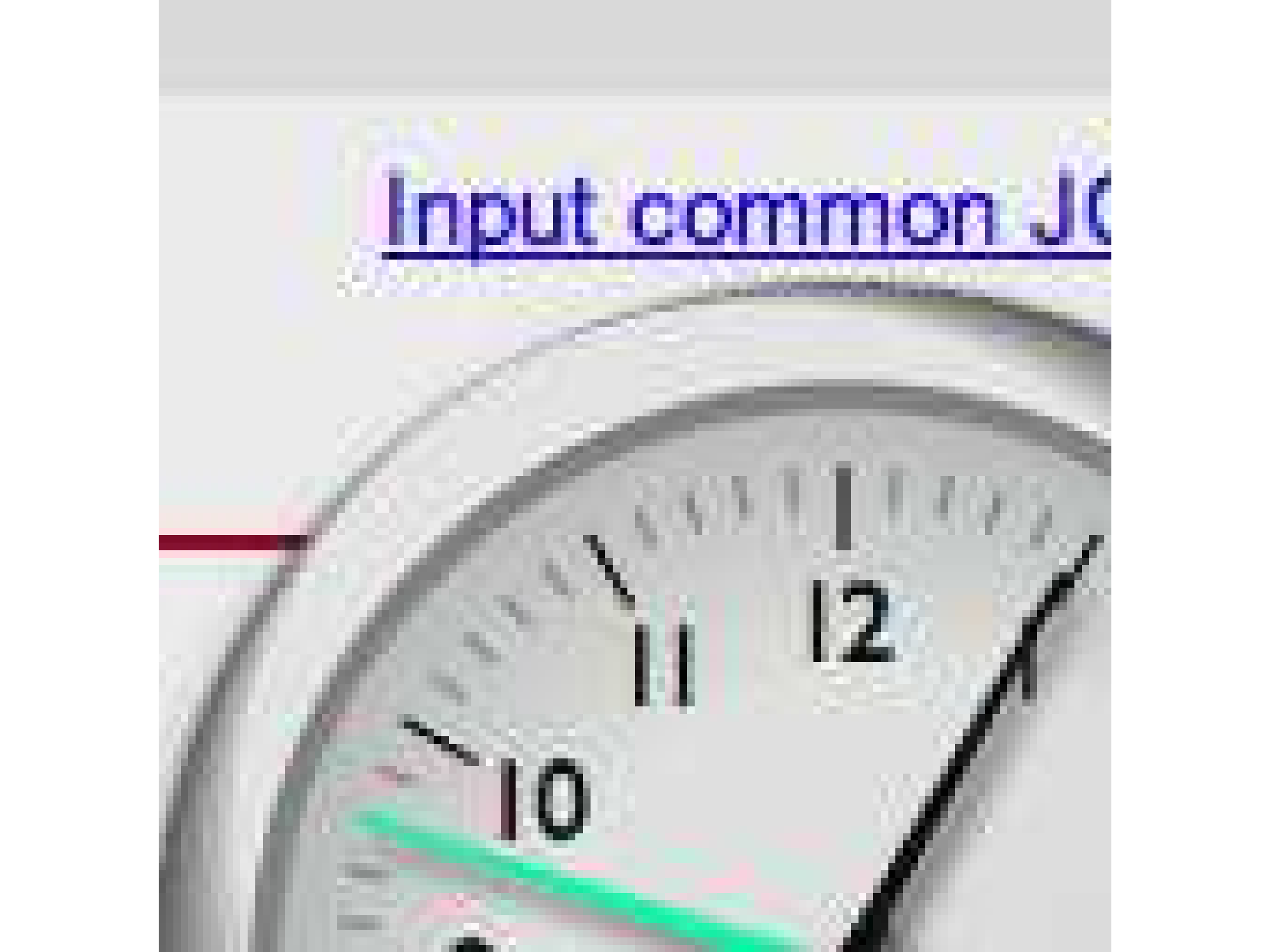}
          \hspace{-1.5cm}    
        \end{subfigure}%
        ~ 
        \begin{subfigure}[b]{0.22\textwidth}
                \includegraphics[width=\textwidth]{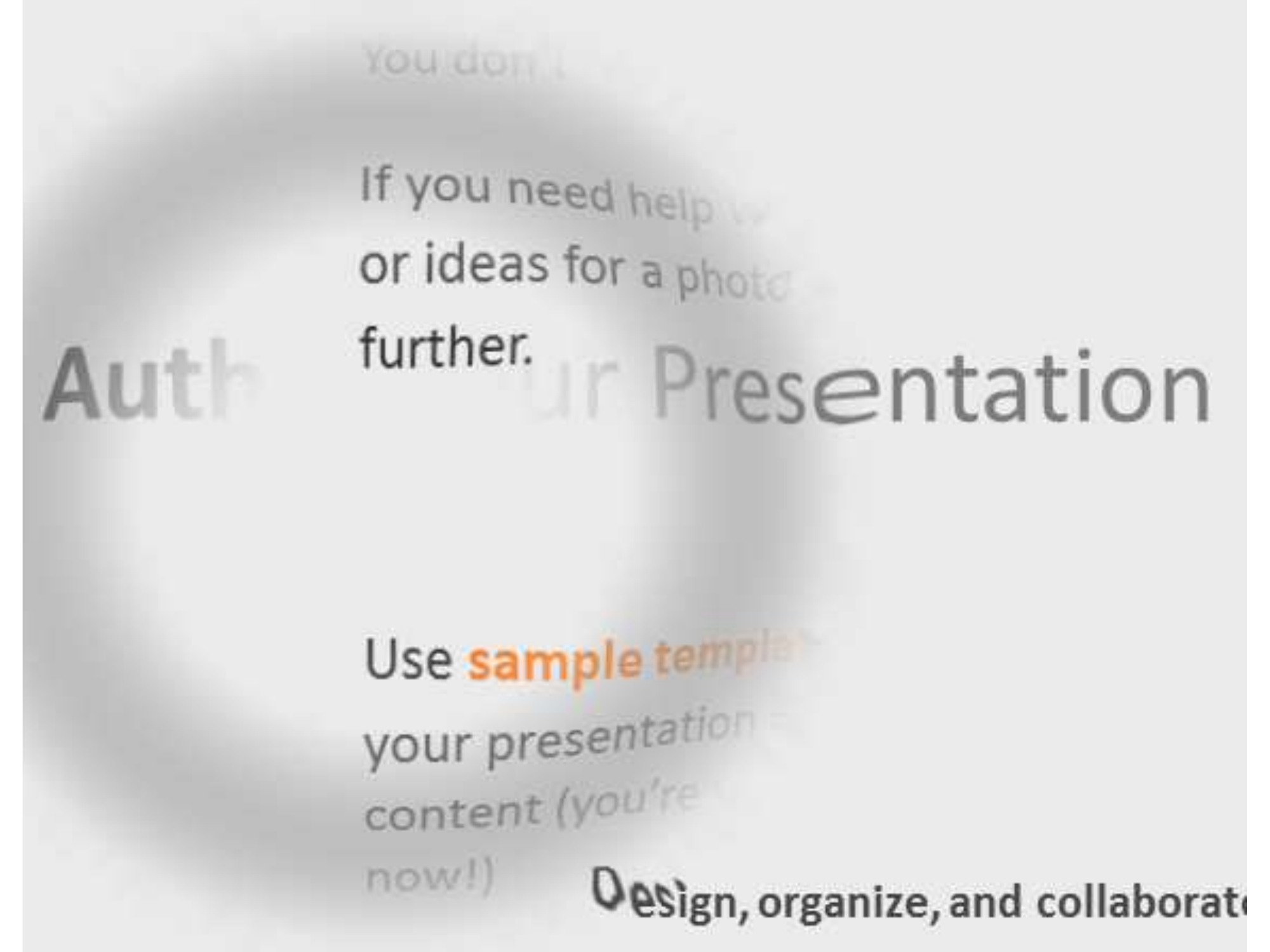}
            \hspace{-6cm} 
        \end{subfigure}%
        \begin{subfigure}[b]{0.22\textwidth}
			~ 
                \includegraphics[width=\textwidth]{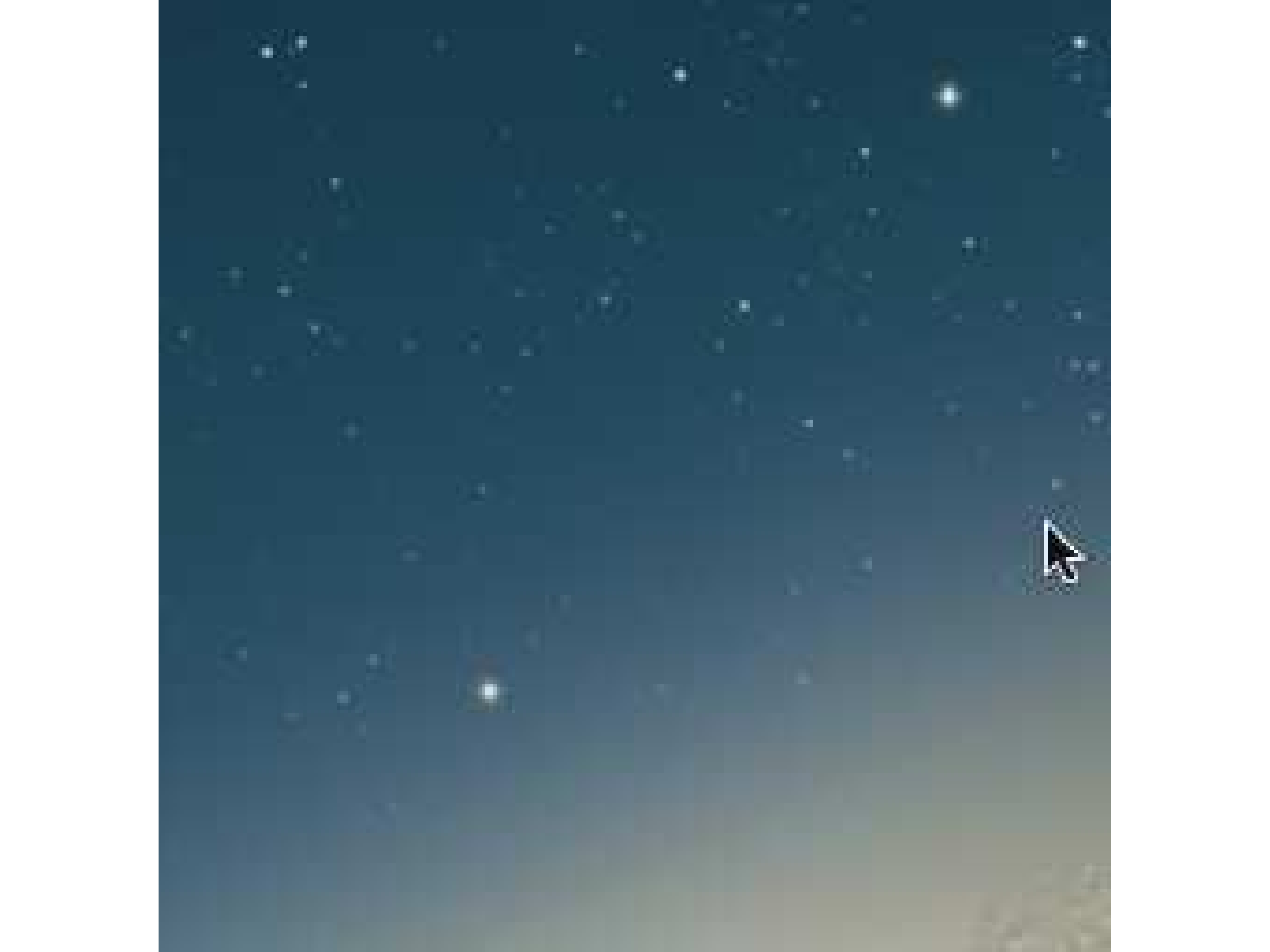}
            \hspace{-5cm} 
        \end{subfigure}%
        \begin{subfigure}[b]{0.22\textwidth}
                \includegraphics[width=\textwidth]{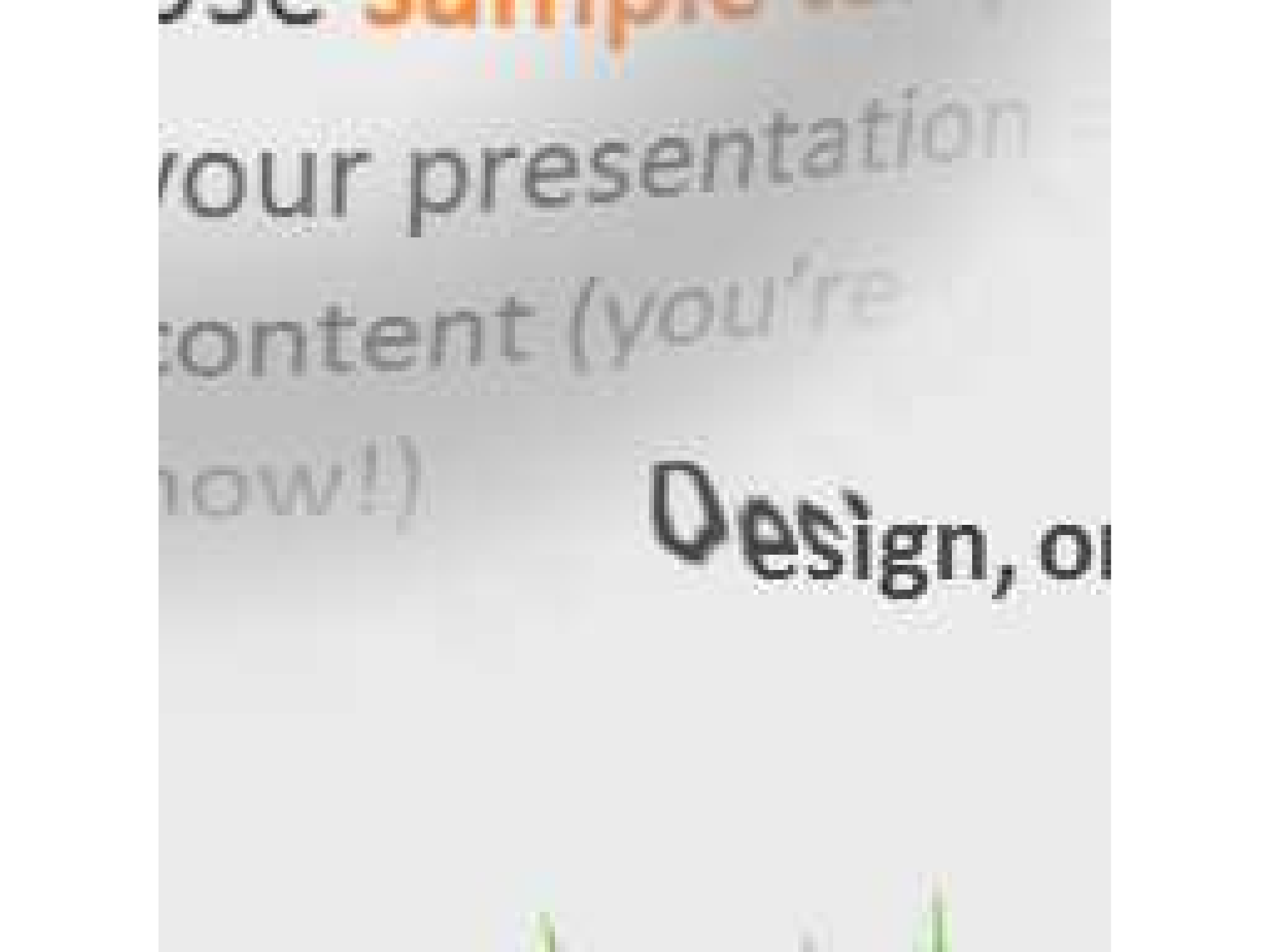}
              \hspace{-4.8cm}
        \end{subfigure}
         \\[1ex]
		\begin{subfigure}[b]{0.22\textwidth}
                \includegraphics[width=\textwidth]{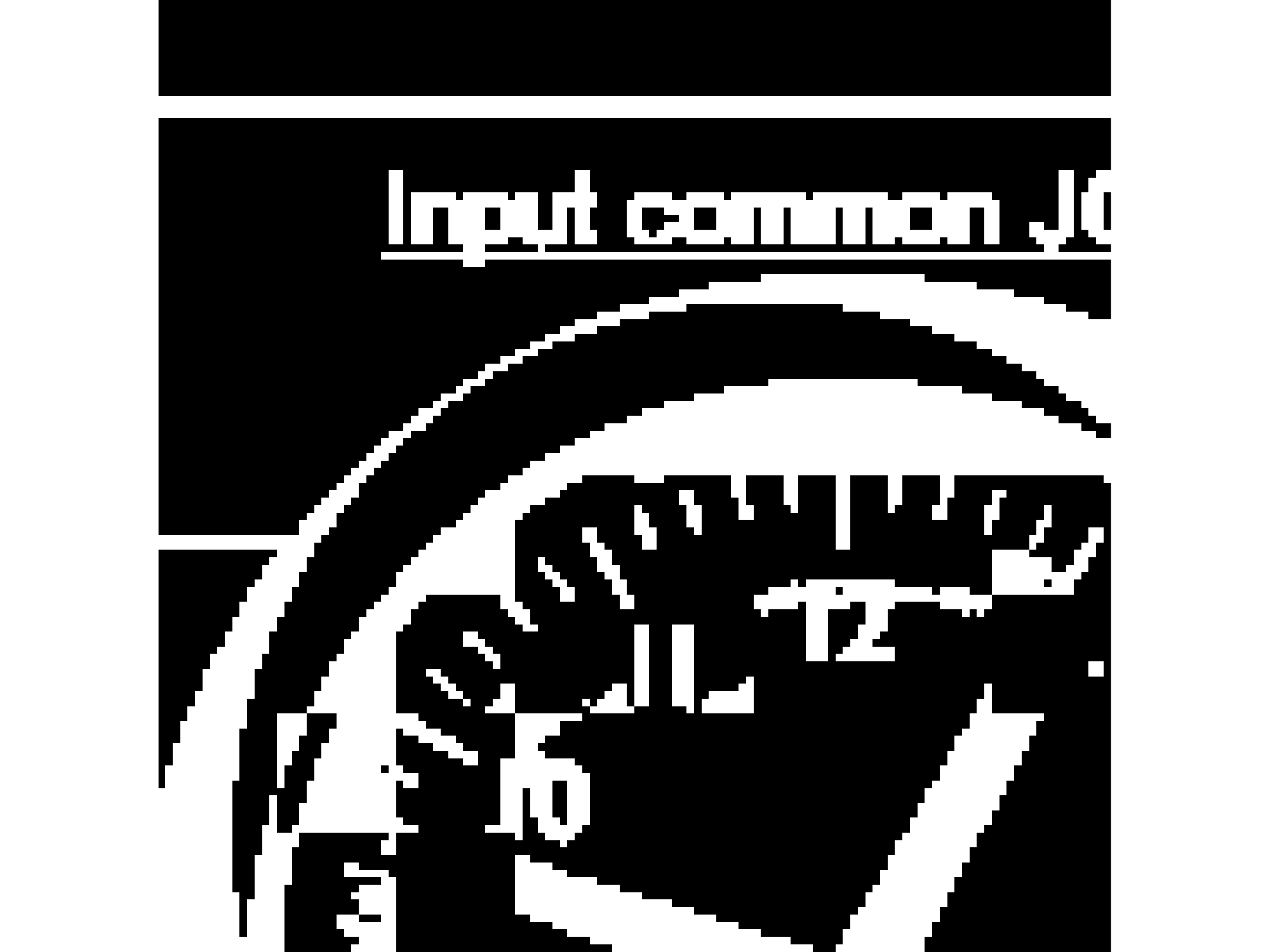}
            \hspace{-3cm} 
        \end{subfigure}%
        ~ 
        \begin{subfigure}[b]{0.22\textwidth}
                \includegraphics[width=\textwidth]{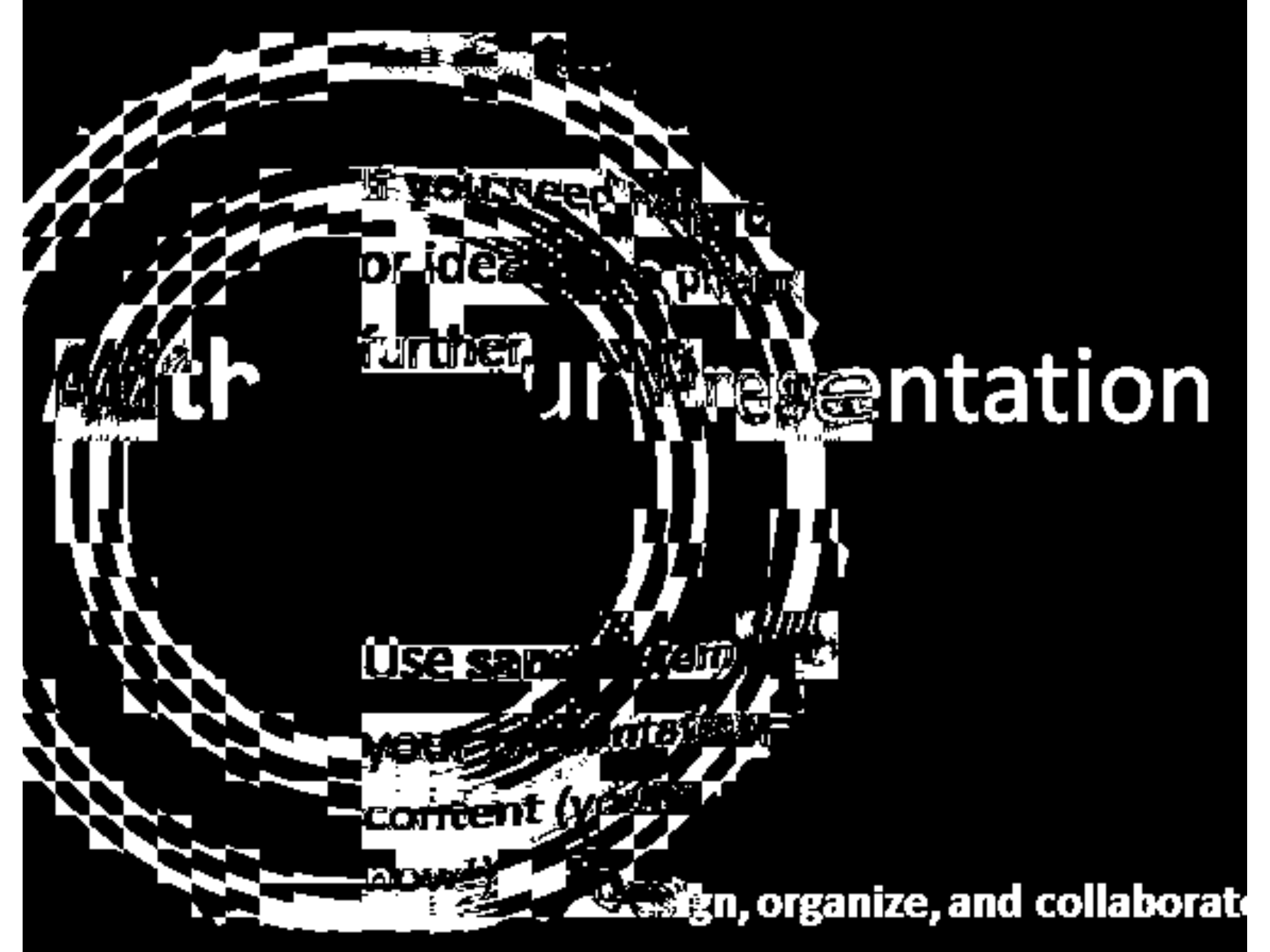}
            \hspace{-3cm} 
        \end{subfigure}%
        \begin{subfigure}[b]{0.22\textwidth}
			~ 
                \includegraphics[width=\textwidth]{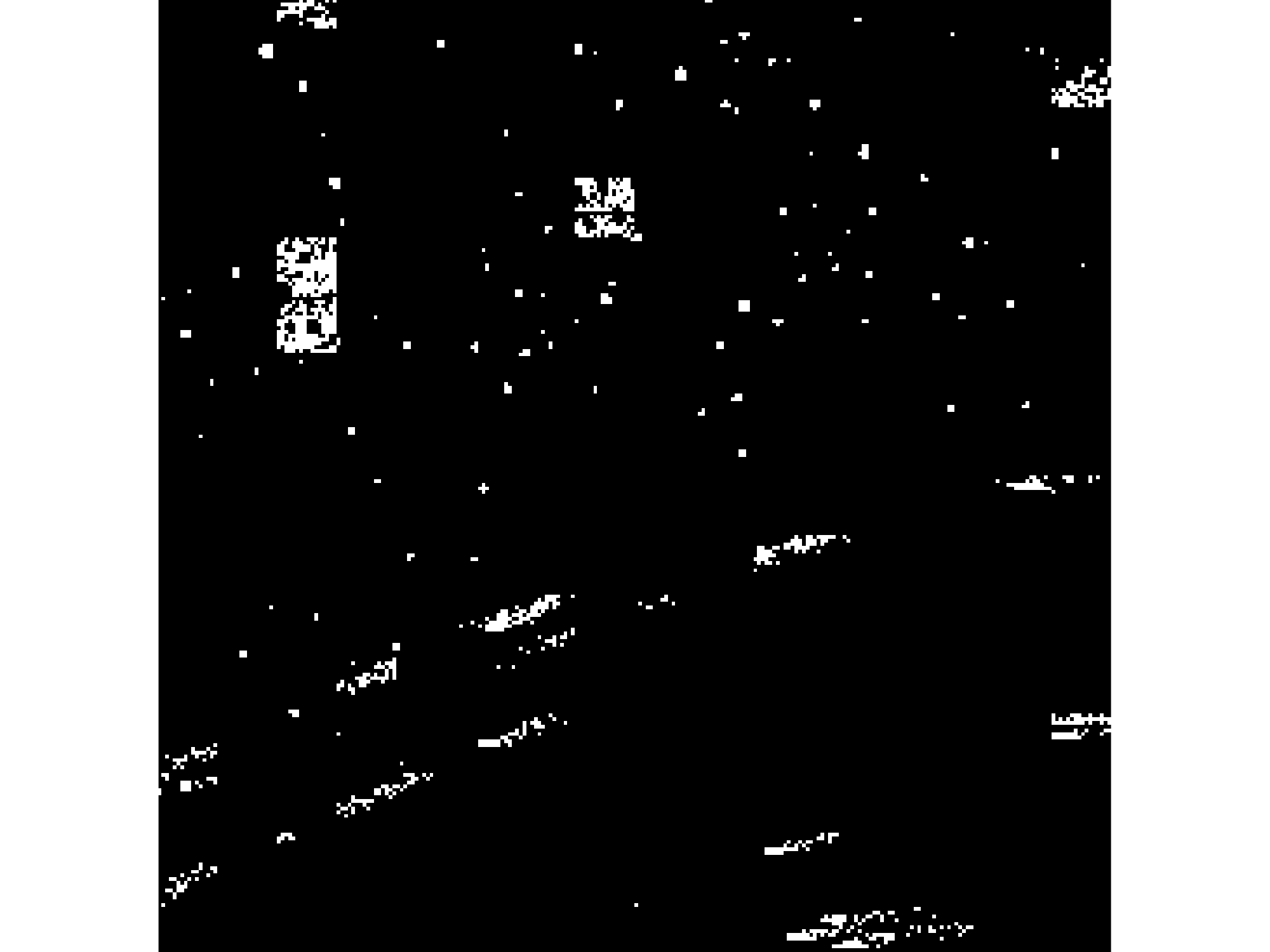}
            \hspace{-3cm} 
        \end{subfigure}%
        \begin{subfigure}[b]{0.22\textwidth}
                \includegraphics[width=\textwidth]{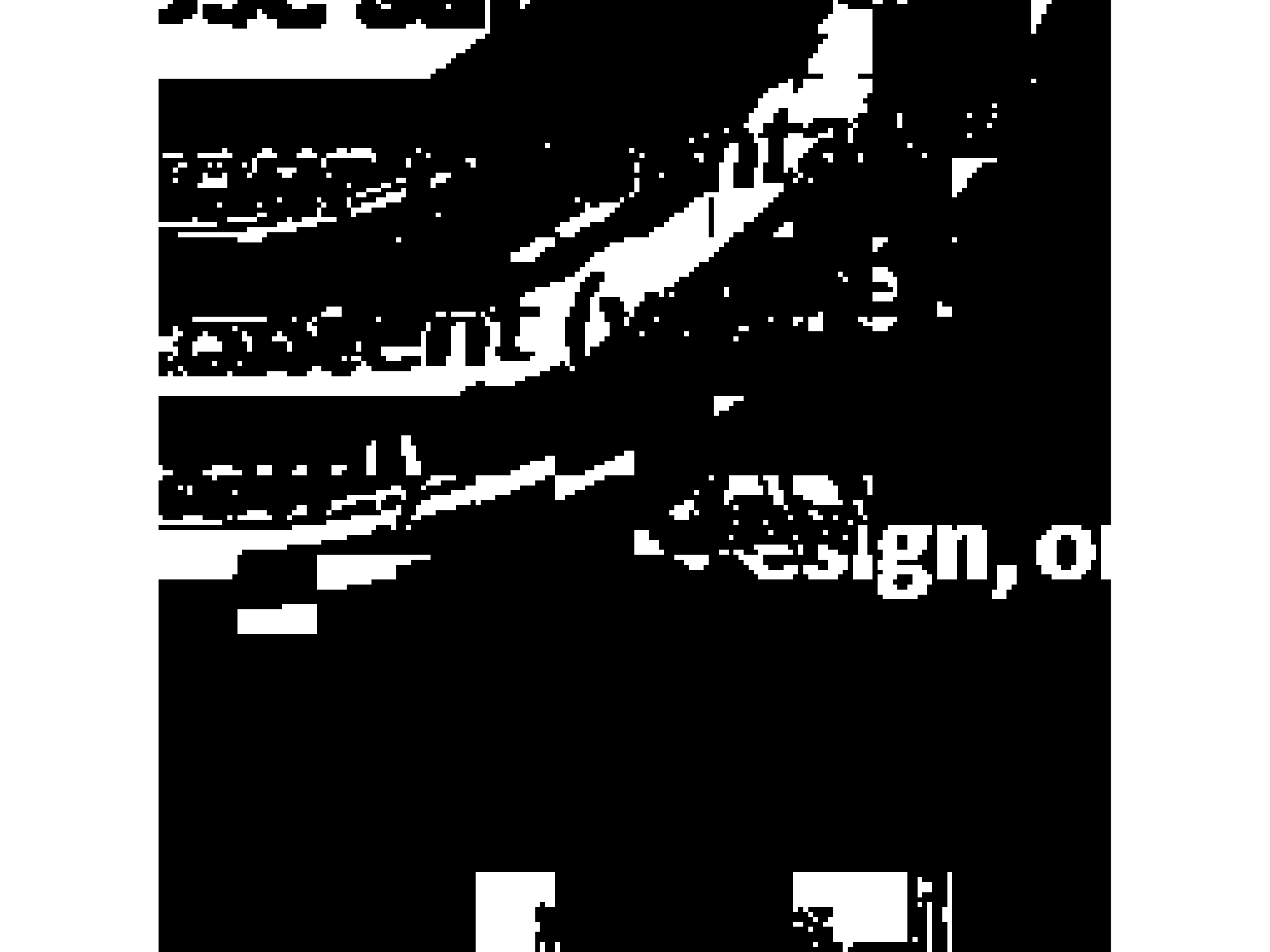}
              \hspace{-4.8cm}
        \end{subfigure} \\[1ex]
        \begin{subfigure}[b]{0.22\textwidth}
                \includegraphics[width=\textwidth]{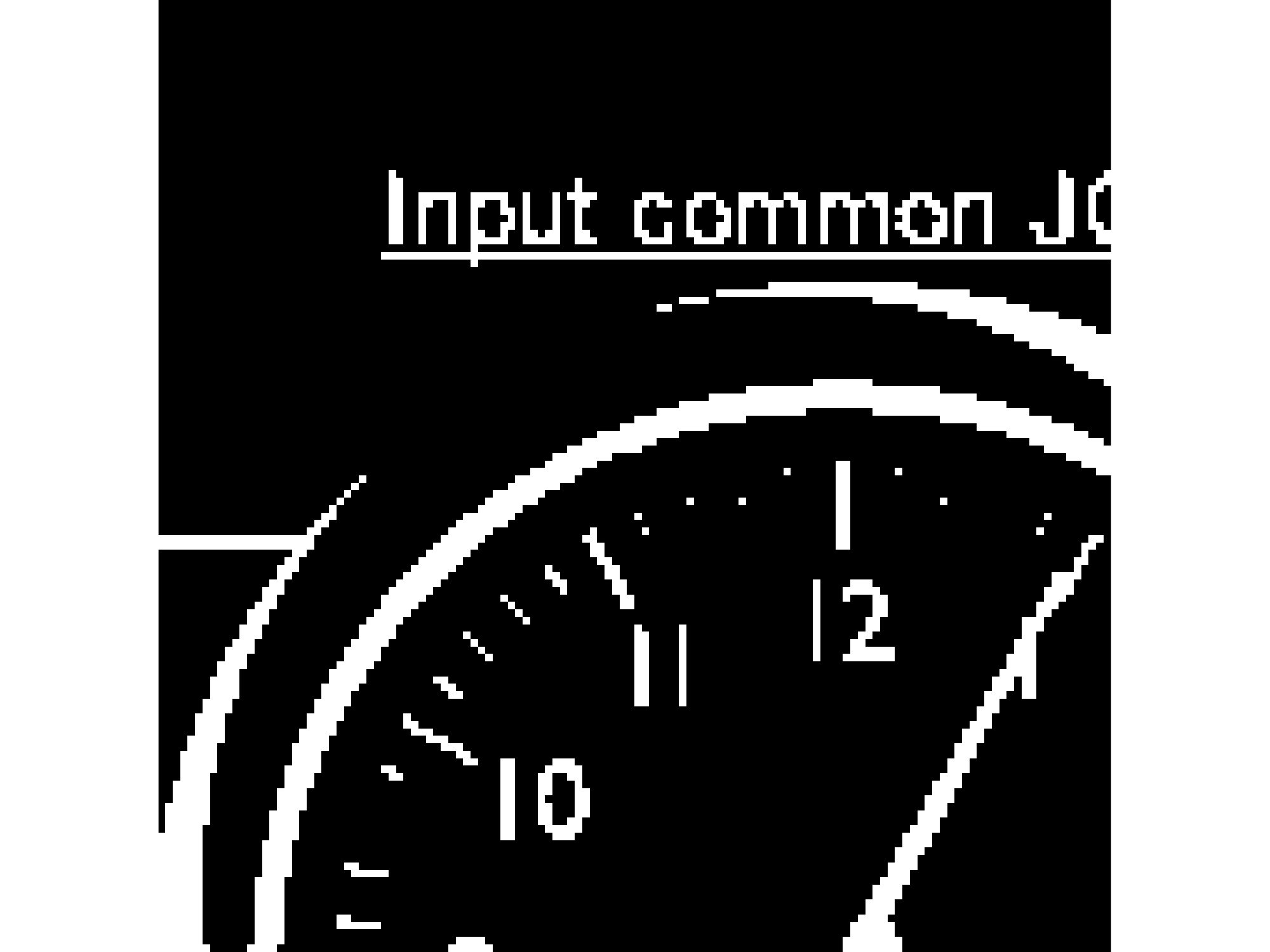}
            \hspace{-3cm} 
        \end{subfigure}%
        ~ 
        \begin{subfigure}[b]{0.22\textwidth}
                \includegraphics[width=\textwidth]{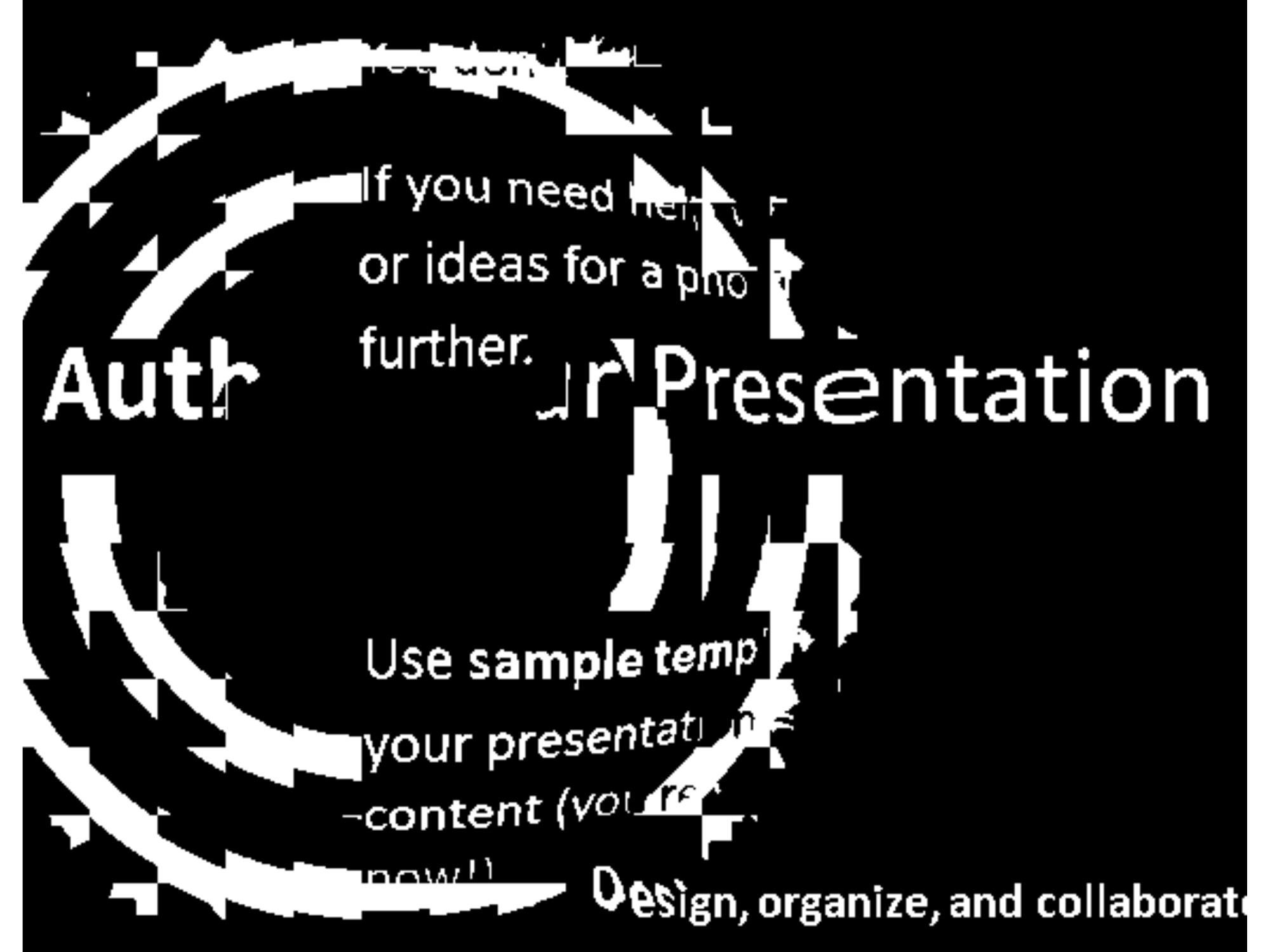}
            \hspace{-3cm} 
        \end{subfigure}%
        \begin{subfigure}[b]{0.22\textwidth}
			~ 
                \includegraphics[width=\textwidth]{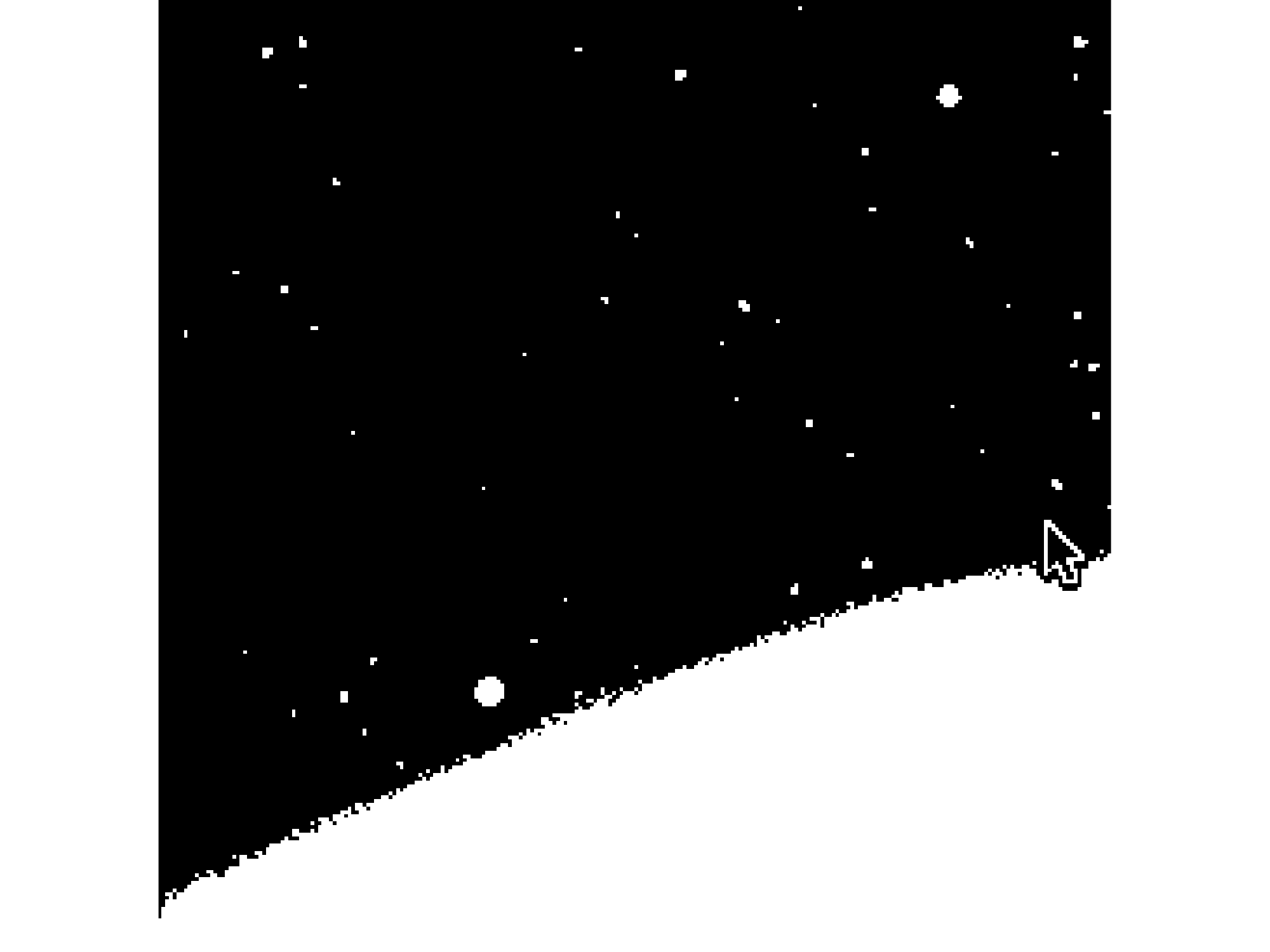}
            \hspace{-3cm} 
        \end{subfigure}%
        \begin{subfigure}[b]{0.22\textwidth}
                \includegraphics[width=\textwidth]{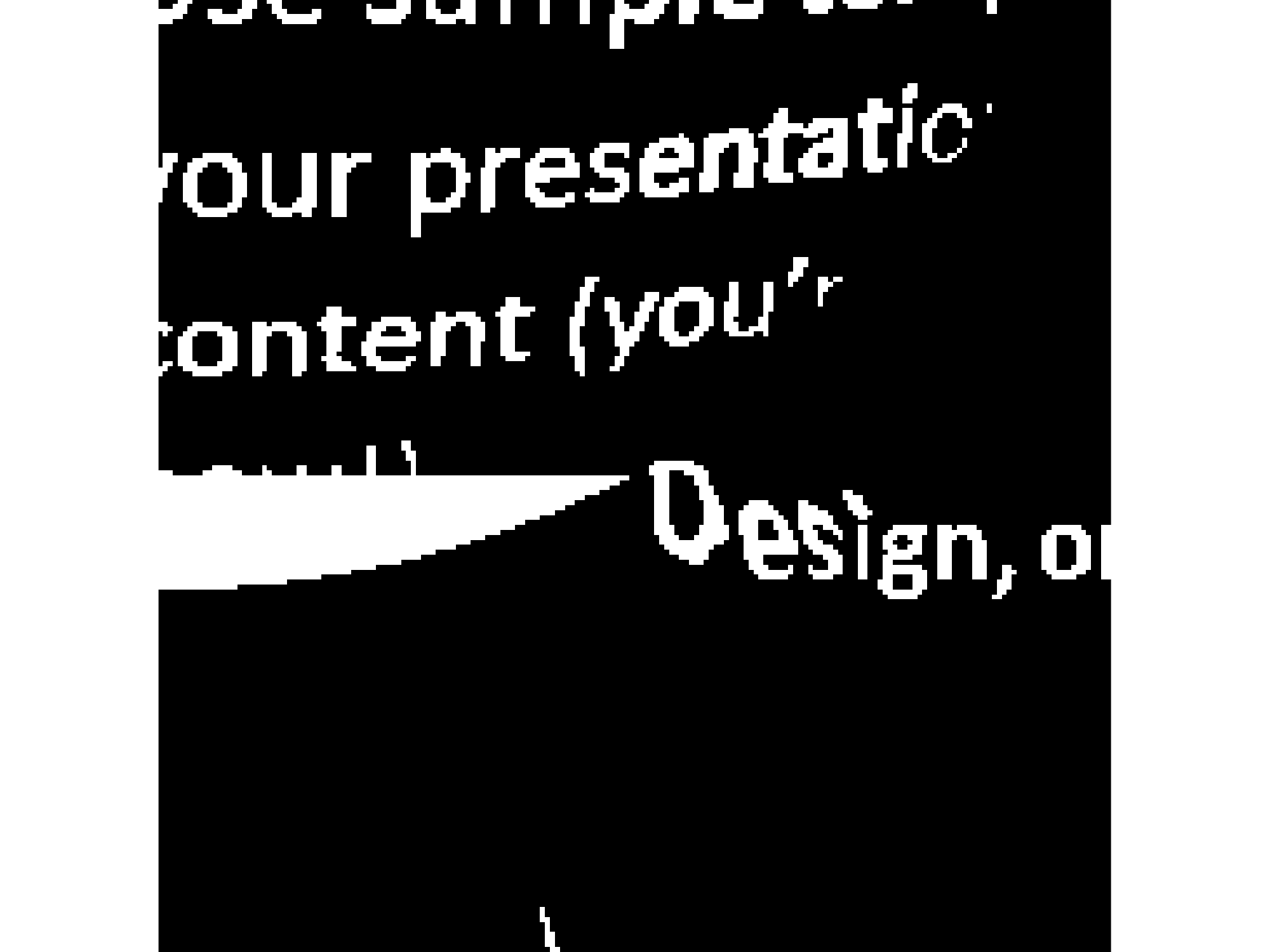}
              \hspace{-4.8cm}
        \end{subfigure}\\[1ex]        
		\begin{subfigure}[b]{0.22\textwidth}
                \includegraphics[width=\textwidth]{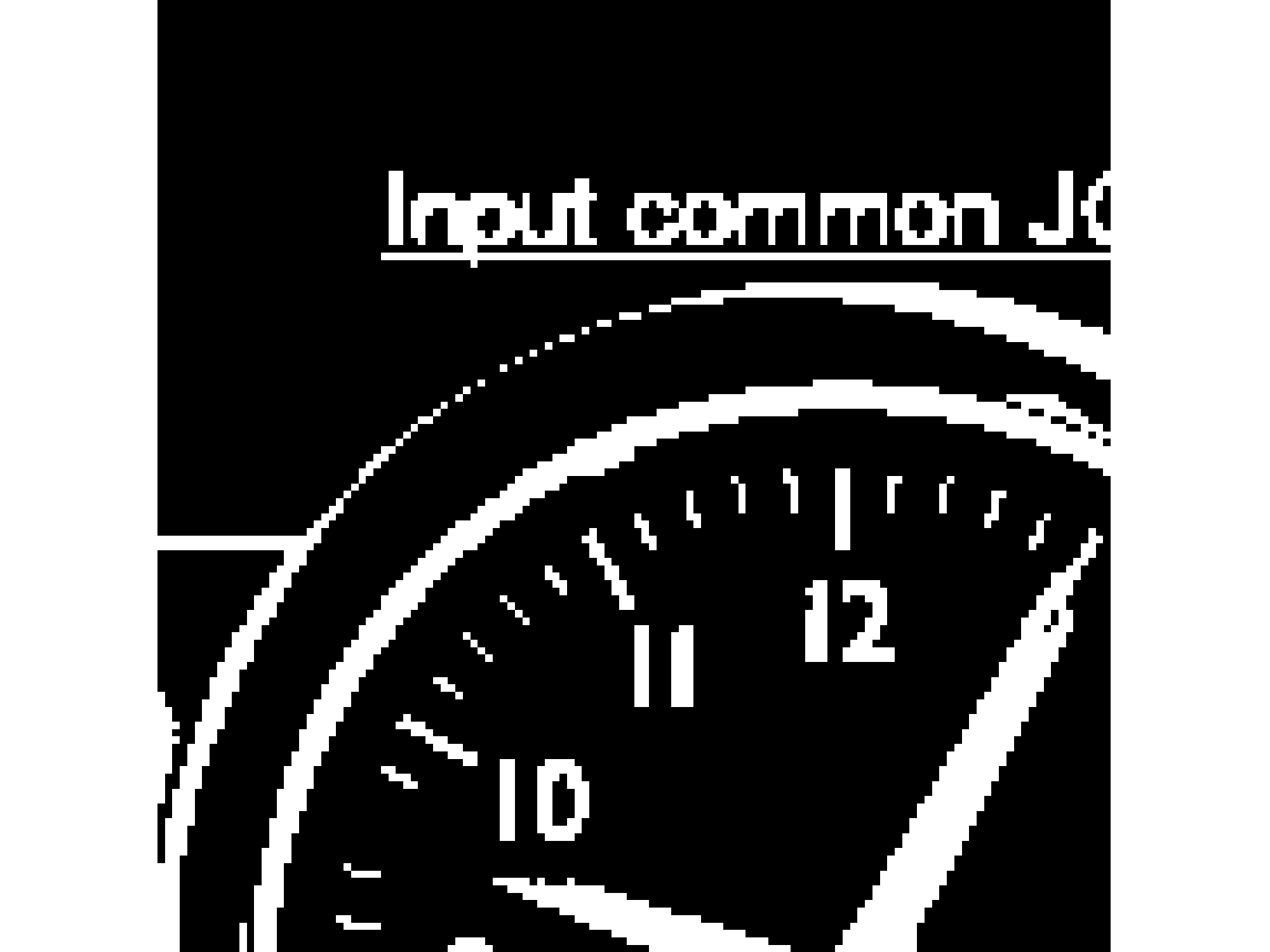}
            \hspace{-3cm} 
        \end{subfigure}%
        ~ 
        \begin{subfigure}[b]{0.22\textwidth}
                \includegraphics[width=\textwidth]{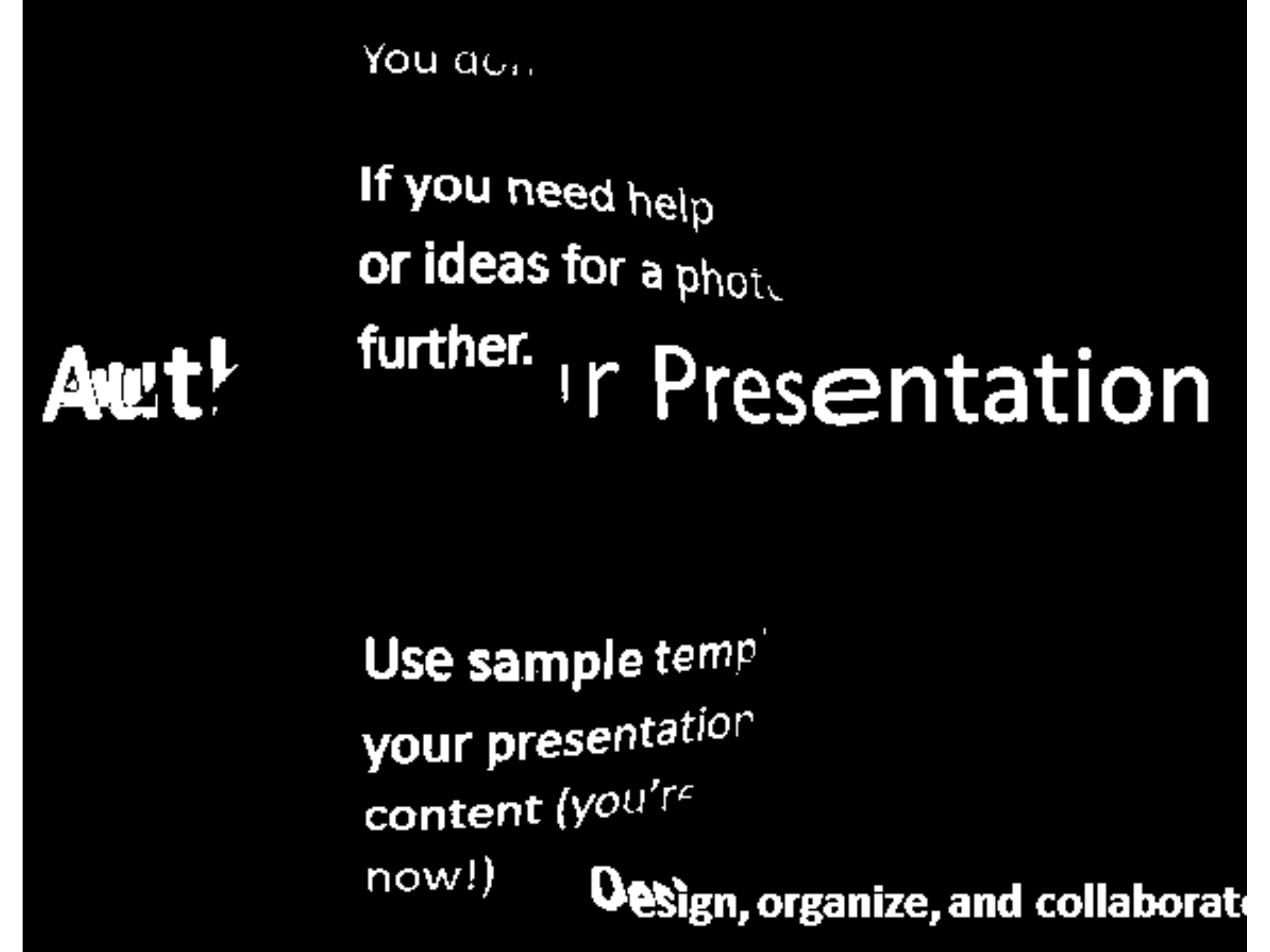}
            \hspace{-3cm} 
        \end{subfigure}%
        \begin{subfigure}[b]{0.22\textwidth}
			~ 
                \includegraphics[width=\textwidth]{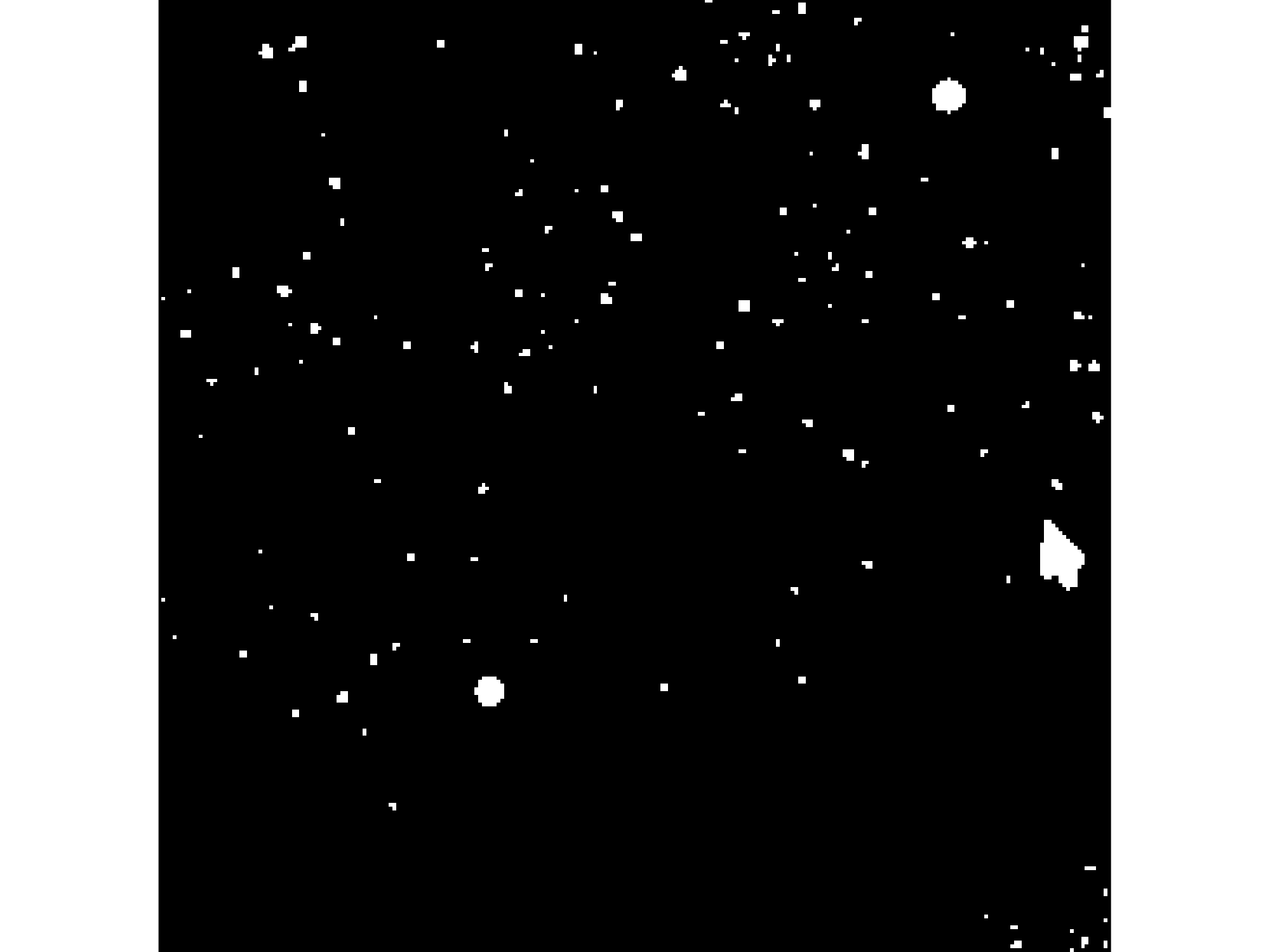}
            \hspace{-3cm} 
        \end{subfigure}%
        \begin{subfigure}[b]{0.22\textwidth}
                \includegraphics[width=\textwidth]{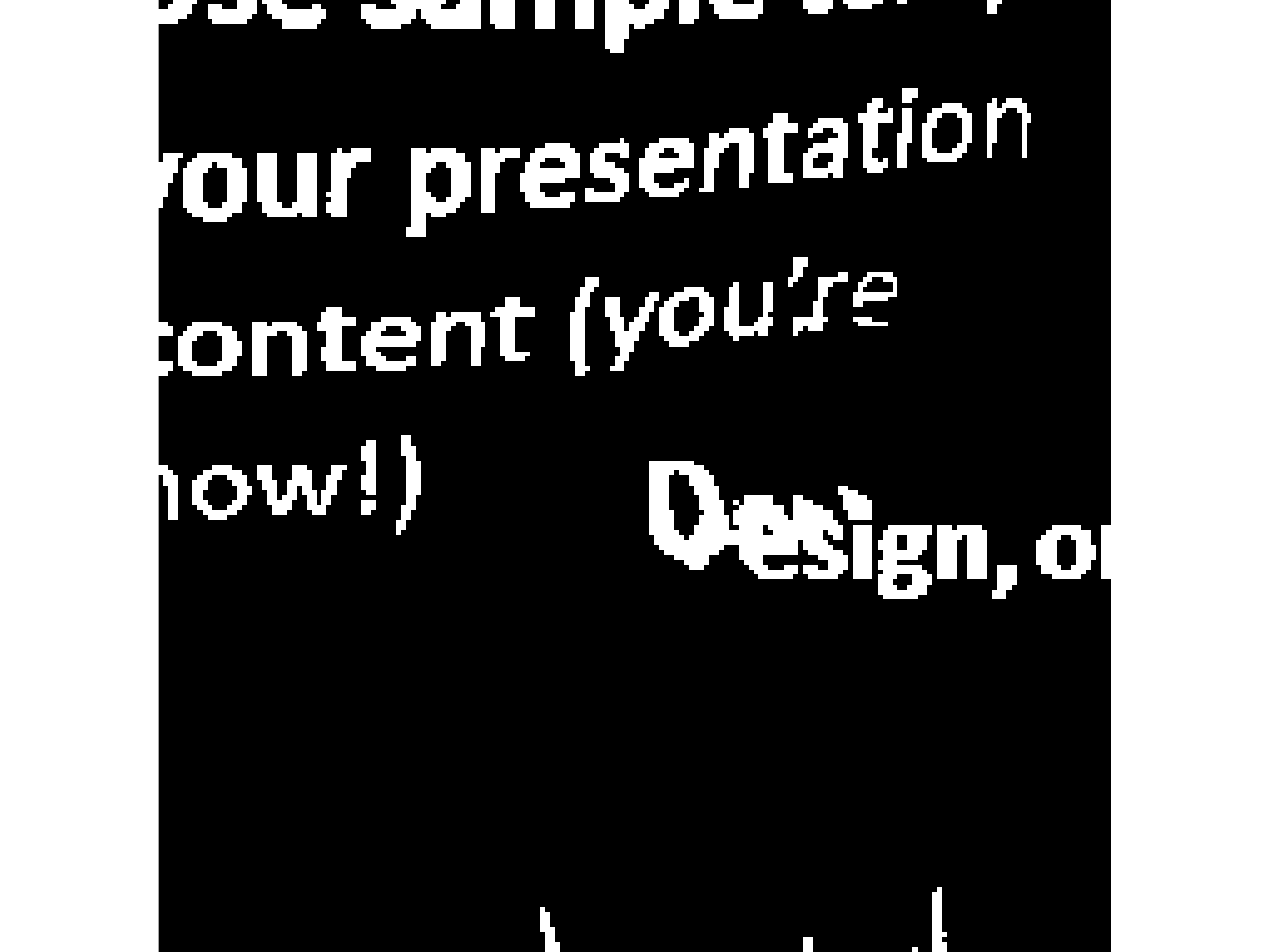}
              \hspace{-4.8cm}
        \end{subfigure} \\[1ex]      
        \begin{subfigure}[b]{0.22\textwidth}
                \includegraphics[width=\textwidth]{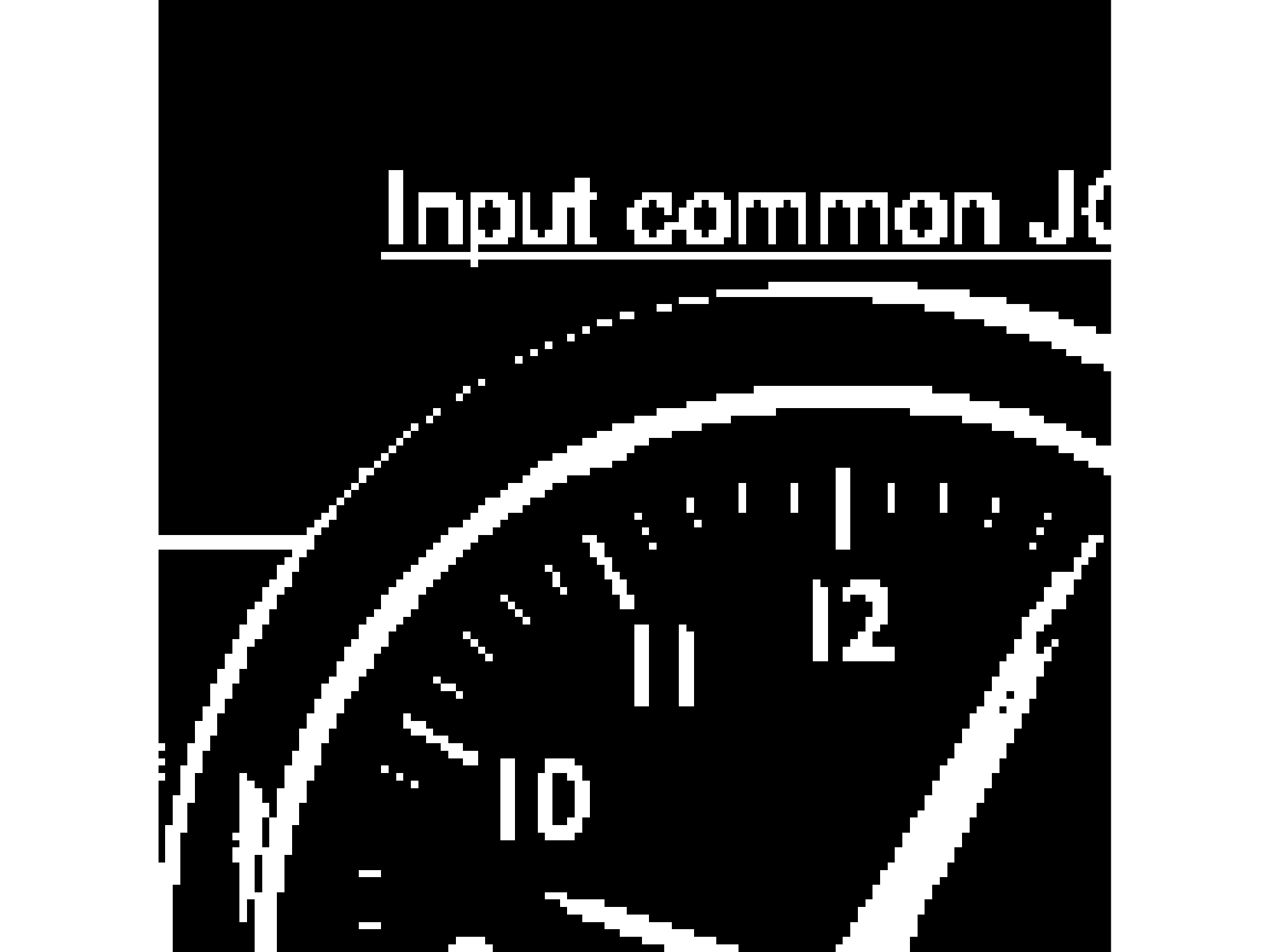}
            \hspace{-3cm} 
        \end{subfigure}%
        ~ 
        \begin{subfigure}[b]{0.22\textwidth}
                \includegraphics[width=\textwidth]{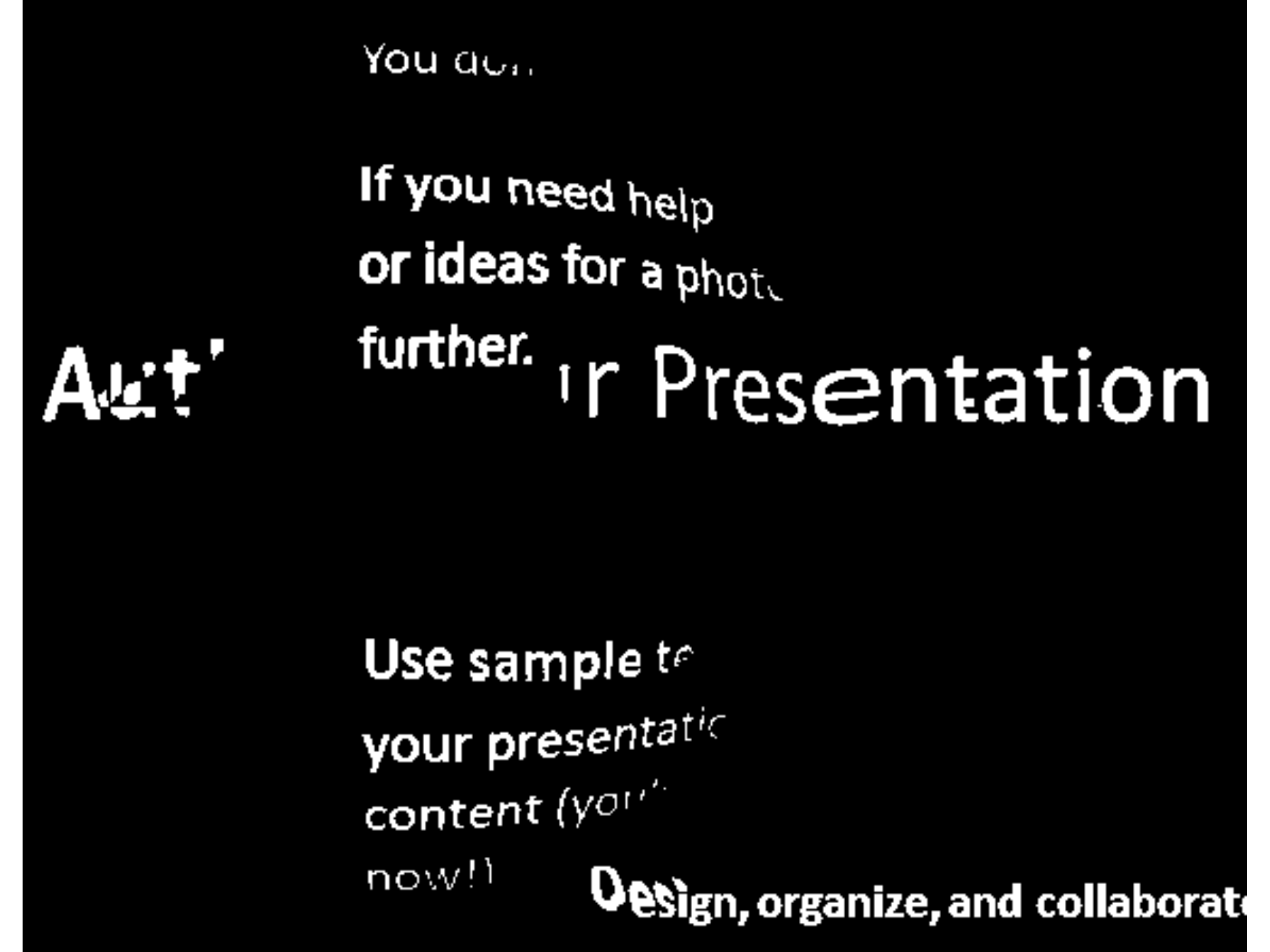}
            \hspace{-3cm} 
        \end{subfigure}%
        \begin{subfigure}[b]{0.22\textwidth}
			~ 
                \includegraphics[width=\textwidth]{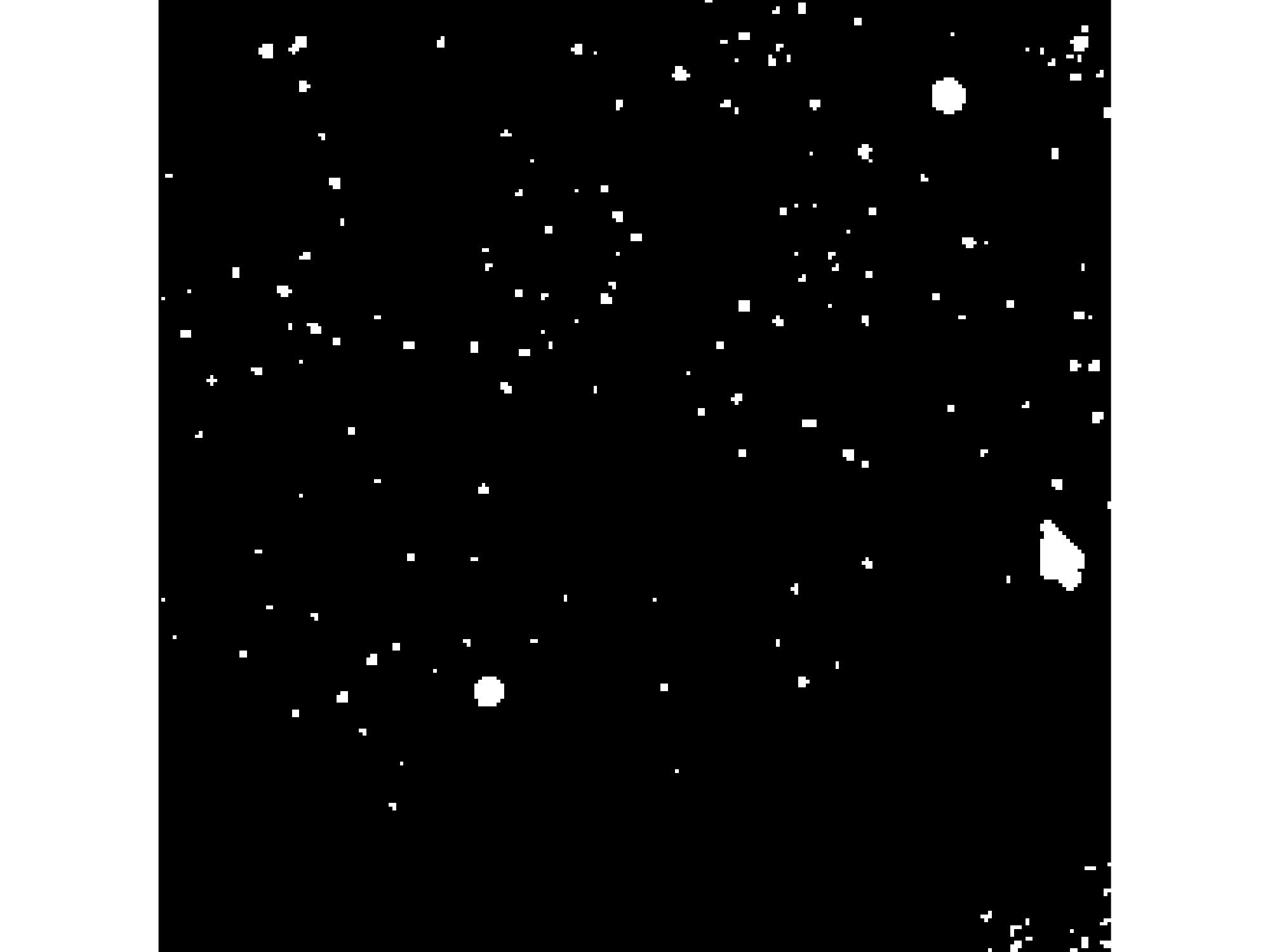} 
            \hspace{-3cm} 
        \end{subfigure}%
        \begin{subfigure}[b]{0.22\textwidth}
                \includegraphics[width=\textwidth]{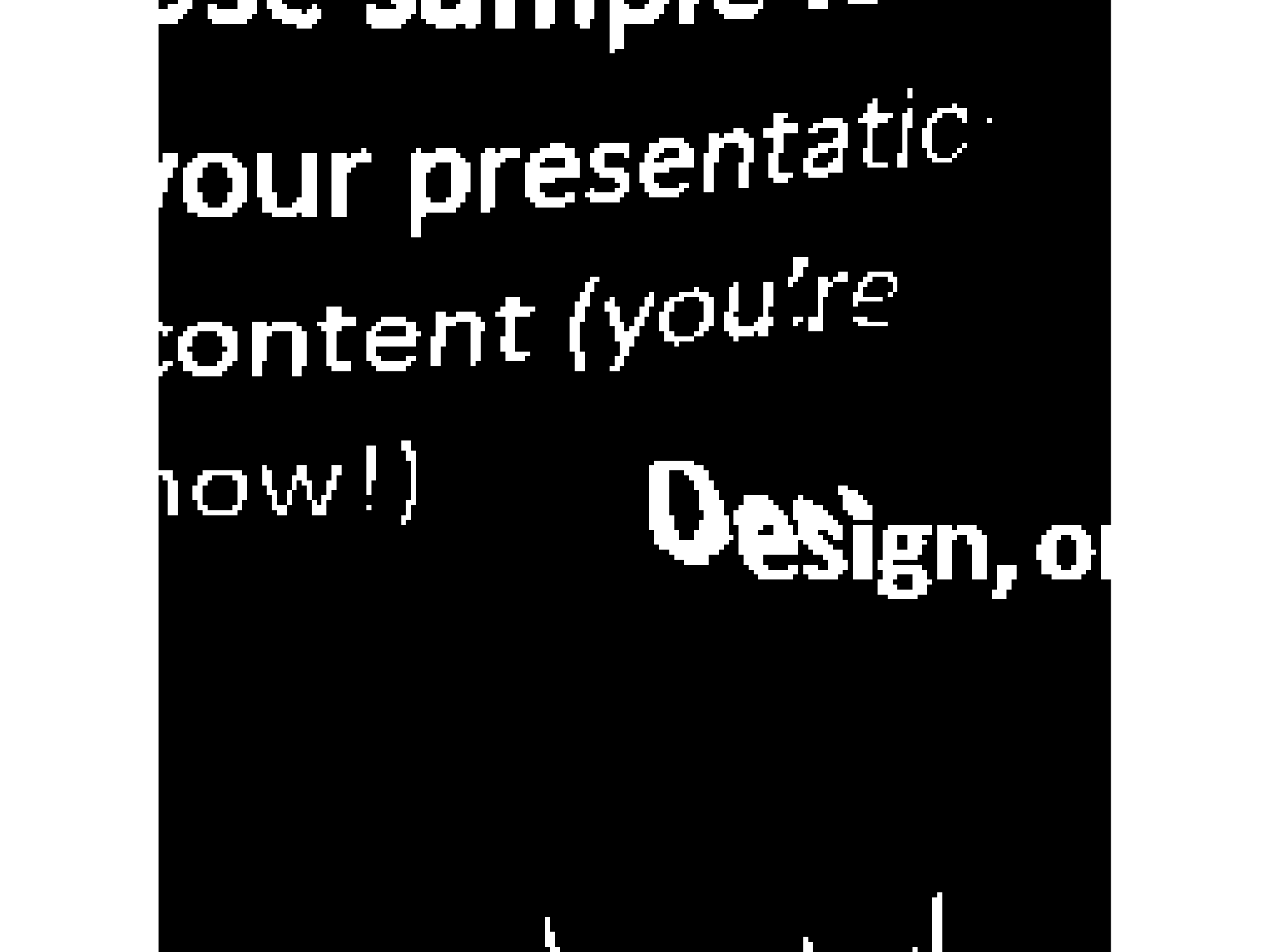}
              \hspace{-4.8cm}
        \end{subfigure}\\[1ex]      
        \begin{subfigure}[b]{0.22\textwidth}
                \includegraphics[width=\textwidth]{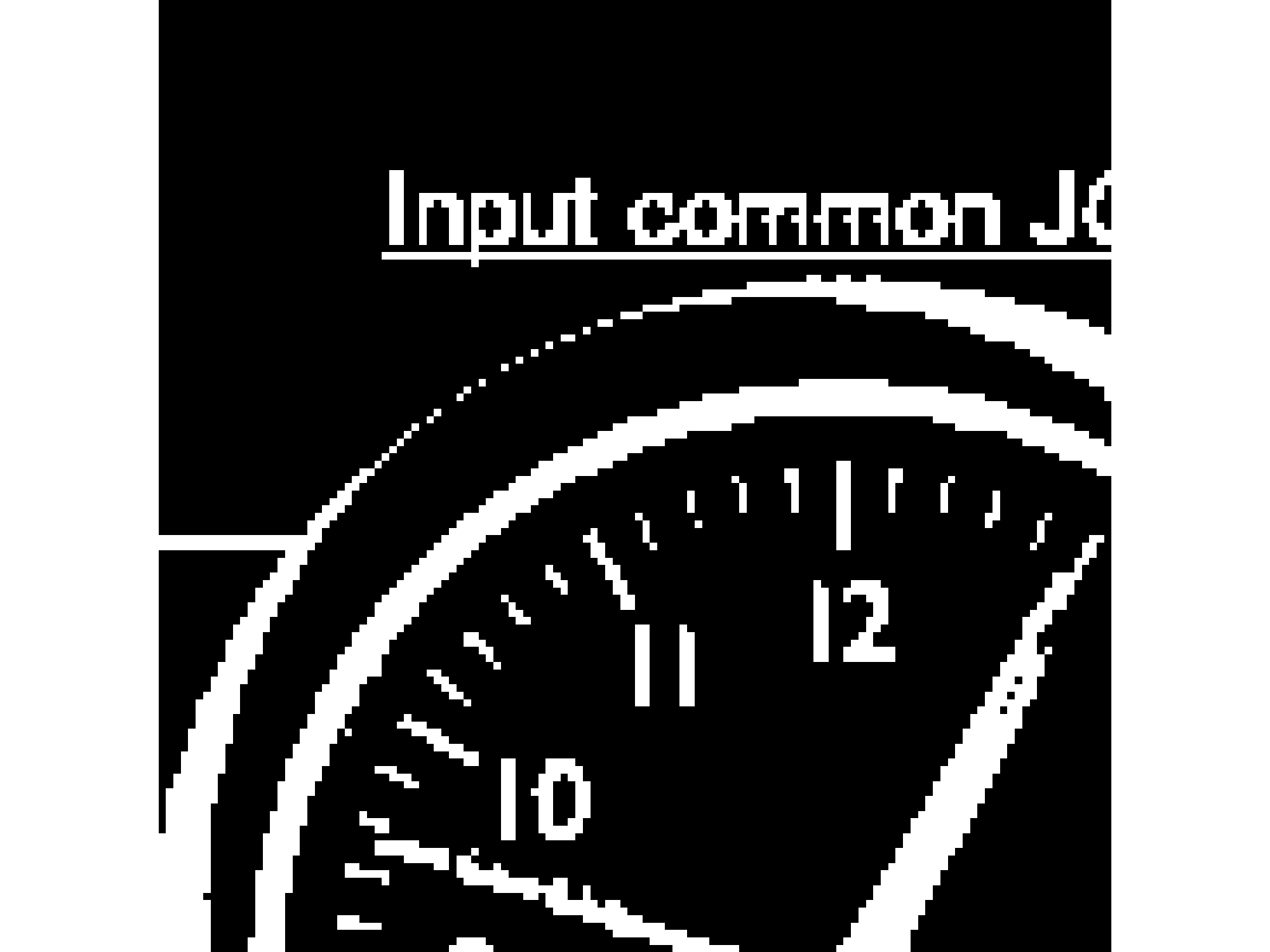}
            \hspace{-3cm} 
        \end{subfigure}%
        ~ 
        \begin{subfigure}[b]{0.22\textwidth}
                \includegraphics[width=\textwidth]{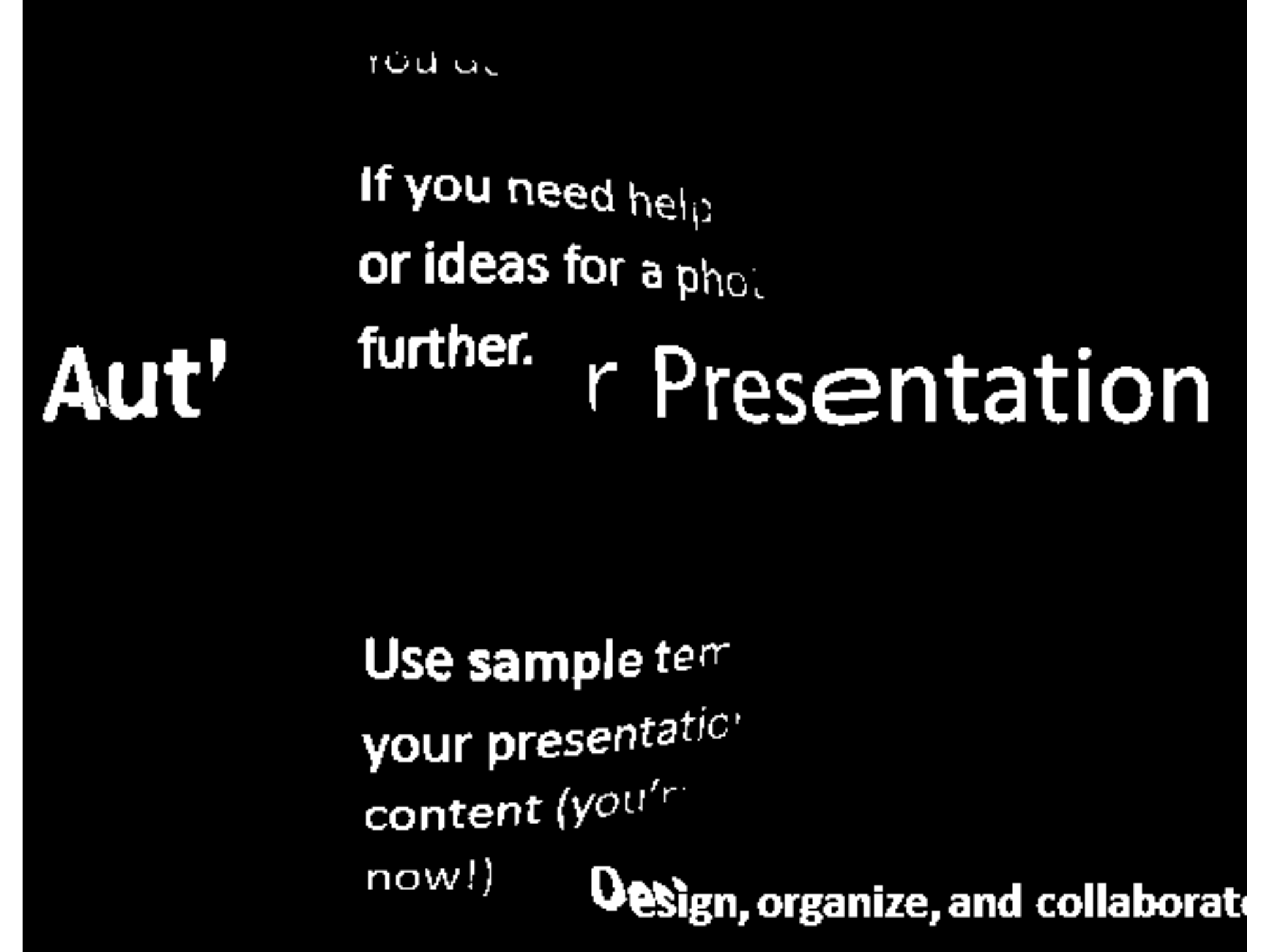}
            \hspace{-3cm} 
        \end{subfigure}%
        \begin{subfigure}[b]{0.22\textwidth}
			~ 
                \includegraphics[width=\textwidth]{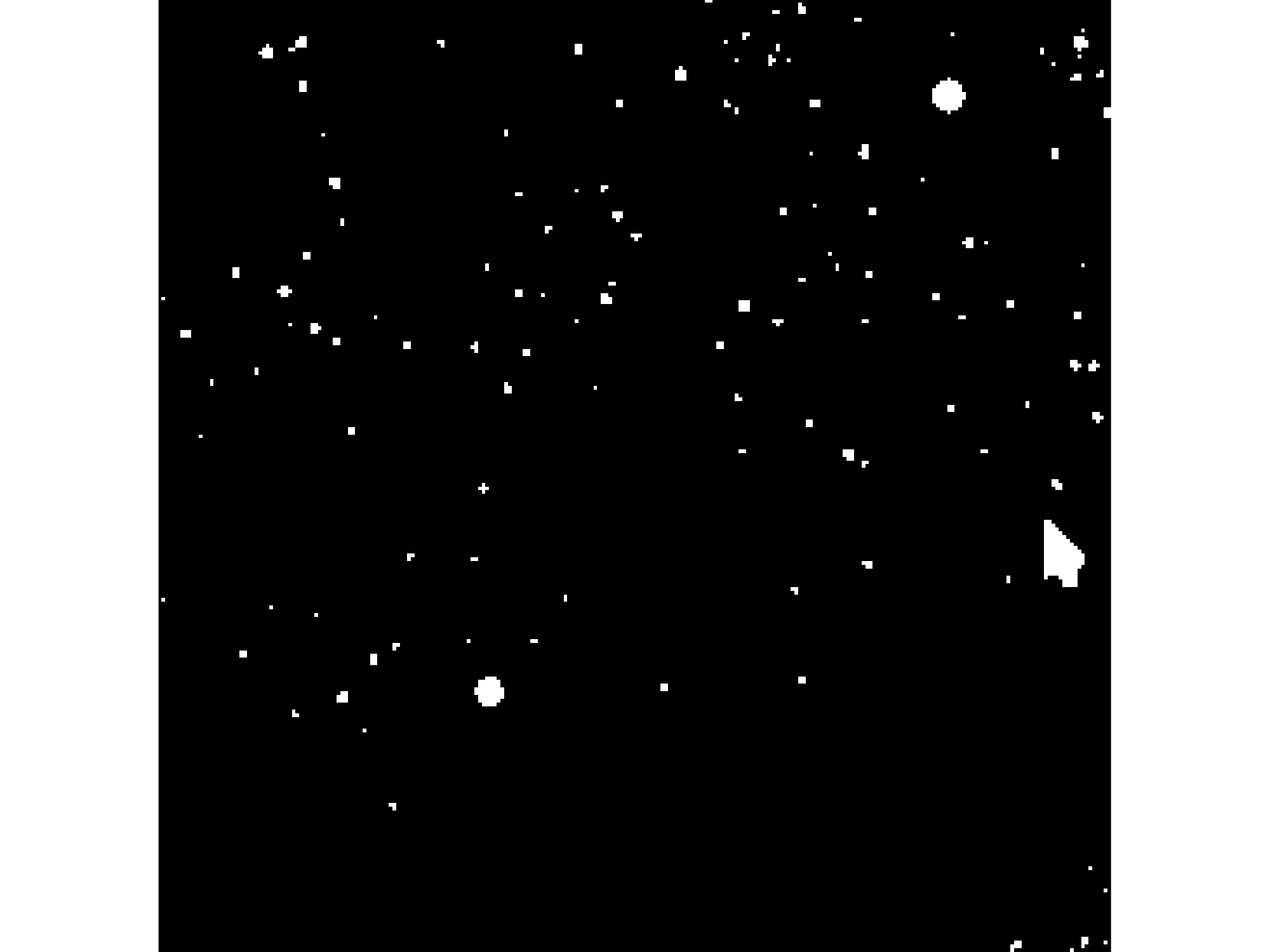} 
            \hspace{-3cm} 
        \end{subfigure}%
        \begin{subfigure}[b]{0.22\textwidth}
                \includegraphics[width=\textwidth]{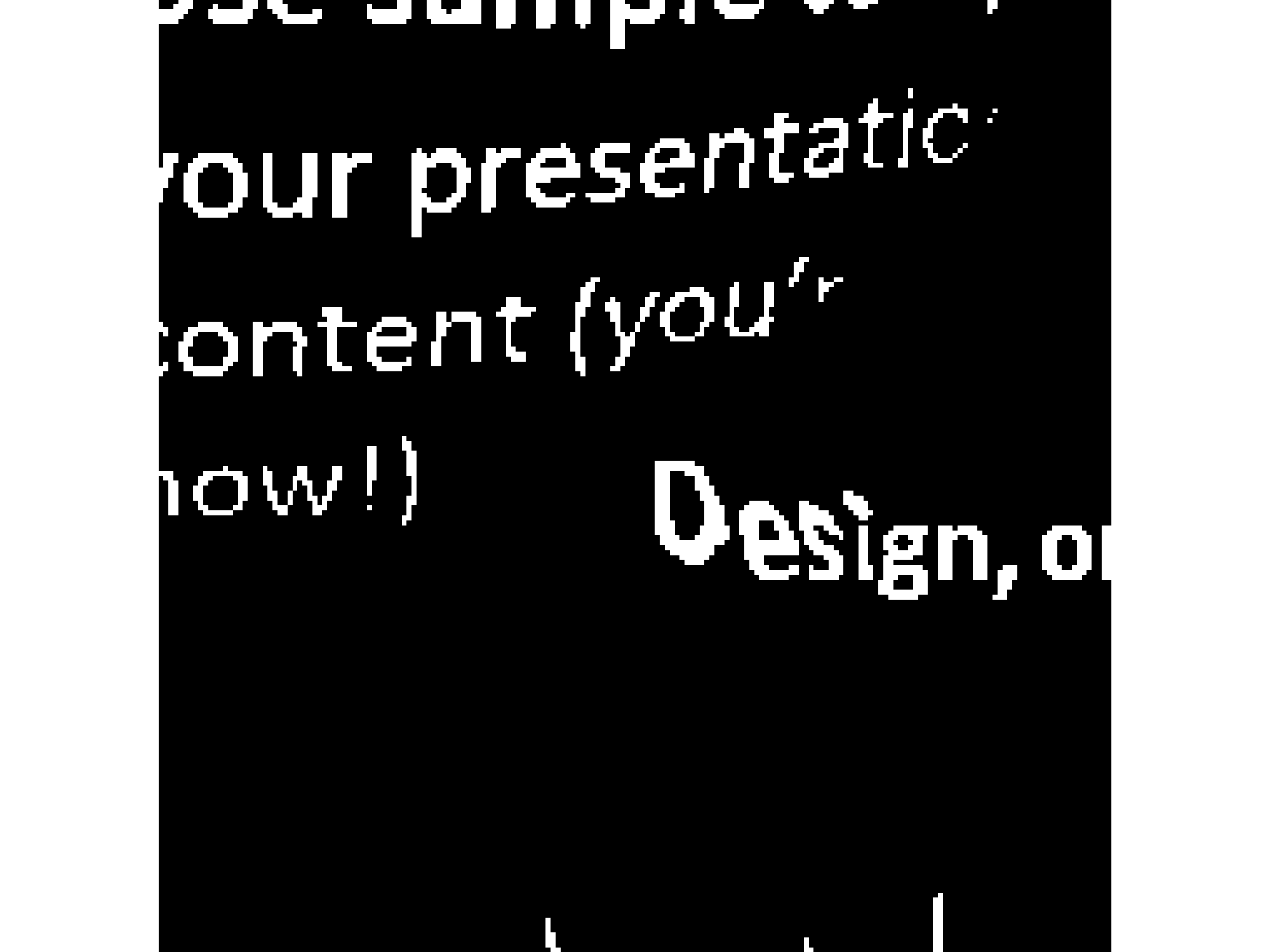}
              \hspace{-4.8cm}
        \end{subfigure}
        \caption{Segmentation result for selected test images for screen content image compression. The images in the first row denotes the original images. And the images in the second, third, forth, fifth and the sixth rows denote the extracted foreground maps by shape primitive extraction and coding, hierarchical k-means clustering, least absolute deviation fitting, sparse decomposition, and the proposed algorithm respectively.}
\end{figure*}

The results for 4 test images (each consisting of multiple 64x64 blocks) are shown in Figure 5. 
It can be seen that the proposed algorithm gives superior performance over DjVu and SPEC in all cases.
There are also noticeable improvement over our prior works on LAD  and sparse decomposition based image segmentation. For example, in the left part of the second image (around the letters AUT), and in the left part of the first image next to the image border, where the LAD algorithm detects some part of background as foreground.
We would like to note that, this dataset mainly consists of challenging images where the background and foreground have overlapping color ranges. For simpler cases where the background has a narrow color range that is quite different from the foreground, both DjVu and least absolute deviation fitting will work well. 


In another experiment, we manually added text on top of an image, and tried to extract them using the proposed algorithm.
Figure 6 shows the comparison between the proposed algorithm and the previous approaches.
For this part we also provide the results derived by the method of sparse and low-rank decomposition \cite{becker1}, using the MATLAB implementation provided in \cite{becker2}.
Essentially this method assumes the background image block is low rank and the text part is sparse.
To derive the foreground map using this approach, we threshold the absolute value of the sparse component after decomposition.
For all images, we see that the proposed method yields significantly better text segmentation. 

\begin{figure*}
        \centering
        \vspace{-0.1cm}
        \begin{subfigure}[b]{0.22\textwidth}
                \includegraphics[width=\textwidth]{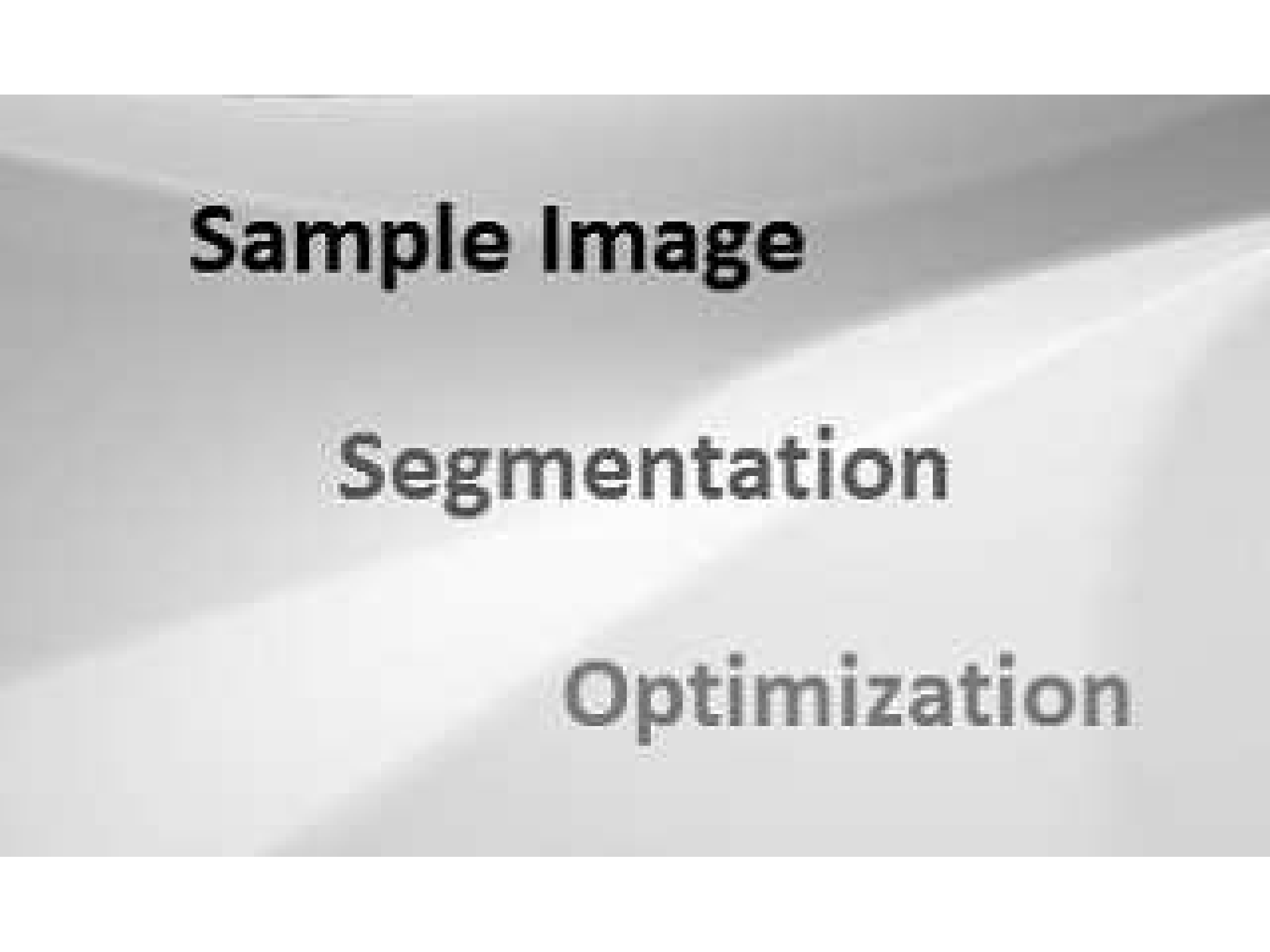}
                                \vspace{-0.3cm}
          \hspace{-1.5cm}    
        \end{subfigure}%
        ~ 
		\vspace{-0.02cm}        
        \begin{subfigure}[b]{0.18\textwidth}
                \includegraphics[width=\textwidth]{texture8_orig-eps-converted-to.pdf}
                \vspace{-0.04cm}
            \hspace{-6cm} 
        \end{subfigure}%
        \begin{subfigure}[b]{0.25\textwidth}
			~ 
                \includegraphics[width=\textwidth]{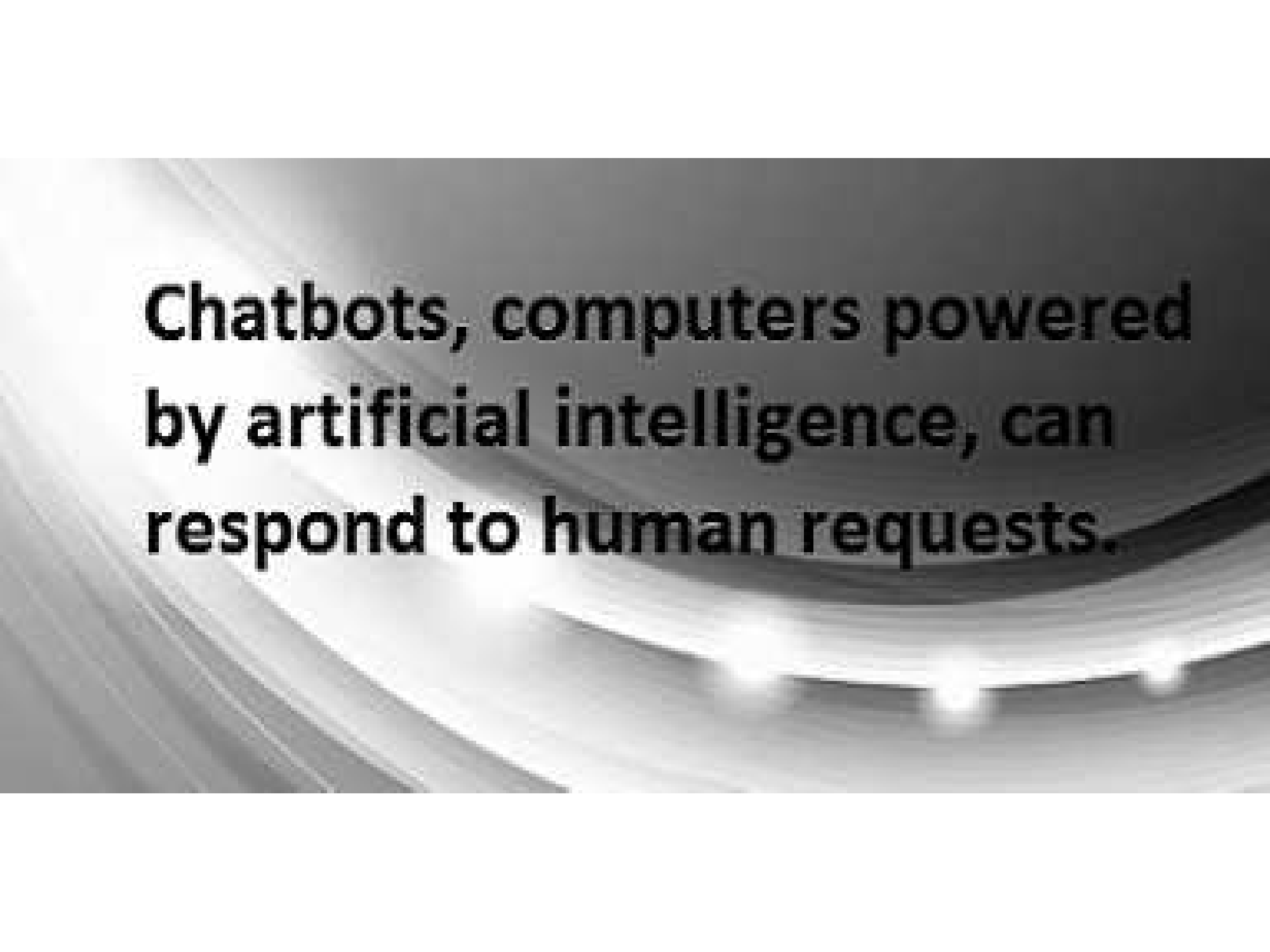}
                \vspace{-0.4cm}
            \hspace{-2cm} 
        \end{subfigure}%
        \begin{subfigure}[b]{0.30\textwidth}
                \includegraphics[width=\textwidth]{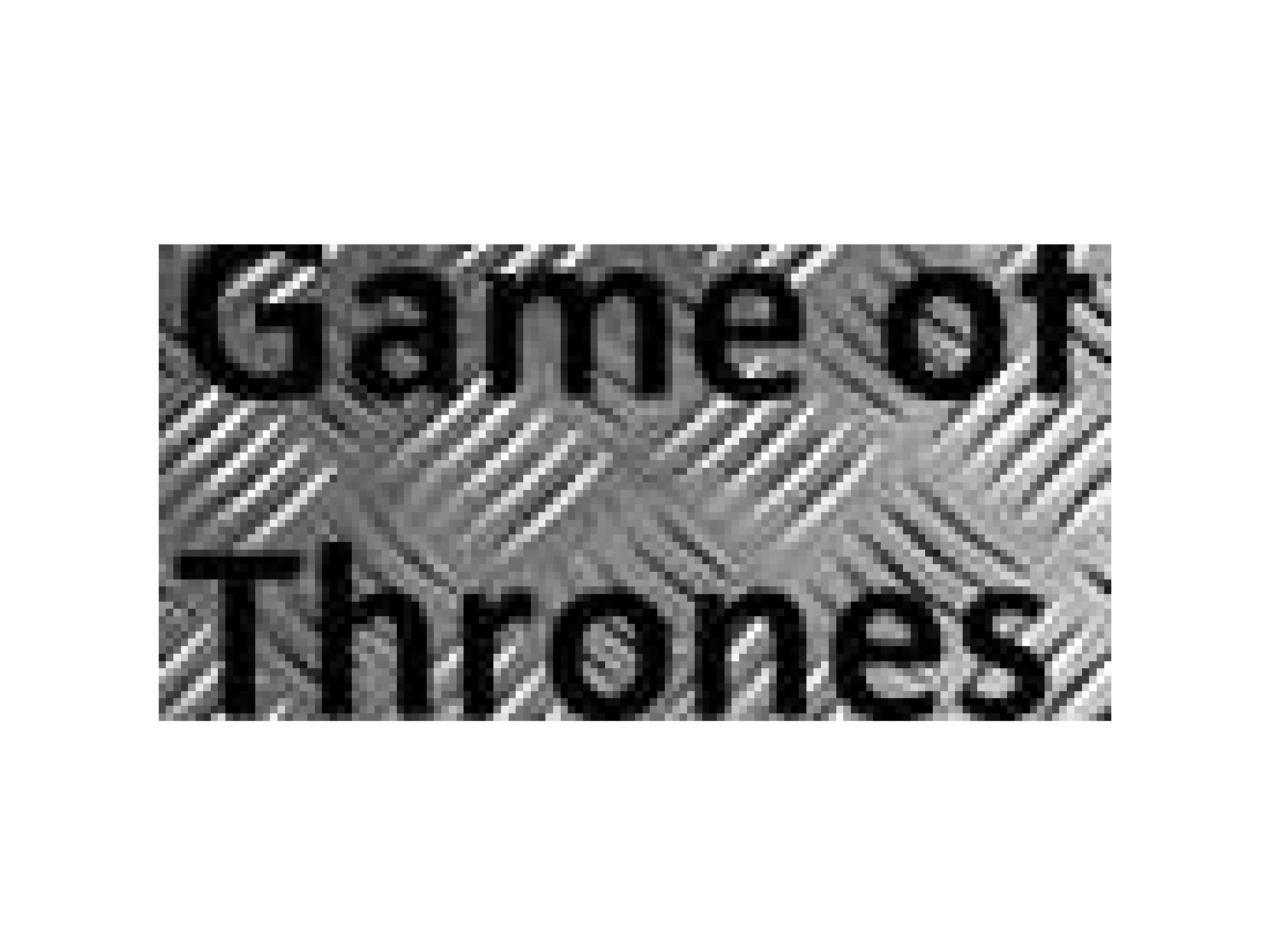}
                 \vspace{-0.61cm}
              \hspace{-4.8cm}
        \end{subfigure}
         \\[1ex]  \vspace{-0.9cm}
        \begin{subfigure}[b]{0.22\textwidth}
                \includegraphics[width=\textwidth]{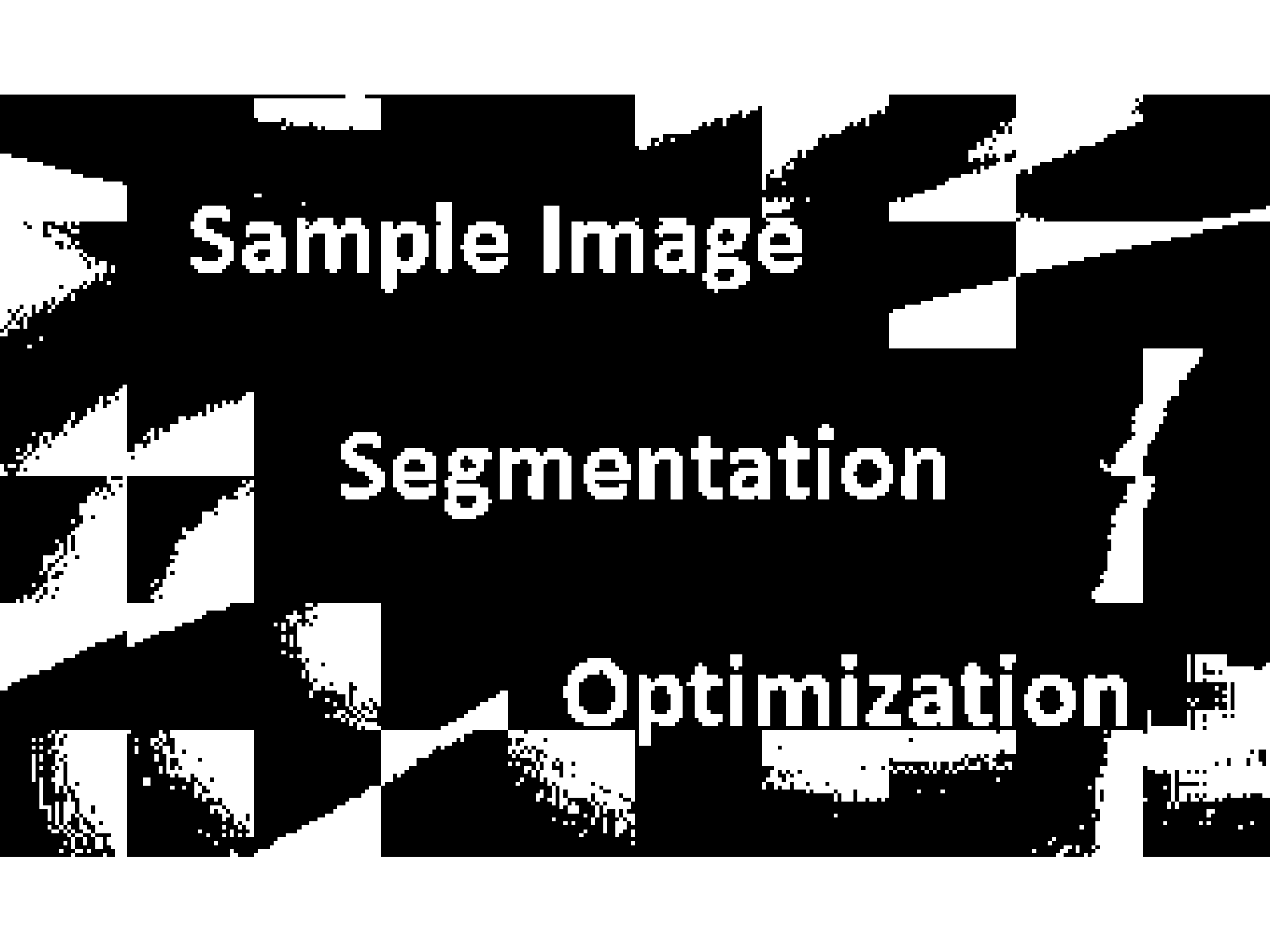}
                                \vspace{-0.32cm}
          \hspace{-1.5cm}    
        \end{subfigure}%
        ~ 
		\vspace{-0.02cm}        
        \begin{subfigure}[b]{0.18\textwidth}
                \includegraphics[width=\textwidth]{textuer8_DjVu-eps-converted-to.pdf}
                \vspace{-0.04cm}
            \hspace{-6cm} 
        \end{subfigure}%
        \begin{subfigure}[b]{0.25\textwidth}
			~ 
                \includegraphics[width=\textwidth]{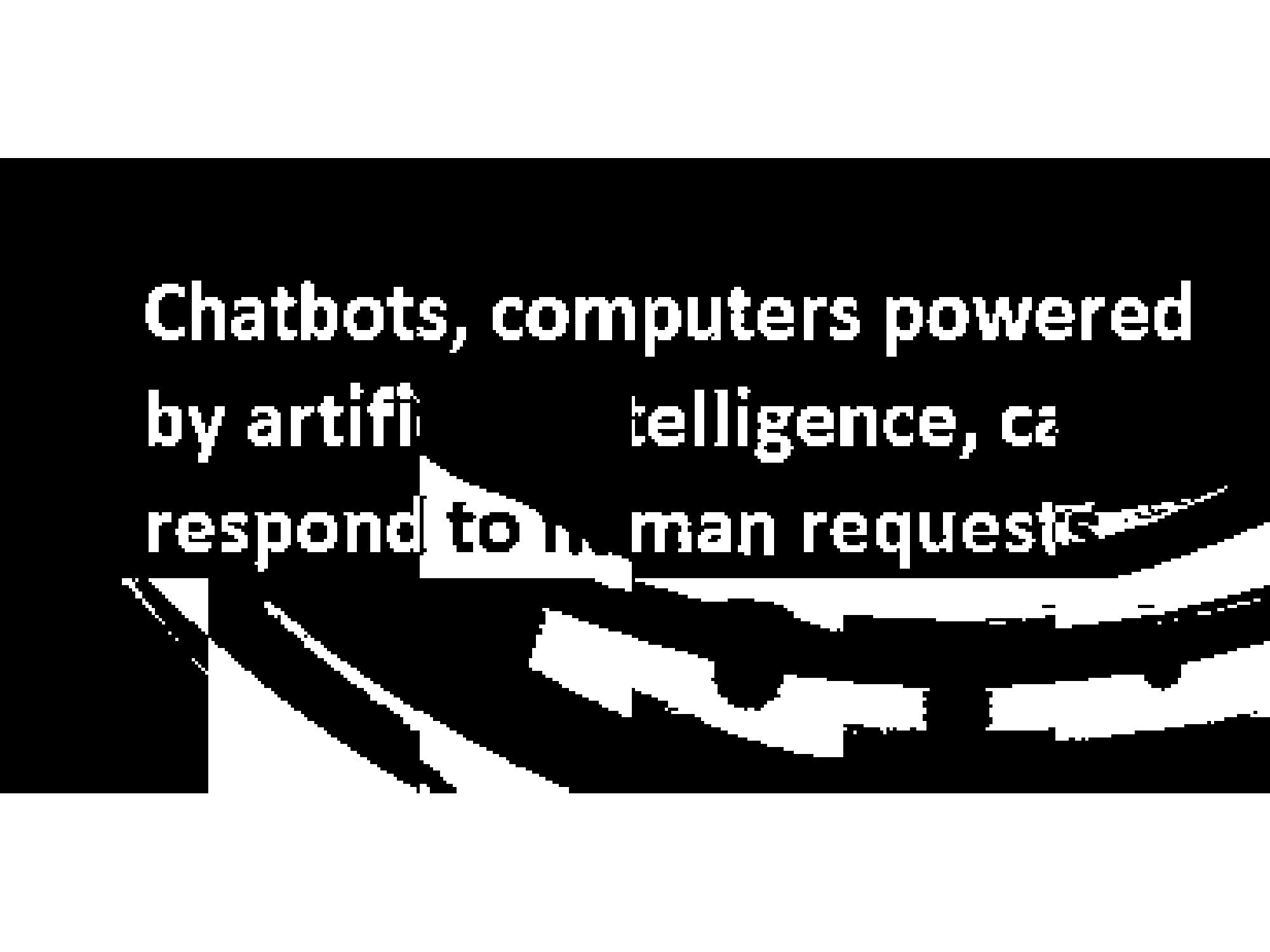}
                \vspace{-0.4cm}
            \hspace{-2cm} 
        \end{subfigure}%
        \begin{subfigure}[b]{0.30\textwidth}
                \includegraphics[width=\textwidth]{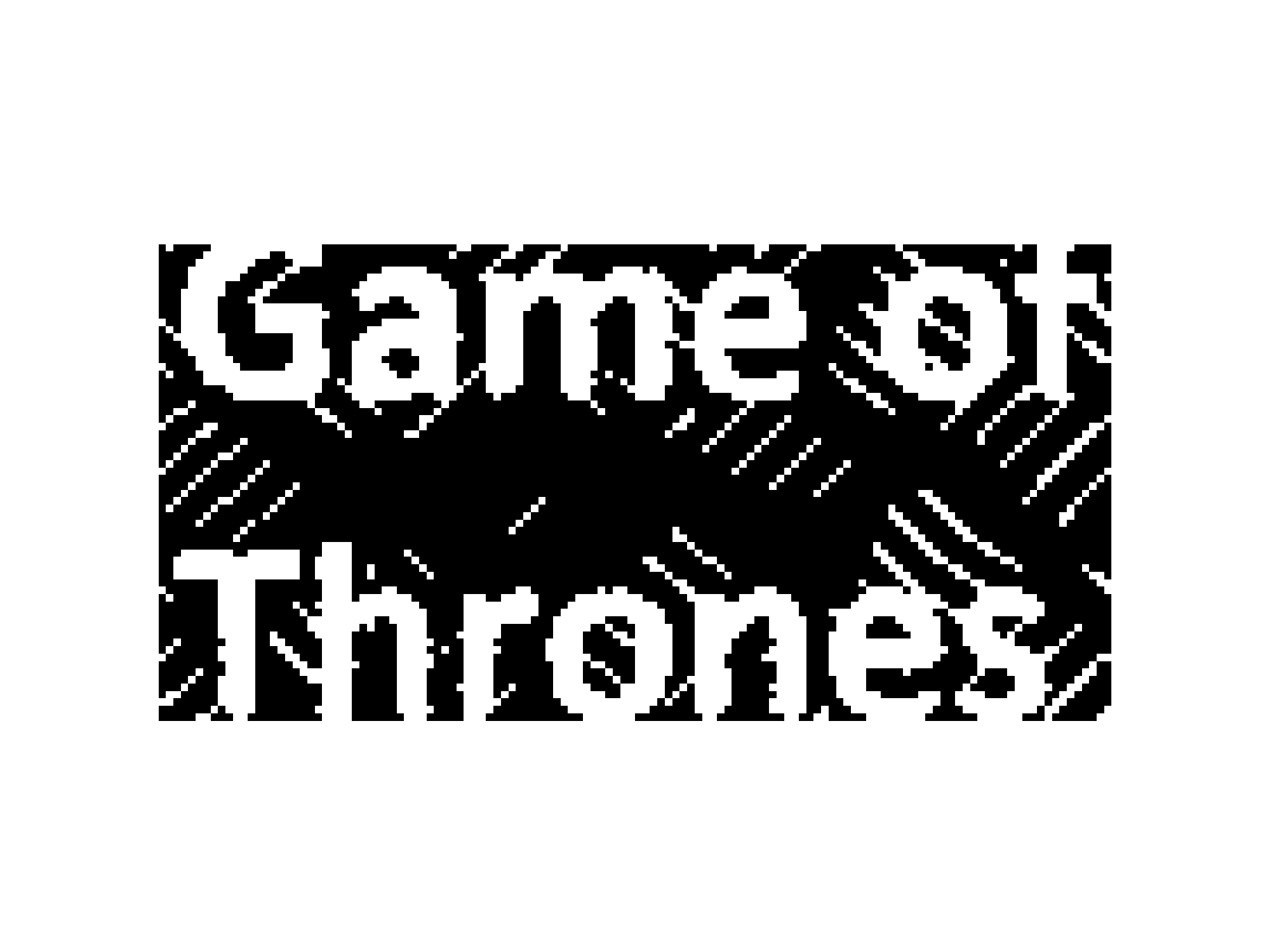}
                 \vspace{-0.61cm}
              \hspace{-4.8cm}
        \end{subfigure}
         \\[1ex]  \vspace{-0.9cm}
        \begin{subfigure}[b]{0.22\textwidth}
                \includegraphics[width=\textwidth]{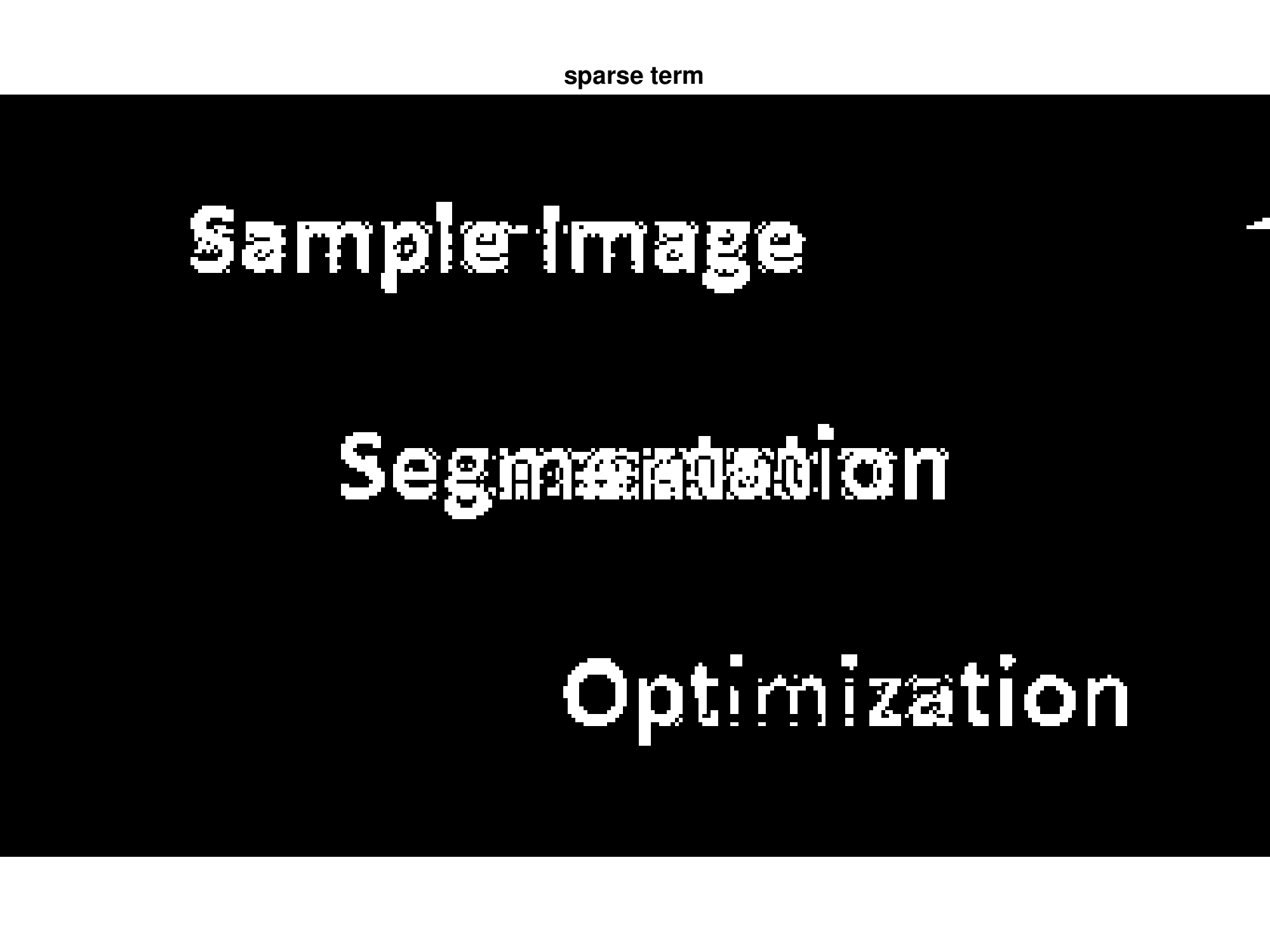}
                                \vspace{-0.32cm}
          \hspace{-1.5cm}    
        \end{subfigure}%
        ~ 
		\vspace{-0.02cm}        
        \begin{subfigure}[b]{0.18\textwidth}
                \includegraphics[width=\textwidth]{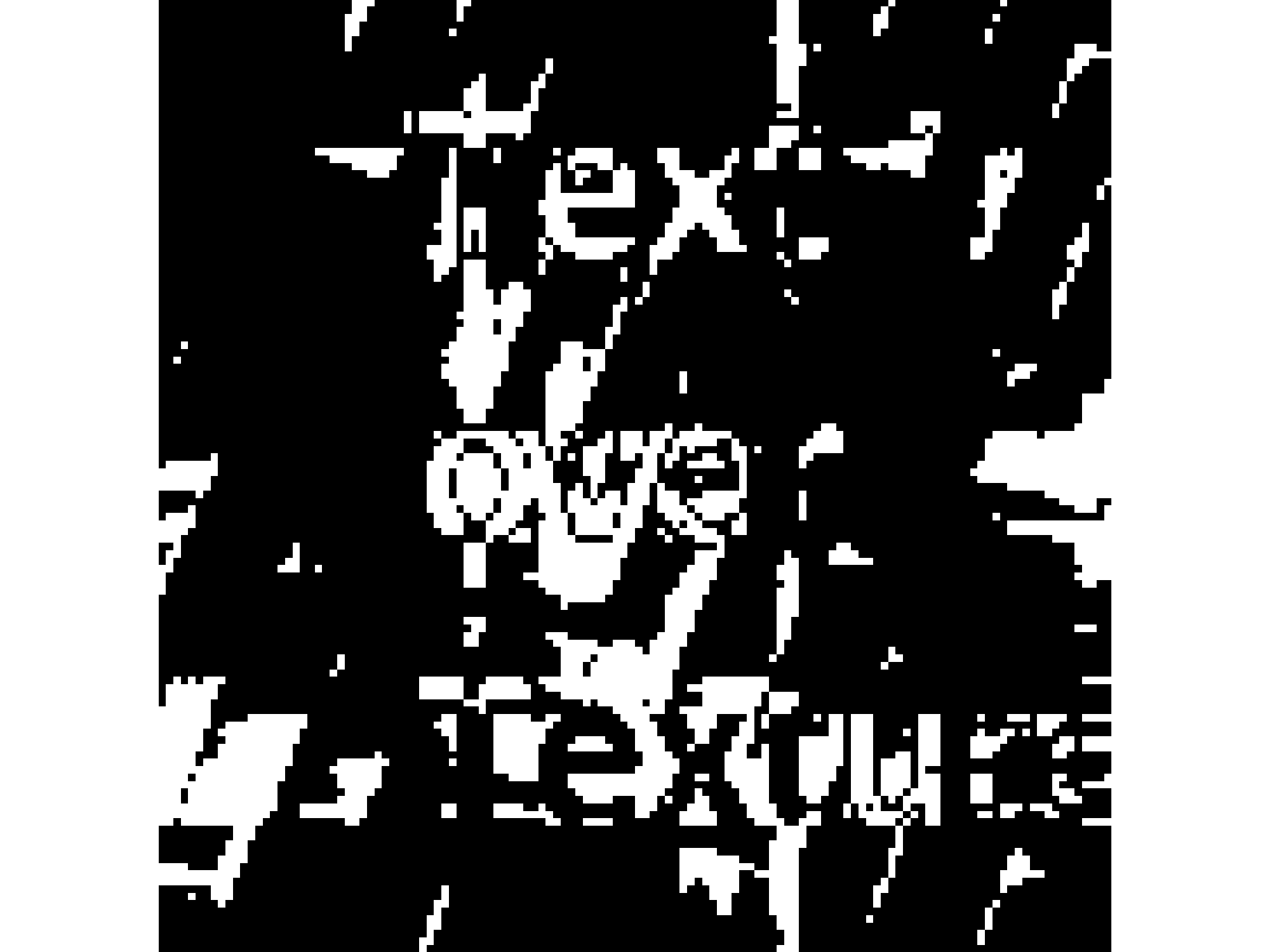}
                \vspace{-0.04cm}
            \hspace{-6cm} 
        \end{subfigure}%
        \begin{subfigure}[b]{0.25\textwidth}
			~ 
                \includegraphics[width=\textwidth]{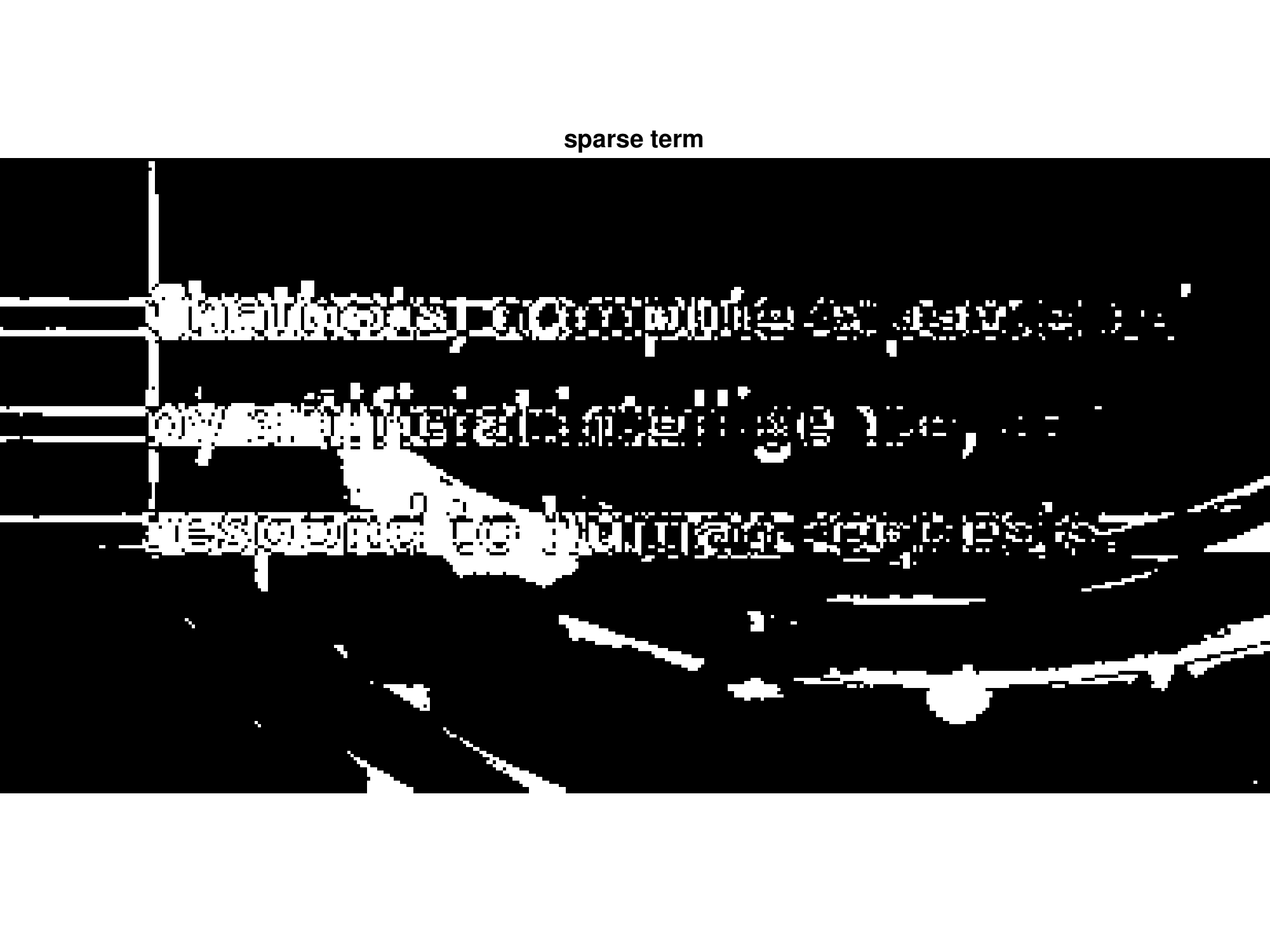}
                \vspace{-0.4cm}
            \hspace{-2cm} 
        \end{subfigure}%
        \begin{subfigure}[b]{0.30\textwidth}
                \includegraphics[width=\textwidth]{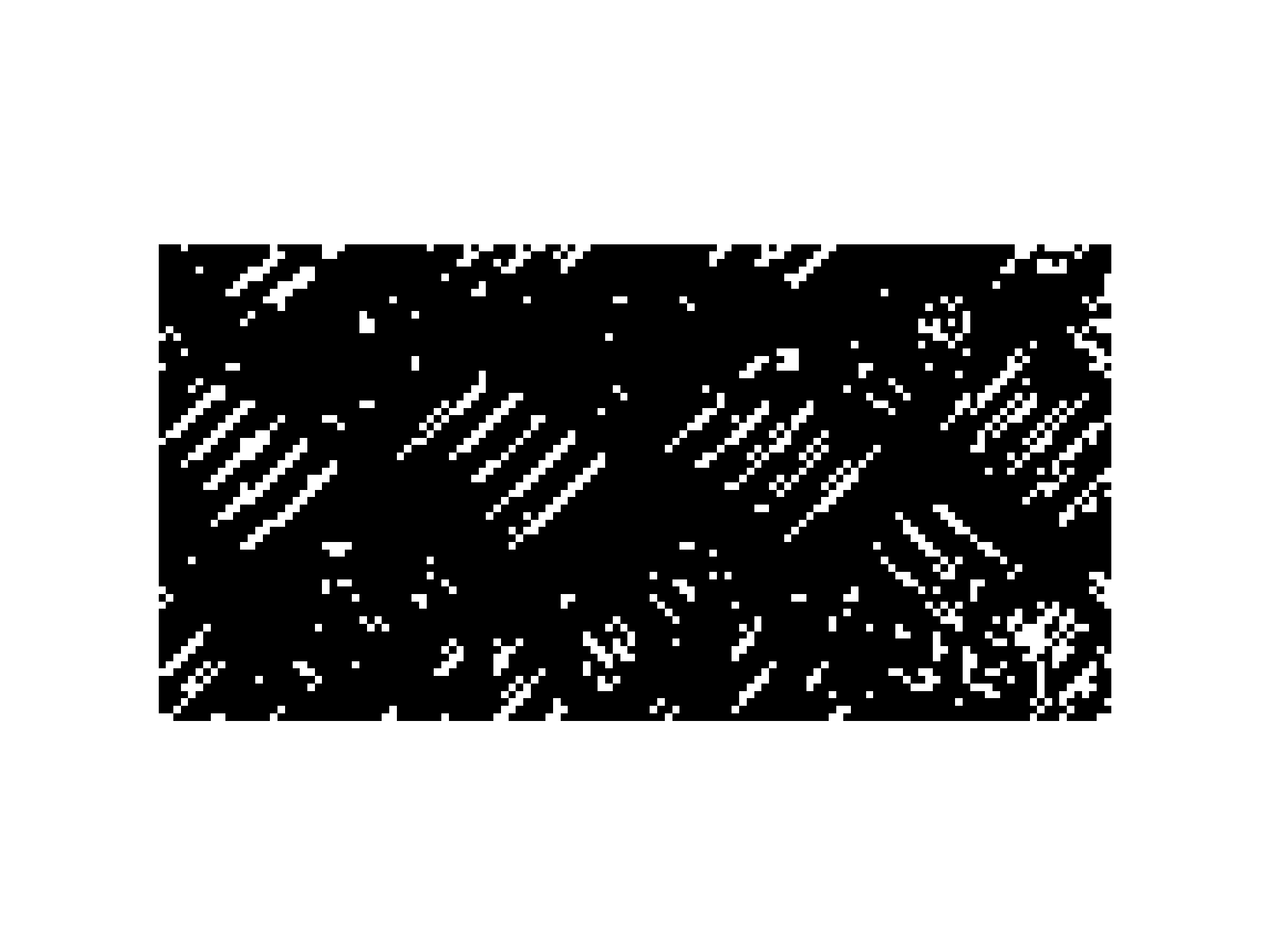}
                 \vspace{-0.61cm}
              \hspace{-4.8cm}
        \end{subfigure}
         \\[1ex]\vspace{-0.9cm}
        \begin{subfigure}[b]{0.22\textwidth}
                \includegraphics[width=\textwidth]{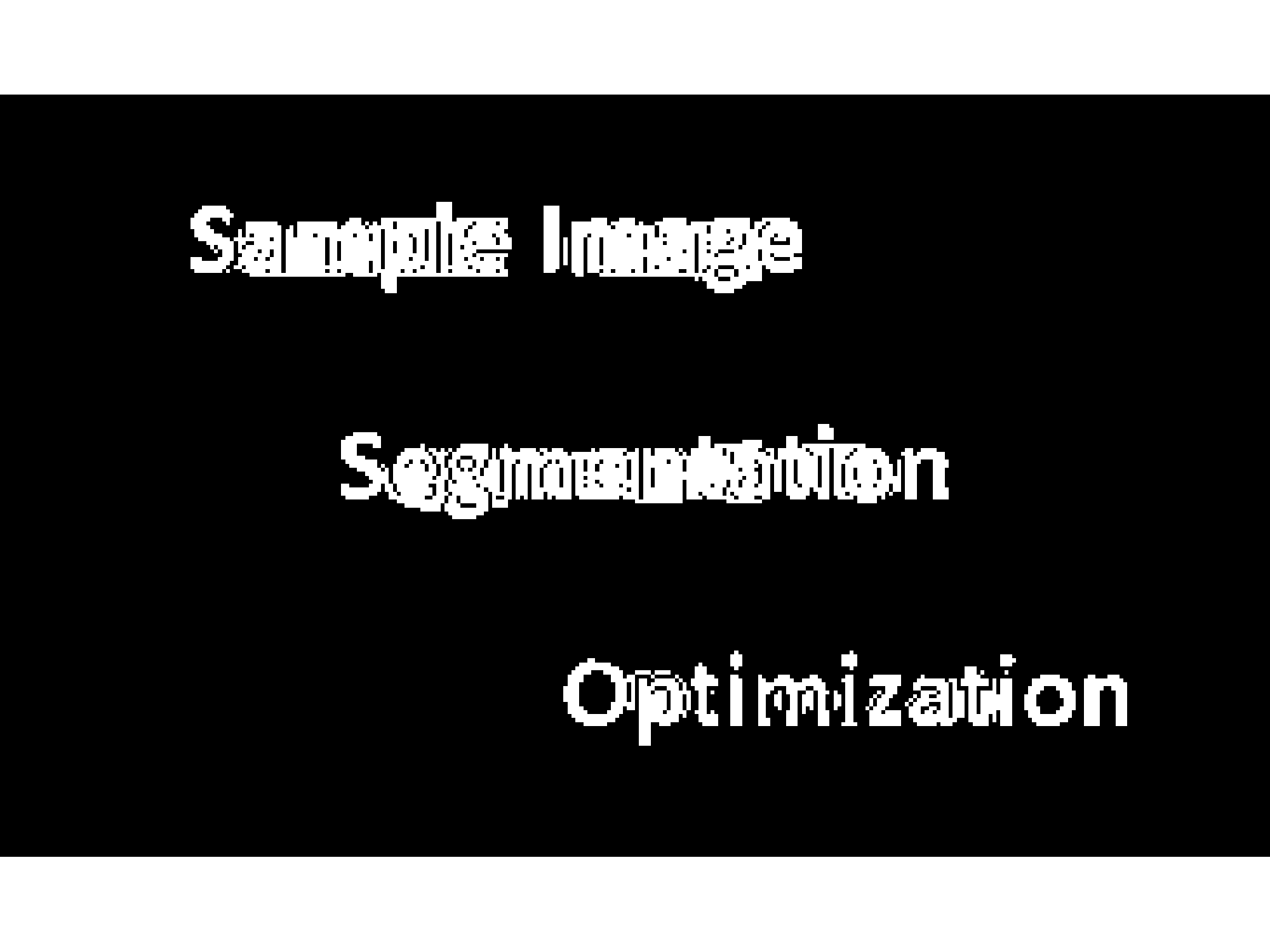}
                                \vspace{-0.32cm}
          \hspace{-1.5cm}    
        \end{subfigure}%
        ~ 
		\vspace{-0.02cm}        
        \begin{subfigure}[b]{0.18\textwidth}
                \includegraphics[width=\textwidth]{texture8_TV-eps-converted-to.pdf}
                \vspace{-0.04cm}
            \hspace{-6cm} 
        \end{subfigure}%
        \begin{subfigure}[b]{0.25\textwidth}
			~ 
                \includegraphics[width=\textwidth]{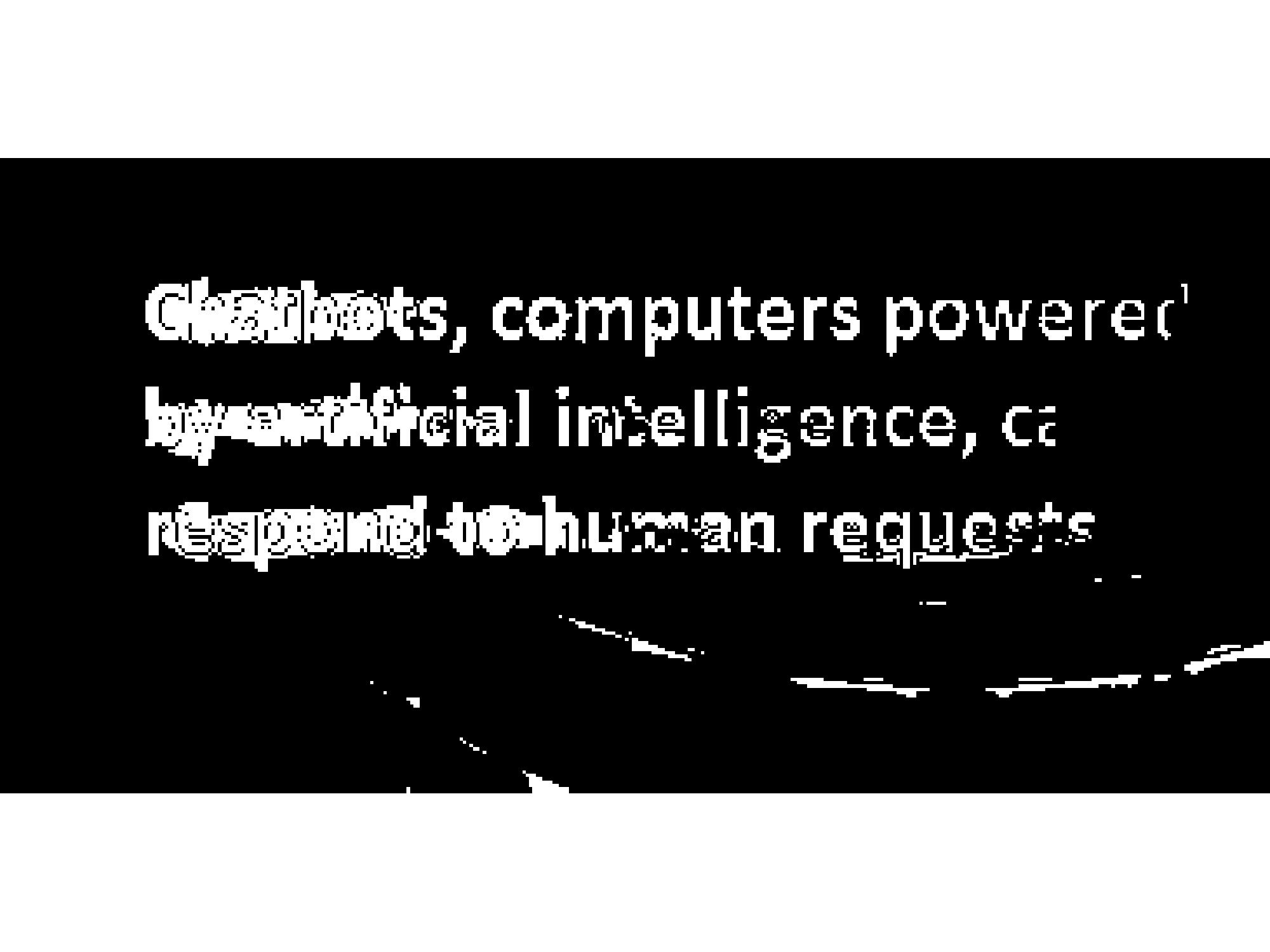}
                \vspace{-0.4cm}
            \hspace{-2cm} 
        \end{subfigure}%
        \begin{subfigure}[b]{0.30\textwidth}
                \includegraphics[width=\textwidth]{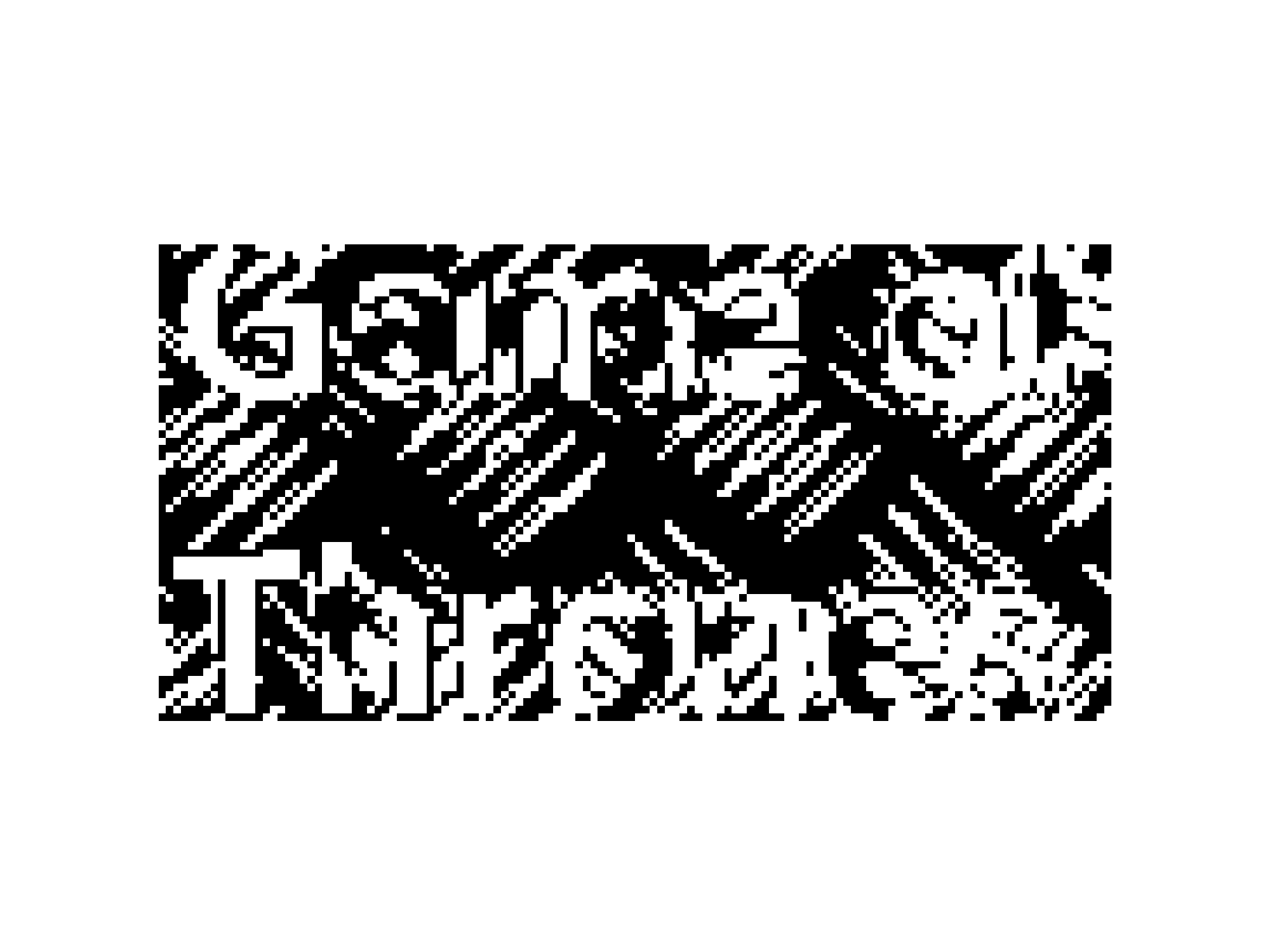}
                 \vspace{-0.61cm}
              \hspace{-4.8cm}
        \end{subfigure}
         \\[1ex]\vspace{-0.9cm}
                 \begin{subfigure}[b]{0.22\textwidth}
                \includegraphics[width=\textwidth]{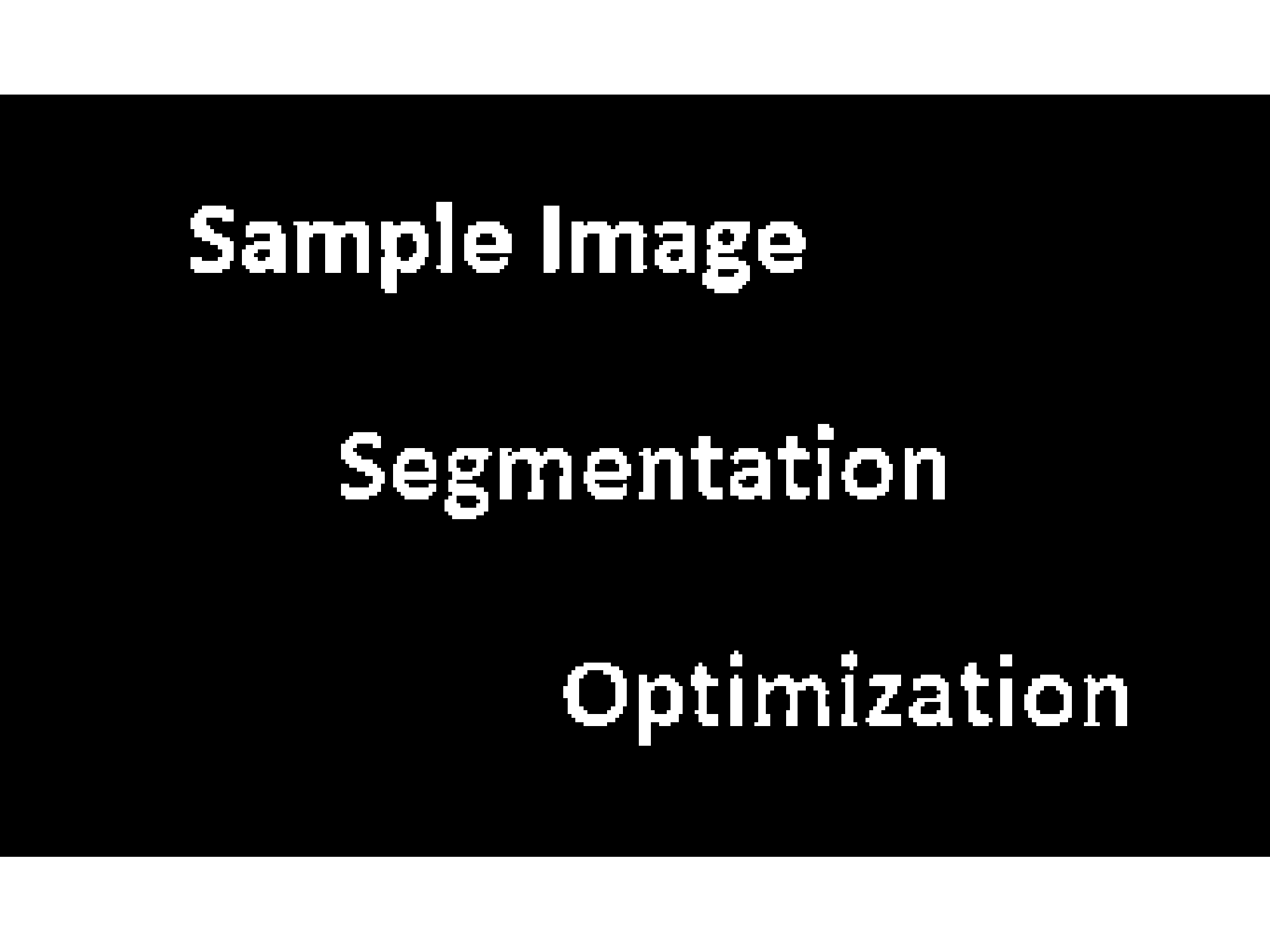}
                                \vspace{-0.3cm}
          \hspace{-1.5cm}    
        \end{subfigure}%
        ~ 
		\vspace{-0.02cm}        
        \begin{subfigure}[b]{0.18\textwidth}
                \includegraphics[width=\textwidth]{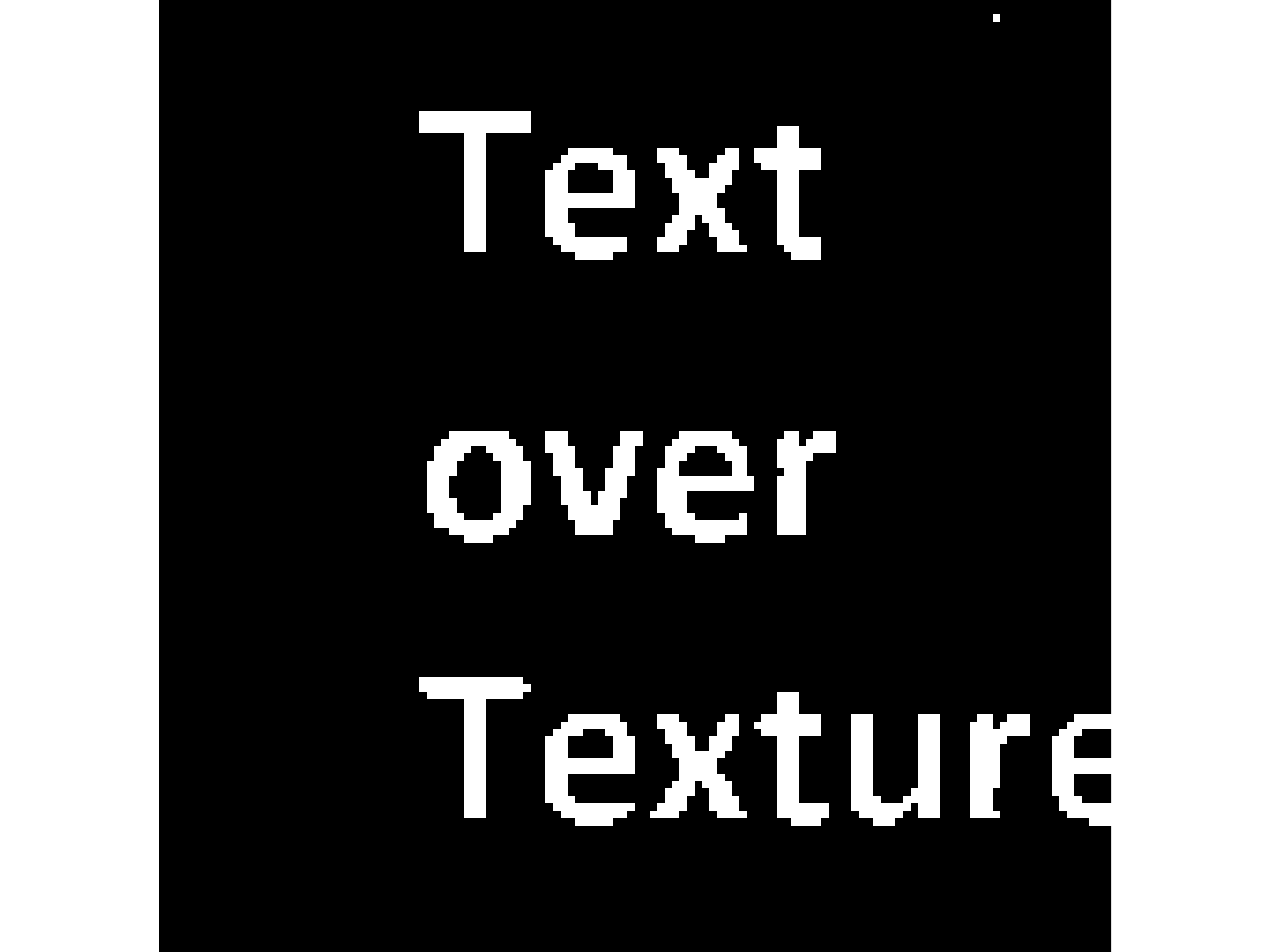}
                \vspace{-0.04cm}
            \hspace{-6cm} 
        \end{subfigure}%
        \begin{subfigure}[b]{0.25\textwidth}
			~ 
                \includegraphics[width=\textwidth]{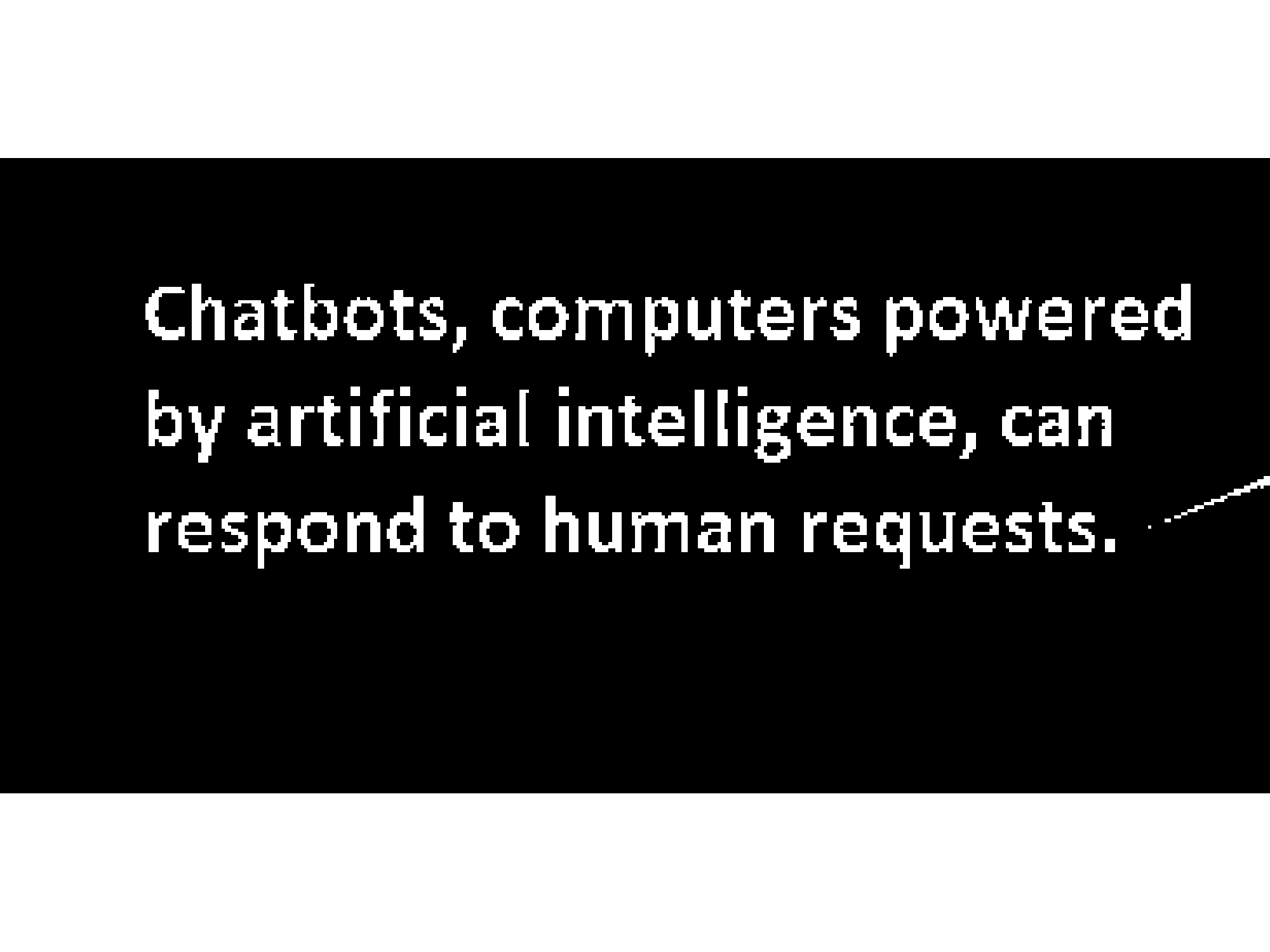}
                \vspace{-0.4cm}
            \hspace{-2cm} 
        \end{subfigure}%
        \begin{subfigure}[b]{0.30\textwidth}
                \includegraphics[width=\textwidth]{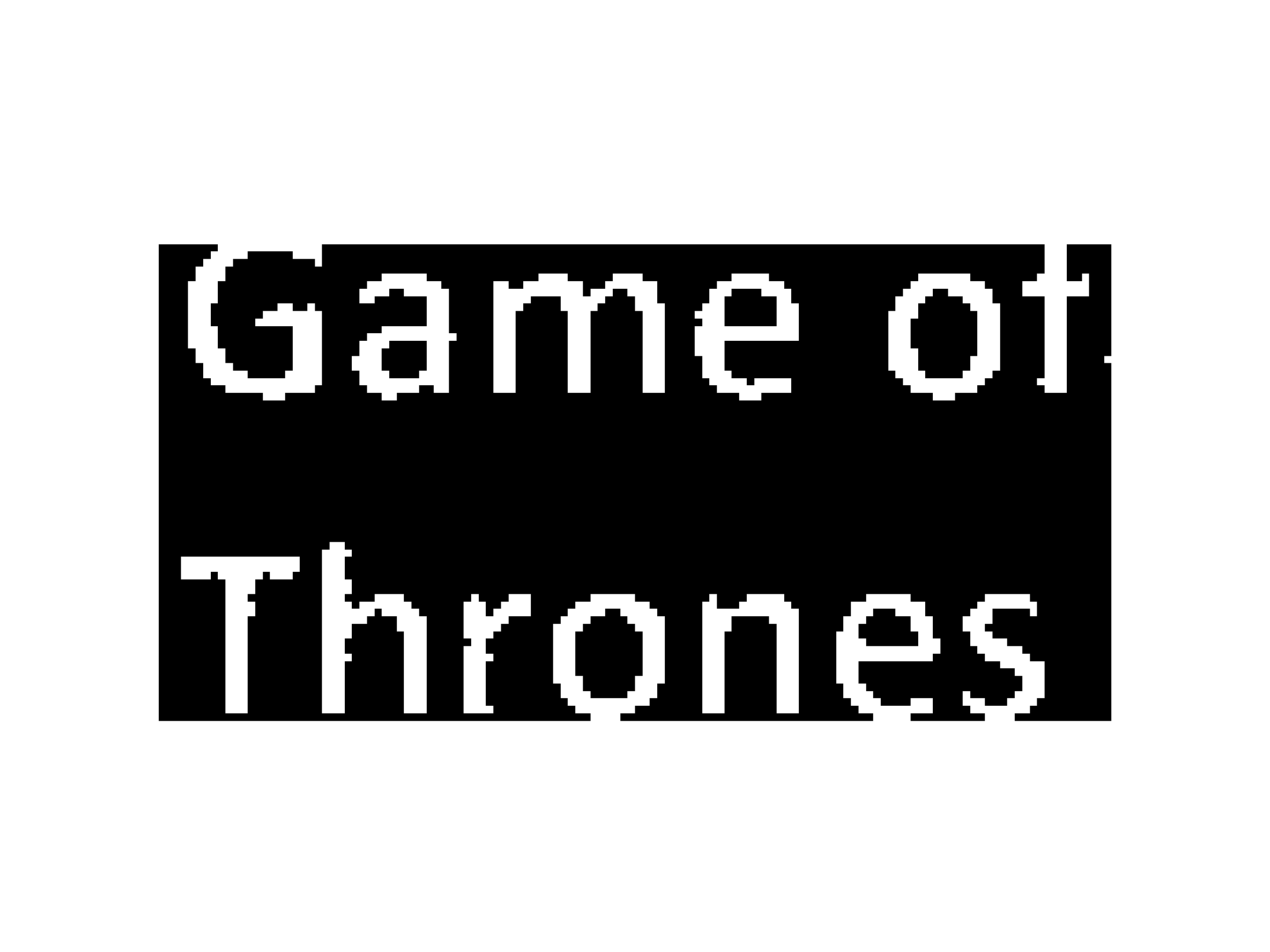}
                 \vspace{-0.61cm}
              \hspace{-4.8cm}
        \end{subfigure}
        \caption{Segmentation result for the text over texture images. The images in the first row denotes the original images. And the images in the second, third, forth and the fifth rows denote the foreground map by hierarchical k-means clustering \cite{clus2}, sparse and low-rank decomposition \cite{candes}, sparse decomposition \cite{mytv}, and the proposed algorithm respectively.}
\end{figure*}

We also provide the average precision, recall and F1 score \cite{metrics} achieved by different algorithms for the above sample images. 
The precision and recall are defined as in Eq. (20), where TP, FP and FN denote true positive, false positive and false negative respectively. In our evaluation, we treat the active elements of the binary mask as positive. 
The balanced F1 score is defined as the harmonic mean of precision and recall, as shown in Eq. (21).
\begin{gather}
 \text{Precision}= \frac{\text{TP}}{\text{TP+FP}} \ , 
\ \ \ \ \text{Recall}= \frac{\text{TP}}{\text{TP+FN}} 
\end{gather}
\begin{gather}
\text{F1}= 2 \ \frac{\text{precision} \times \text{recall}}{\text{precision+recall}}.
\end{gather}
The average precision, recall and F1 score by different algorithms are given in Table 1.
As it can be seen, the proposed scheme achieves much higher precision and recall than hierarchical k-means clustering and sparse decomposition approach. 
We did not provide the results by SPEC \cite{spec} algorithm for these images, since the derived segmentation masks for these test images using SPEC was not satisfactory.

\begin{table}[ht]
\centering
  \caption{Comparison of accuracy of different algorithms for text image segmentation for images in Fig 6.}
  \centering
\begin{tabular}{|m{3.4cm}|m{1.2cm}|m{1.2cm}|m{1.2cm}|}
\hline
Segmentation Algorithm  &  \  \ Precision & \ \  Recall & \  F1 score\\
\hline
 Hierarchical Clustering \cite{clus2} & \ \ \ 66.5\% & \ \ \ 92\% & \ \ \ 77.2\% \\
\hline
 Sparse and Low-rank \cite{candes} & \ \ \ 54\% & \ \ \ 62.8\% & \ \ \ 57.7\% \\
\hline 
 Sparse Dec. with TV \cite{mytv} & \ \ \  71\% & \ \ \  91.7\% & \ \ \  80\% \\
\hline
 The proposed algorithm & \ \ \ 95\%  & \ \ \ 92.5\%  & \ \ \  93.7\%\\
\hline
\end{tabular}
\label{TblComp}
\end{table}

\subsection{Application for Motion Segmentation}
In this section, we demonstrate the application of the proposed algorithm for motion based object segmentation in video. We assume the video undergoes a global camera motion (modeled by a homography mapping) as well as localized object motion (modeled by a sparse component). The optical flow field between two frames can thus being modeled by a masked decomposition of the global motion and object motion. 

To extract the optical flow, we use the optical flow implementation in \cite{of1}, \cite{of2}.
We then use the formulation in Eq.~(18), with $\lambda_1= 1$, $\lambda_2= 0.8$ and $\lambda_3= 0.5$, to find both the global motion parameters and the object mask $w$. Note that the estimated $w$ from Eq. (18) is a continuous mask where each element is in [0,1], and we threshold these values to derive the binary mask for foreground.
We compare our work with the simple least squares fitting method  where we fit the homography model to the whole optical flow by solving the optimization problem in (22), and detect foreground pixels by thresholding the fitting error image.
\begin{equation}
\begin{aligned}
& \underset{a}{\text{min}}
 \  \ \frac{1}{2} \| b_x- P_xa  \|_2^2+\frac{1}{2} \| b_y- P_ya \|_2^2.
\end{aligned}
\end{equation}

The motion segmentation results using the proposed algorithm, and the comparison with the least squares fitting for two videos are provided in Figure 7 and Figure 8. 
As we can see the proposed algorithm achieves better segmentation compared to the baseline.
We would like to note that, this is a preliminary study to show the motion segmentation as one of the potential applications of this work, and the result could be much improved by using more accurate optical flow extraction scheme.

\begin{figure}
        \centering
        \vspace{-1.4cm}
        \hspace{0.8cm}
        \begin{subfigure}[b]{0.35\textwidth}
        \hspace{-0.55cm}
                \includegraphics[width=\textwidth]{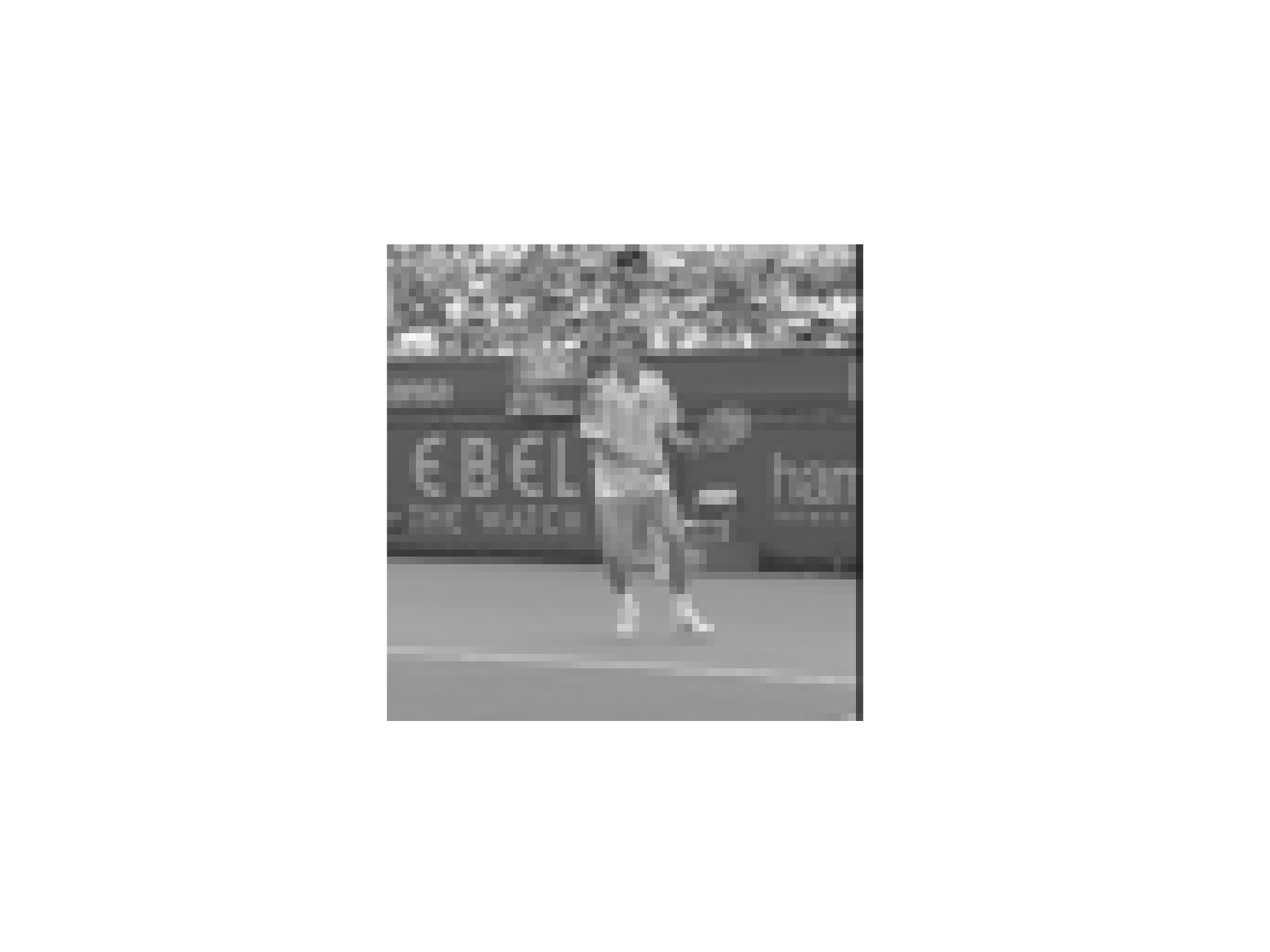}
                \vspace{-0.5cm}
            \hspace{-3.5cm} 
        \end{subfigure}%
        ~ 
        \begin{subfigure}[b]{0.35\textwidth}
        \hspace{-2.8cm}
                \includegraphics[width=\textwidth]{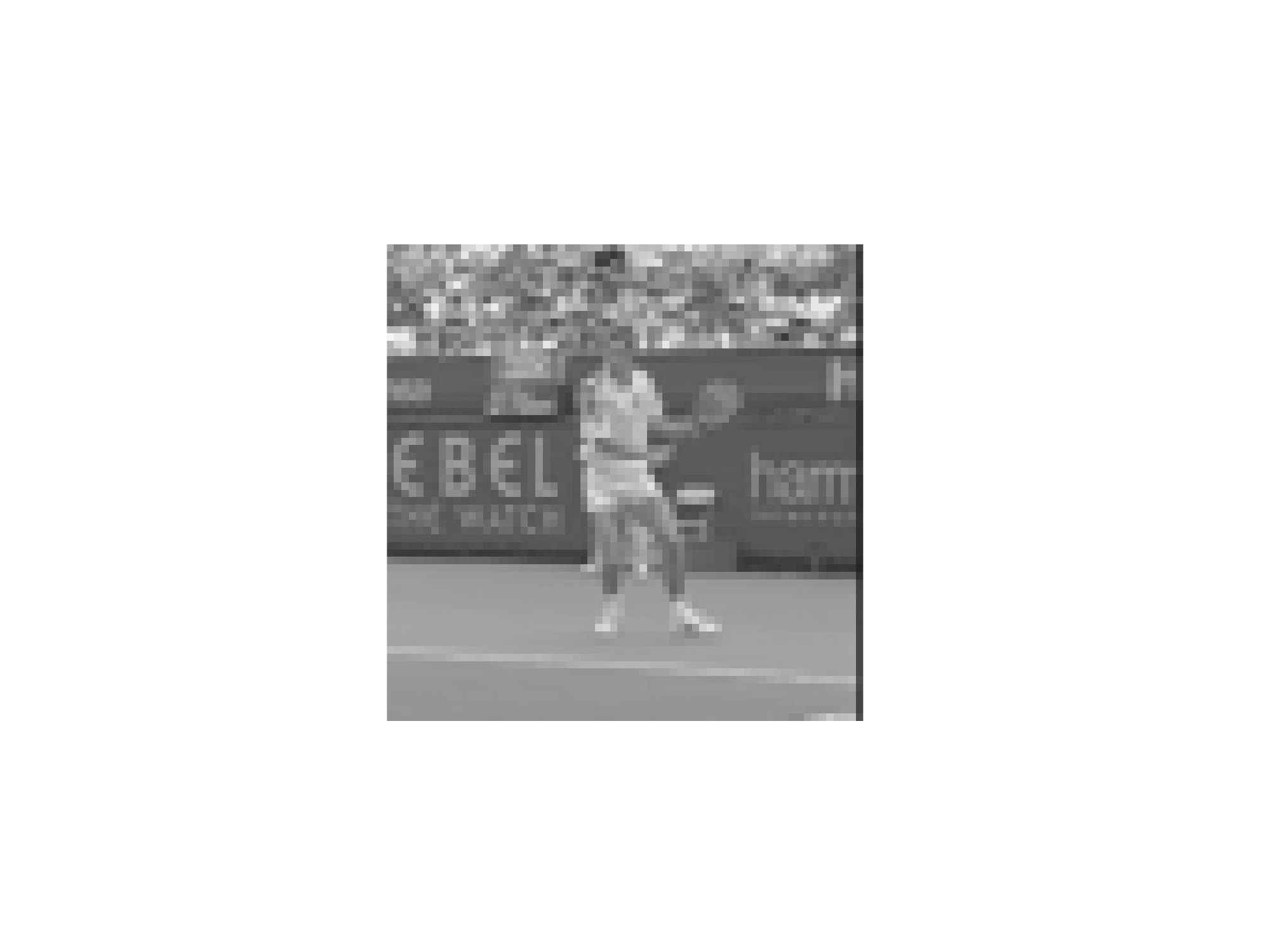}
                 \vspace{-0.45cm}
              \hspace{-4.8cm}
        \end{subfigure}
         \\[1ex]
                 \centering
        \vspace{-1.5cm}
        \hspace{-1.4cm}

        \begin{subfigure}[b]{0.35\textwidth}
        \hspace{-0.55cm}
                \includegraphics[width=\textwidth]{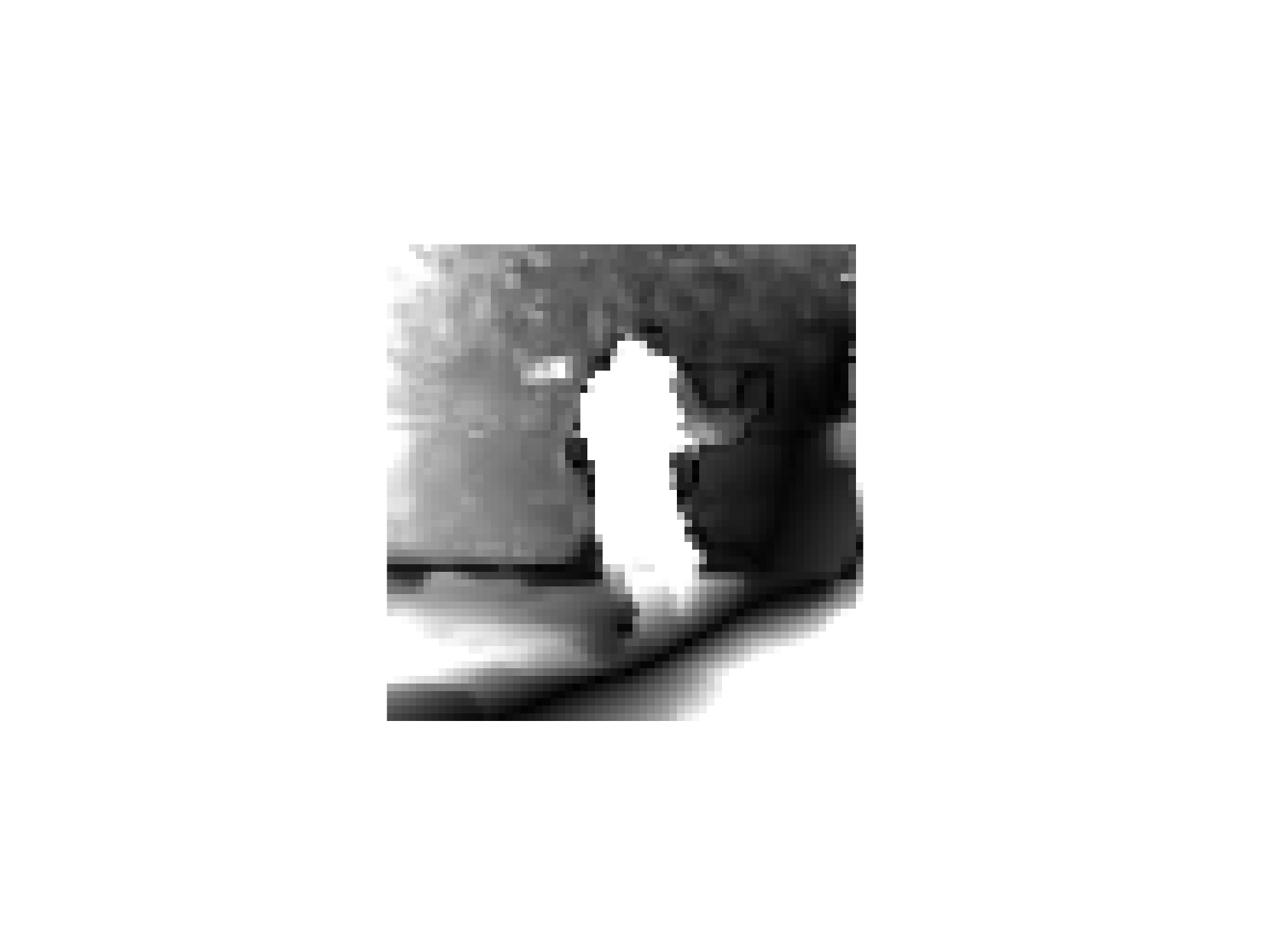}
                \vspace{-0.5cm}
            \hspace{-3cm} 
        \end{subfigure}%
        ~ 
        \begin{subfigure}[b]{0.35\textwidth}
       \hspace{-2.8cm}
                \includegraphics[width=\textwidth]{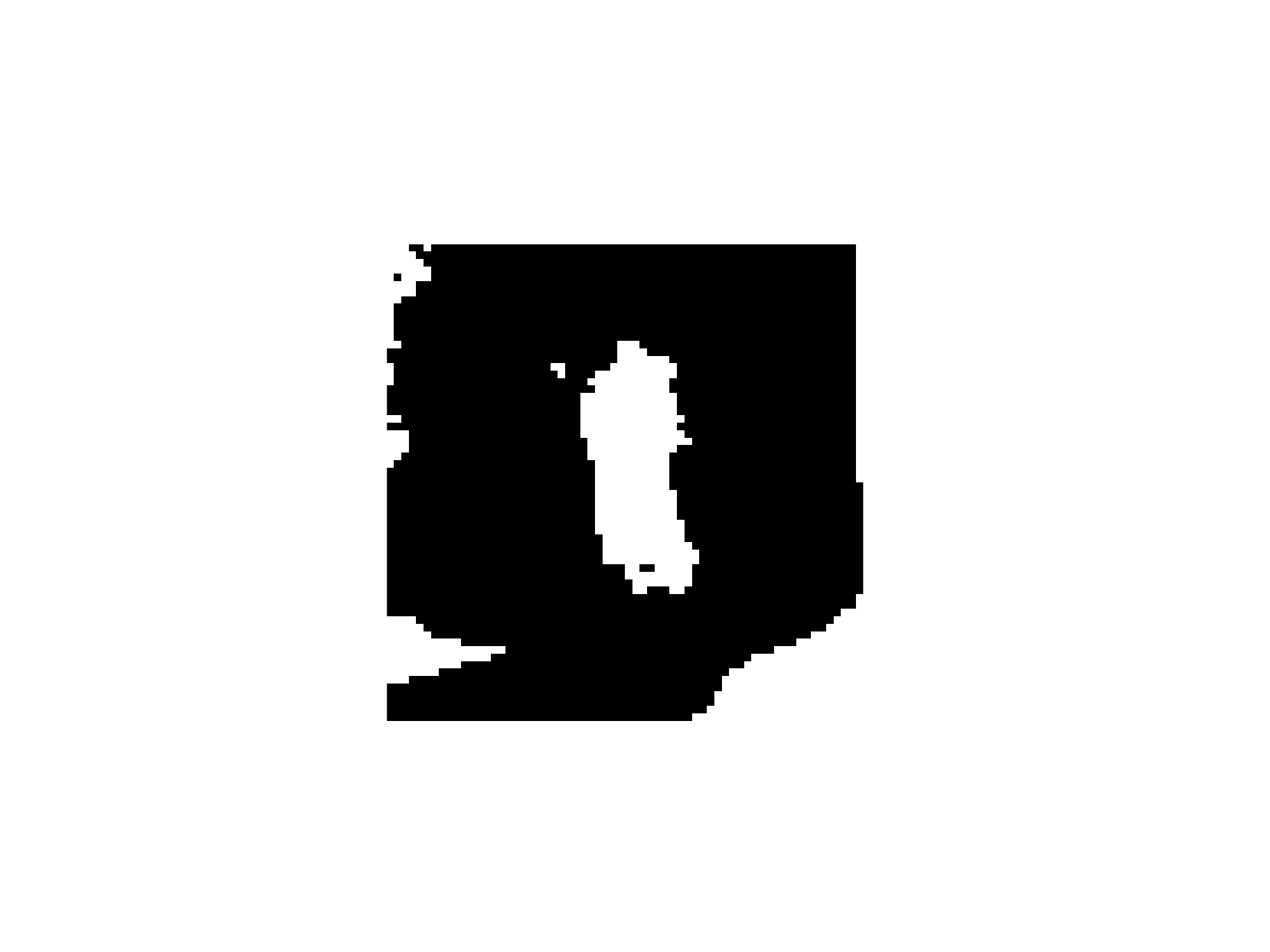}
                 \vspace{-0.45cm}
              \hspace{-7.8cm}
        \end{subfigure}
         \\[1ex]
                 \centering
        \vspace{-1.5cm}
        \hspace{-1.4cm}

        \begin{subfigure}[b]{0.35\textwidth}
        \hspace{-0.55cm}
                \includegraphics[width=\textwidth]{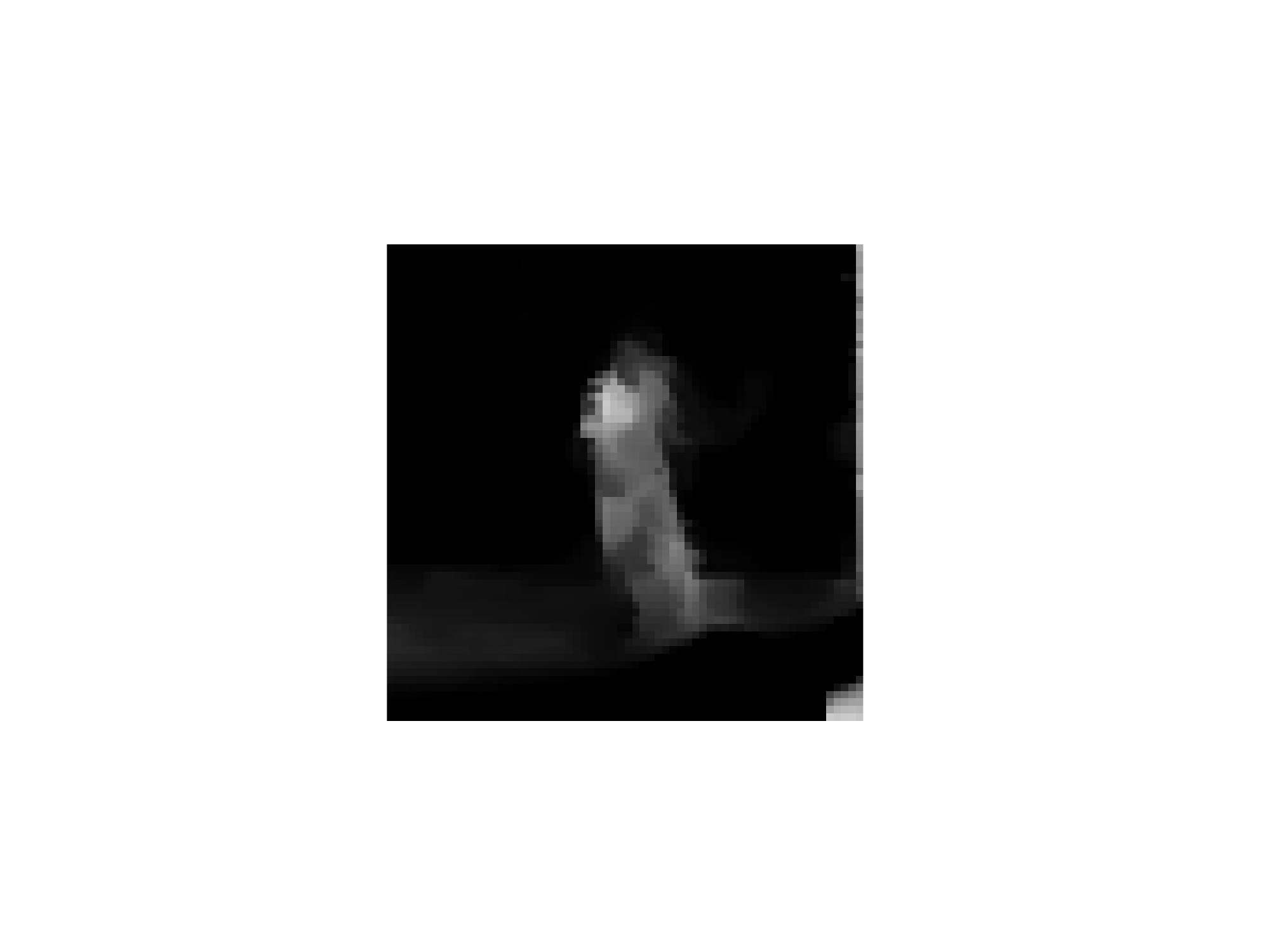}
                \vspace{-0.5cm}
            \hspace{-3cm} 
        \end{subfigure}%
        ~ 
        \begin{subfigure}[b]{0.35\textwidth}
        \hspace{-2.8cm}
                \includegraphics[width=\textwidth]{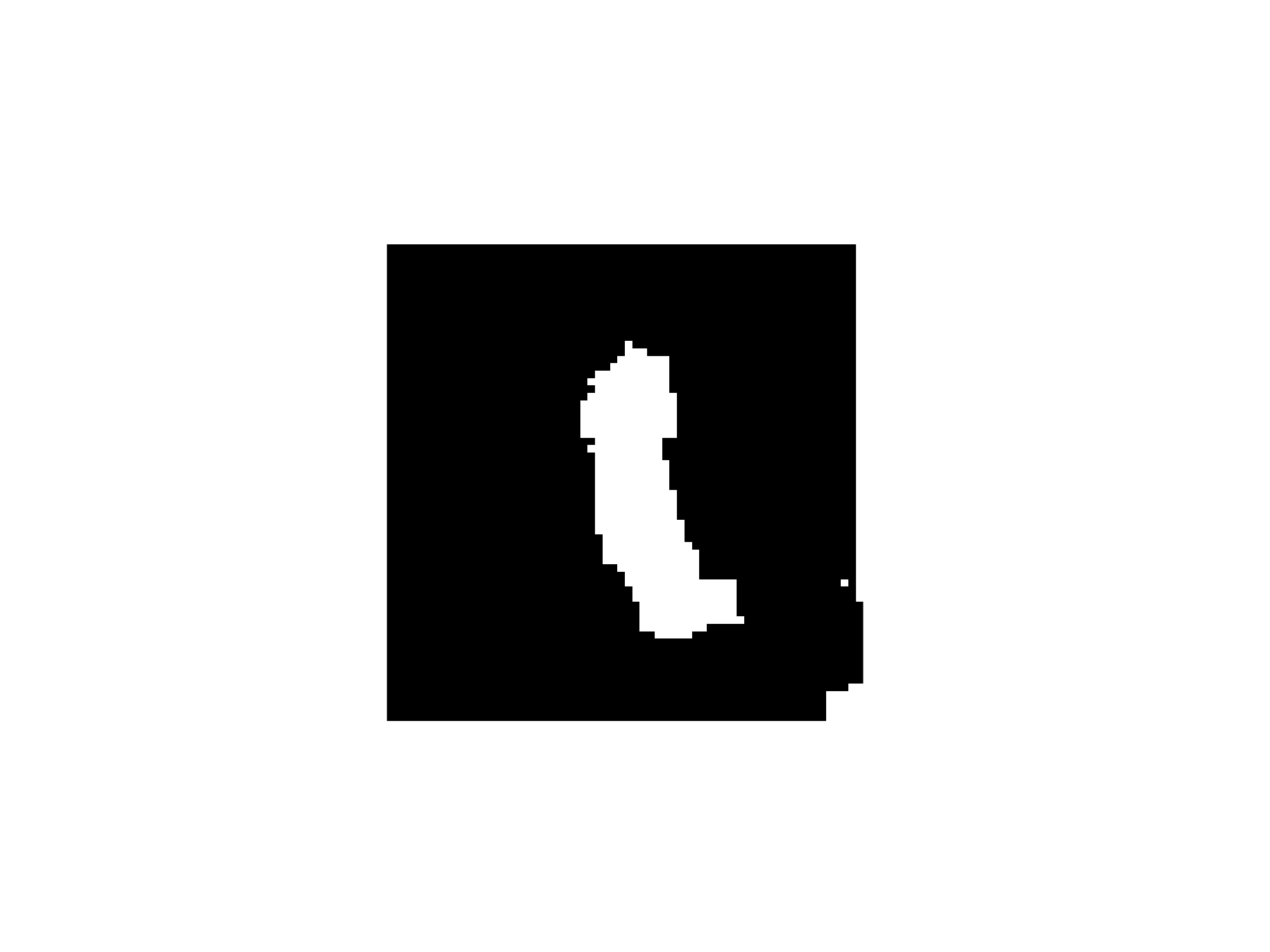}
                 \vspace{-0.45cm}
              \hspace{-4.8cm}
        \end{subfigure}   \\[1ex]
        \caption{ Motion segmentation result for Stefan video. The images in the first row denote two consecutive frames from Stefan test video. The images in the second row denote the global motion estimation error and its corresponding binary mask. The images in the last row denote the continuous and the binary motion masks using the proposed algorithm.}
\end{figure}

\begin{figure}
        \centering
        \vspace{-1.35cm}
        \hspace{0.8cm}
        \begin{subfigure}[b]{0.35\textwidth}
        \hspace{-0.55cm}
                \includegraphics[width=\textwidth]{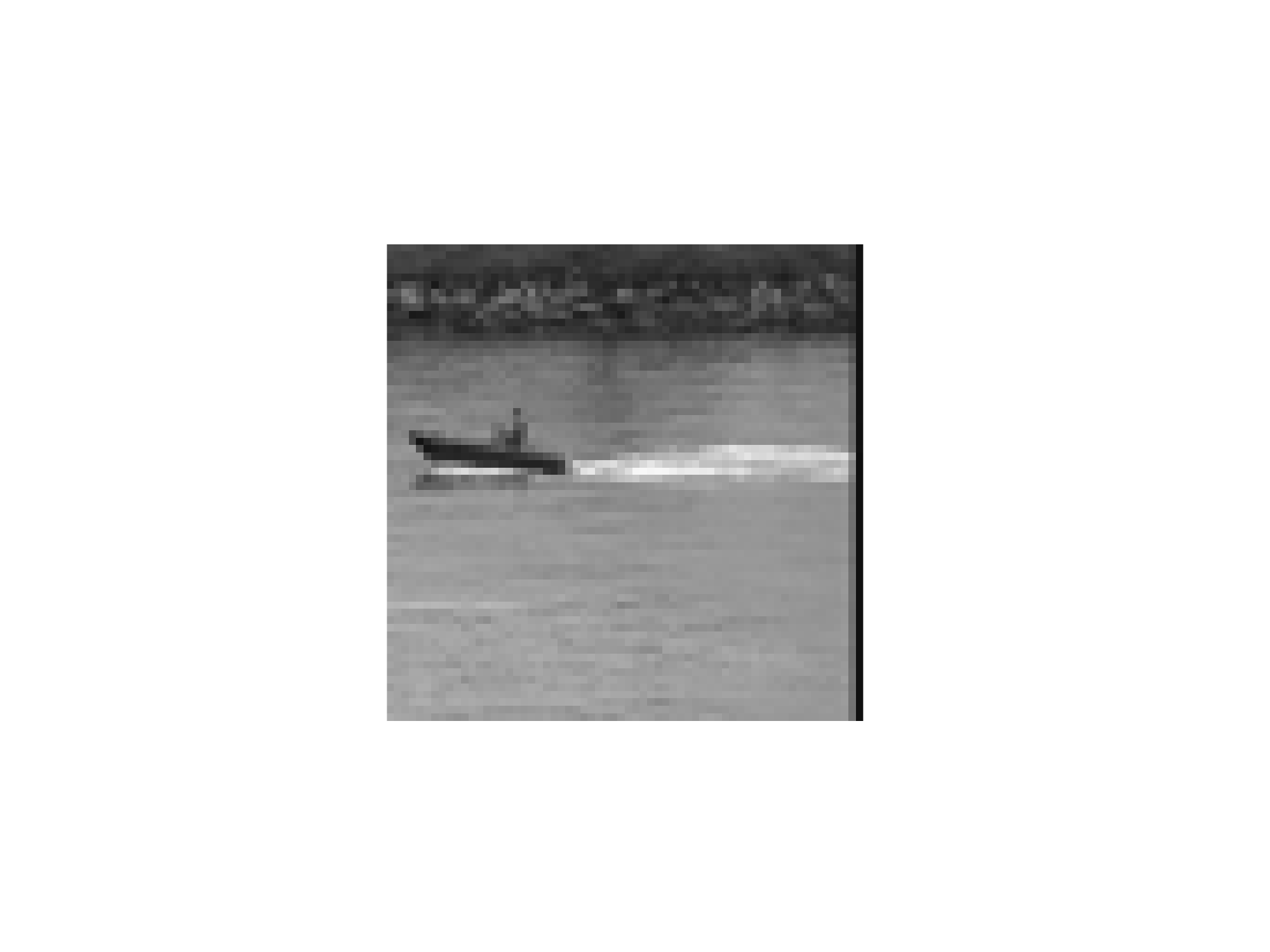}
                \vspace{-0.5cm}
            \hspace{-3.5cm} 
        \end{subfigure}%
        ~ 
        \begin{subfigure}[b]{0.35\textwidth}
        \hspace{-2.8cm}
                \includegraphics[width=\textwidth]{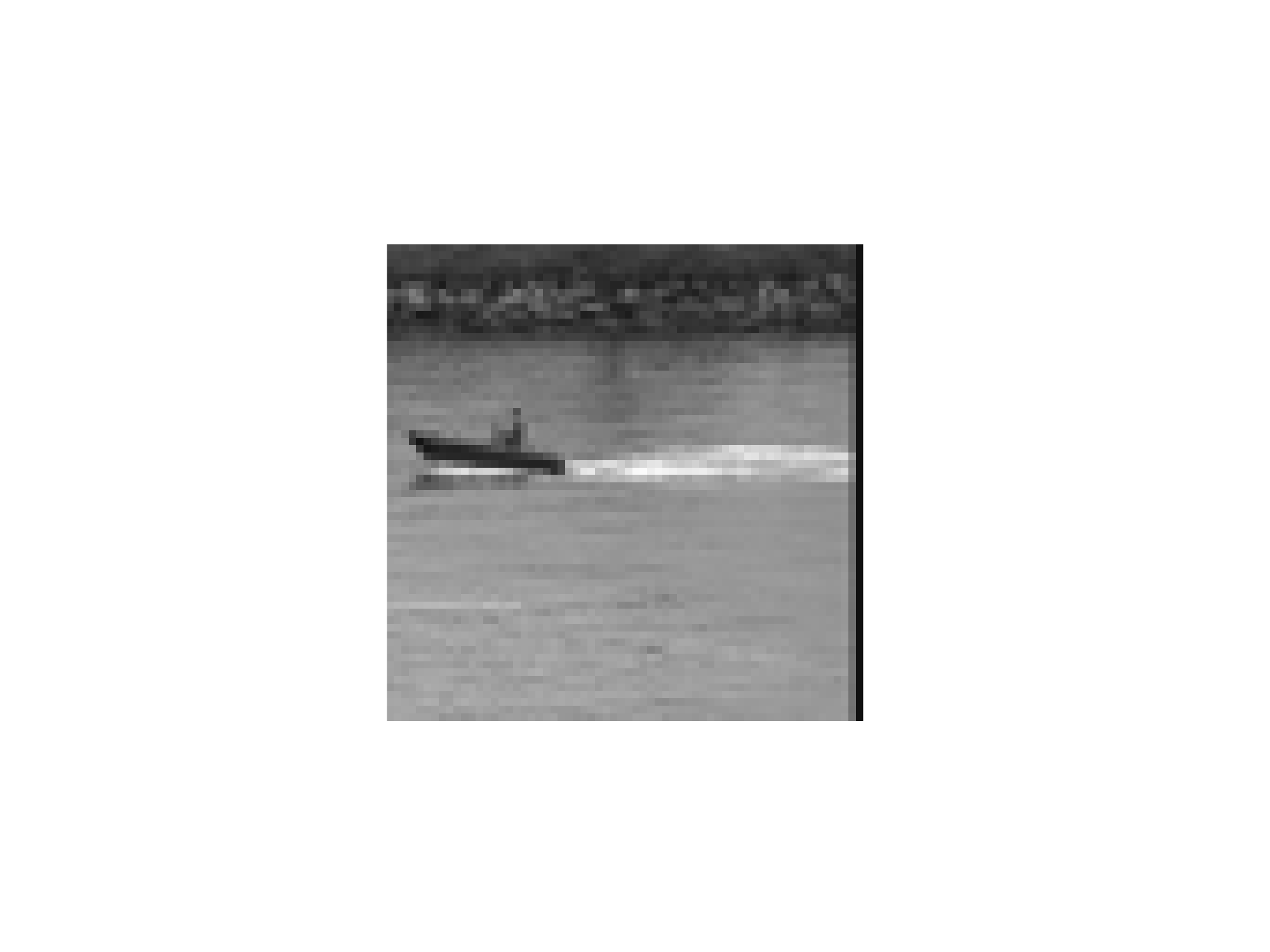}
                 \vspace{-0.45cm}
              \hspace{-4.8cm}
        \end{subfigure}
         \\[1ex]
                 \centering
        \vspace{-1.5cm}
        \hspace{-1.4cm}

        \begin{subfigure}[b]{0.35\textwidth}
        \hspace{-0.55cm}
                \includegraphics[width=\textwidth]{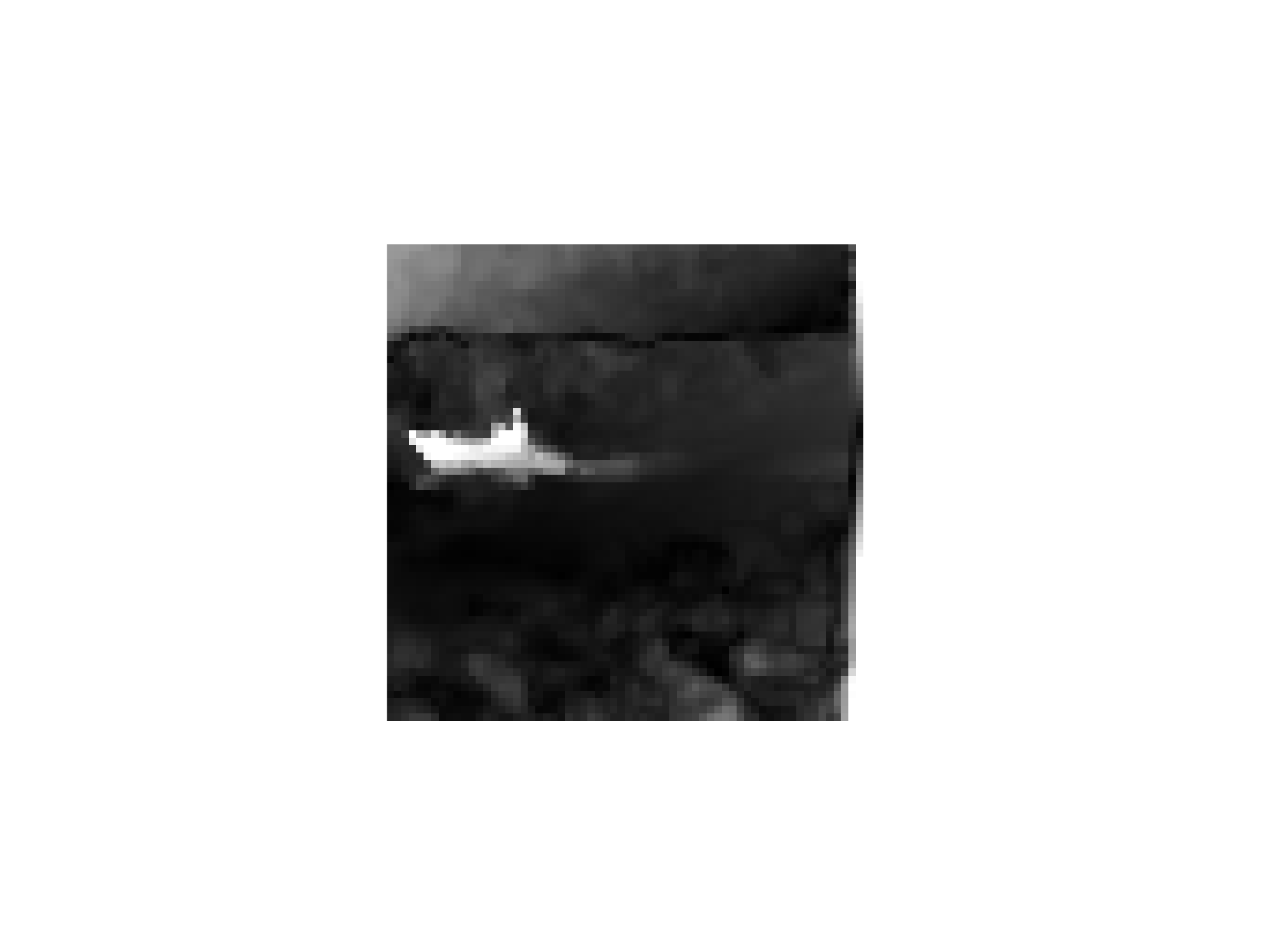}
                \vspace{-0.5cm}
            \hspace{-3cm} 
        \end{subfigure}%
        ~ 
        \begin{subfigure}[b]{0.35\textwidth}
       \hspace{-2.8cm}
                \includegraphics[width=\textwidth]{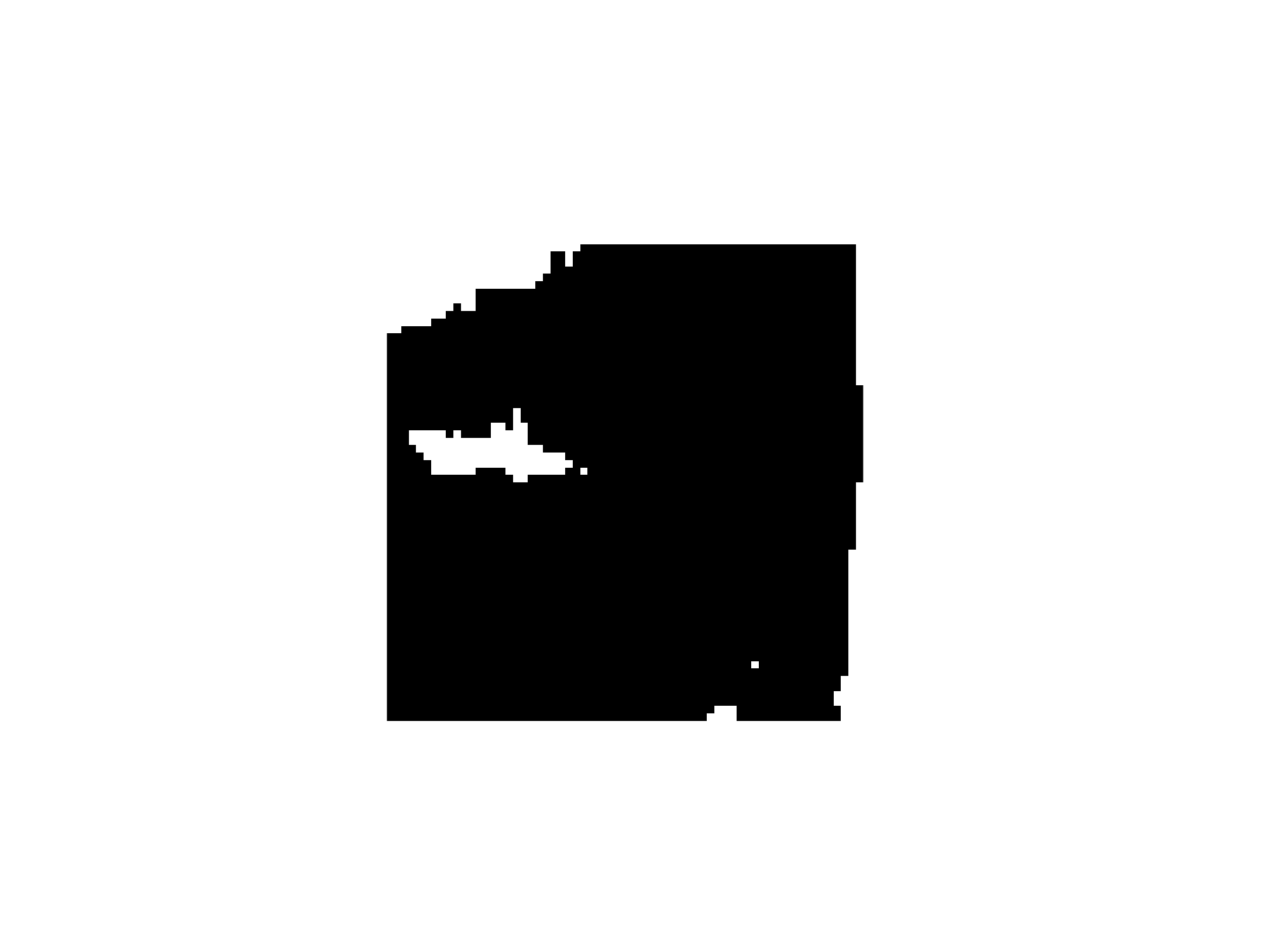}
                 \vspace{-0.45cm}
              \hspace{-7.8cm}
        \end{subfigure}
         \\[1ex]
                 \centering
        \vspace{-1.5cm}
        \hspace{-1.4cm}

        \begin{subfigure}[b]{0.35\textwidth}
        \hspace{-0.55cm}
                \includegraphics[width=\textwidth]{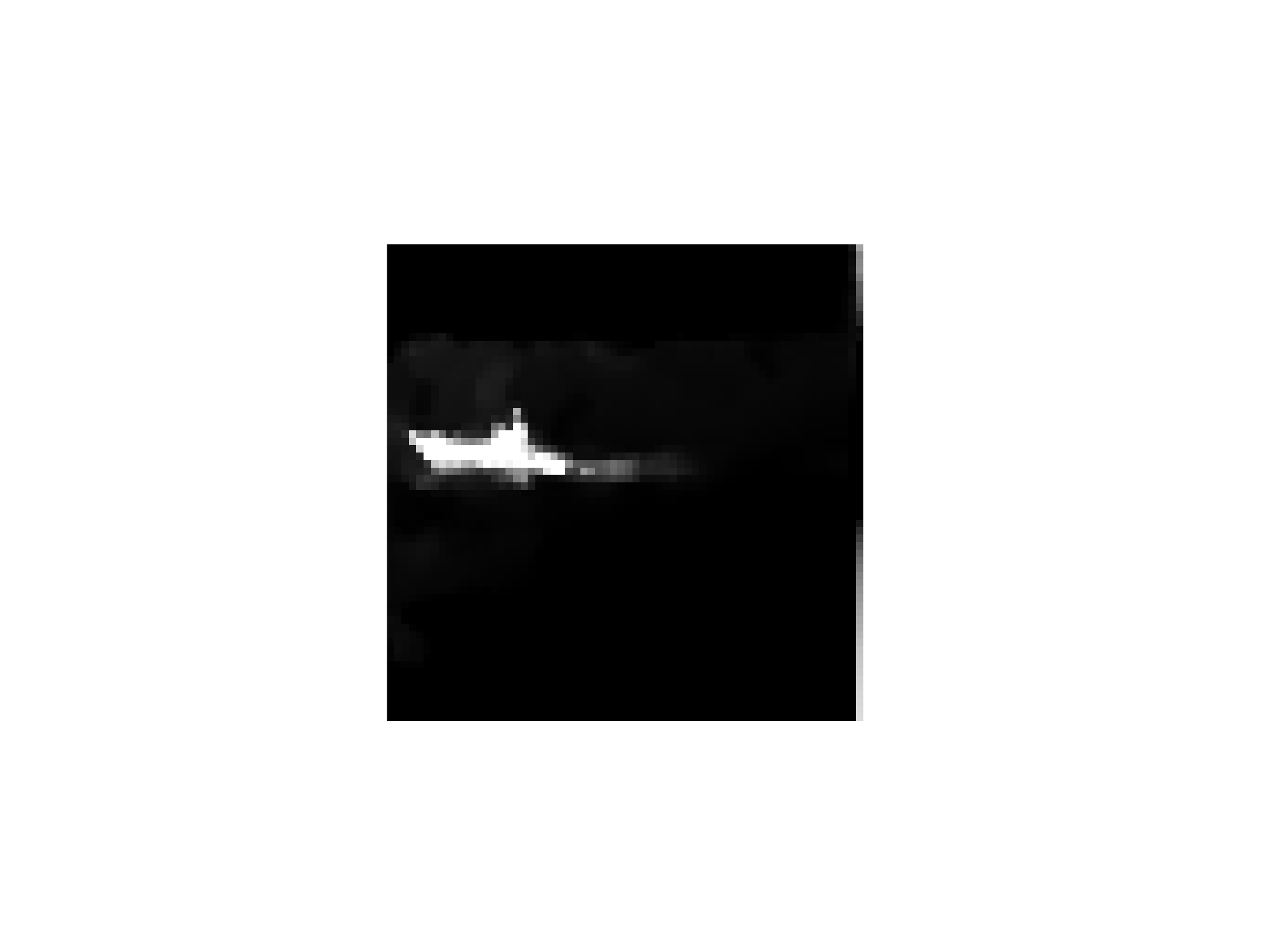}
                \vspace{-0.5cm}
            \hspace{-3cm} 
        \end{subfigure}%
        ~ 
        \begin{subfigure}[b]{0.35\textwidth}
        \hspace{-2.8cm}
                \includegraphics[width=\textwidth]{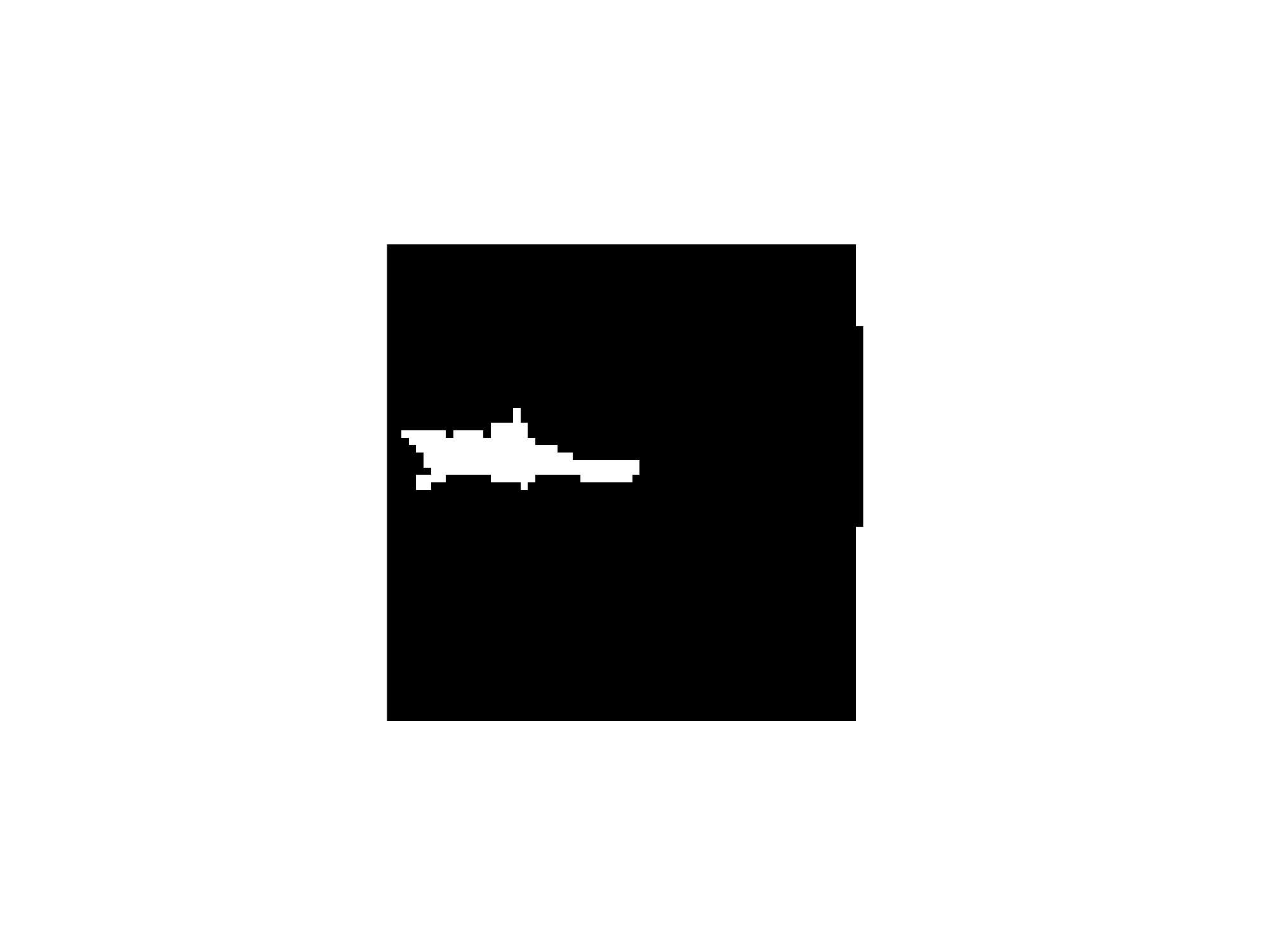}
                 \vspace{-0.45cm}
              \hspace{-4.8cm}
        \end{subfigure}   \\[1ex]
        \caption{ Motion segmentation result for Coastguard video. The images in the first row denote two consecutive frames from Coastguard test video. The images in the second row denote the global motion estimation error and its corresponding binary mask. The images in the last row denote the continuous and the binary motion masks using the proposed algorithm.}
\end{figure}

\subsection{Binarization at each step vs. at the end}
As mentioned earlier, the original masked decomposition problem requires the solution of a binary optimization problem.
To make it a tractable problem, we approximate the binary variables with continuous variables in $[0,1]$ (called linear relaxation), and binarize them after solving the relaxed optimization problem.
There are two ways to do this binarization: 
The first approach solves the optimization problem in Eq. (12), and binarizes the variables $w$ at the very end;
The second approach binarizes the variables $w$ after each update of $w$ in algorithm 1. 
We have tested both these approaches for some of the test images, and provided the results in Fig 9.
As we can see, doing the binarization at the very end works better for all images.
Results presented previously in Secs. A-C are all obtained with the first approach.

\begin{figure*}
        \centering
        \vspace{-0.1cm}
        \begin{subfigure}[b]{0.22\textwidth}
                \includegraphics[width=\textwidth]{test_orig-eps-converted-to.pdf}
                                \vspace{-0.3cm}
          \hspace{-1.5cm}    
        \end{subfigure}%
        ~ 
		\vspace{-0.02cm}        
        \begin{subfigure}[b]{0.18\textwidth}
                \includegraphics[width=\textwidth]{texture8_orig-eps-converted-to.pdf}
                \vspace{-0.04cm}
            \hspace{-6cm} 
        \end{subfigure}%
        \begin{subfigure}[b]{0.25\textwidth}
			~ 
                \includegraphics[width=\textwidth]{test2_orig-eps-converted-to.pdf}
                \vspace{-0.4cm}
            \hspace{-2cm} 
        \end{subfigure}%
        \begin{subfigure}[b]{0.30\textwidth}
                \includegraphics[width=\textwidth]{texture10_orig-eps-converted-to.pdf}
                 \vspace{-0.61cm}
              \hspace{-4.8cm}
        \end{subfigure}
         \\[1ex] \vspace{-0.9cm}
        \begin{subfigure}[b]{0.22\textwidth}
                \includegraphics[width=\textwidth]{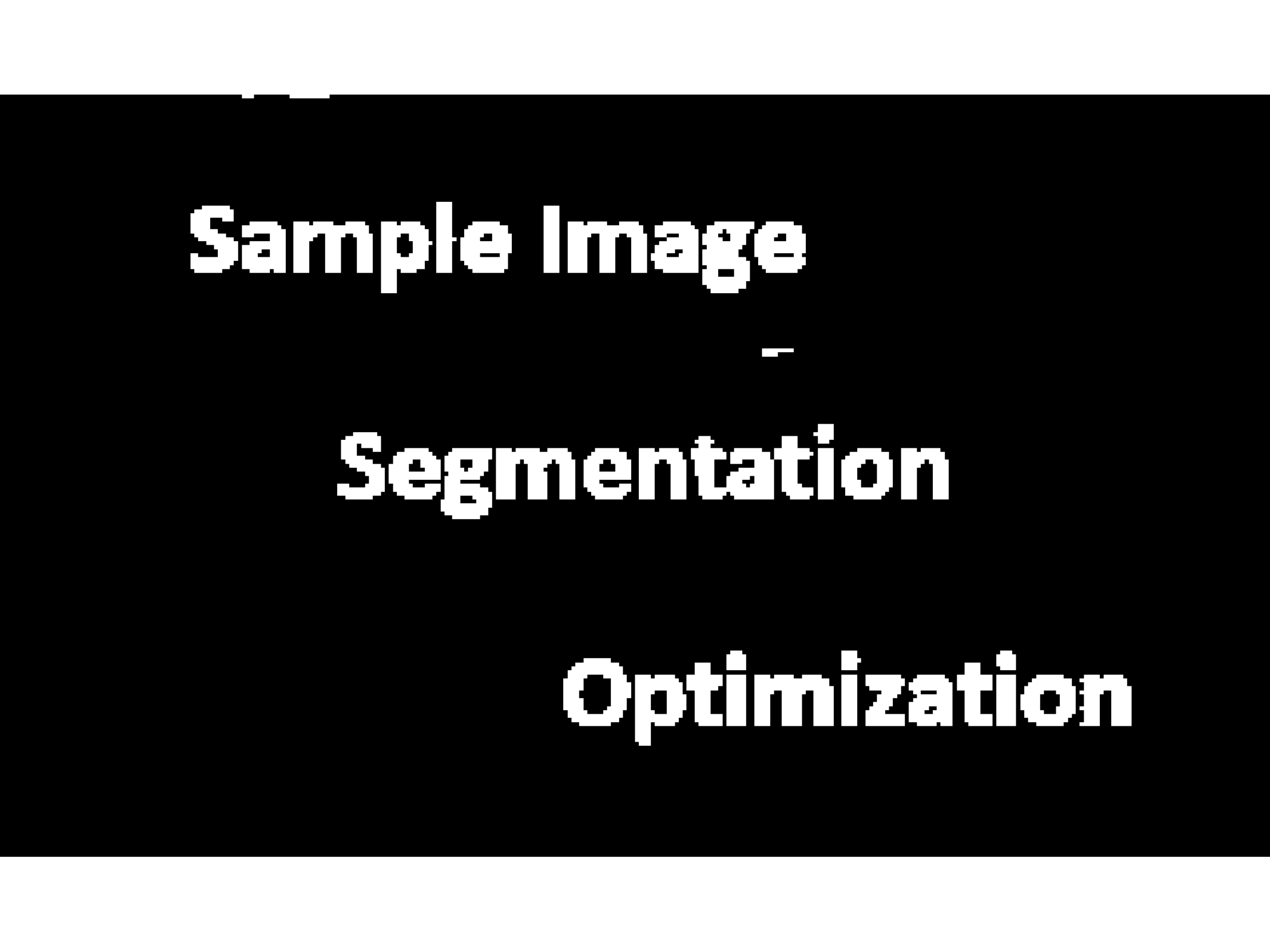}
                                \vspace{-0.32cm}
          \hspace{-1.5cm}    
        \end{subfigure}%
        ~ 
		\vspace{-0.02cm}        
        \begin{subfigure}[b]{0.18\textwidth}
                \includegraphics[width=\textwidth]{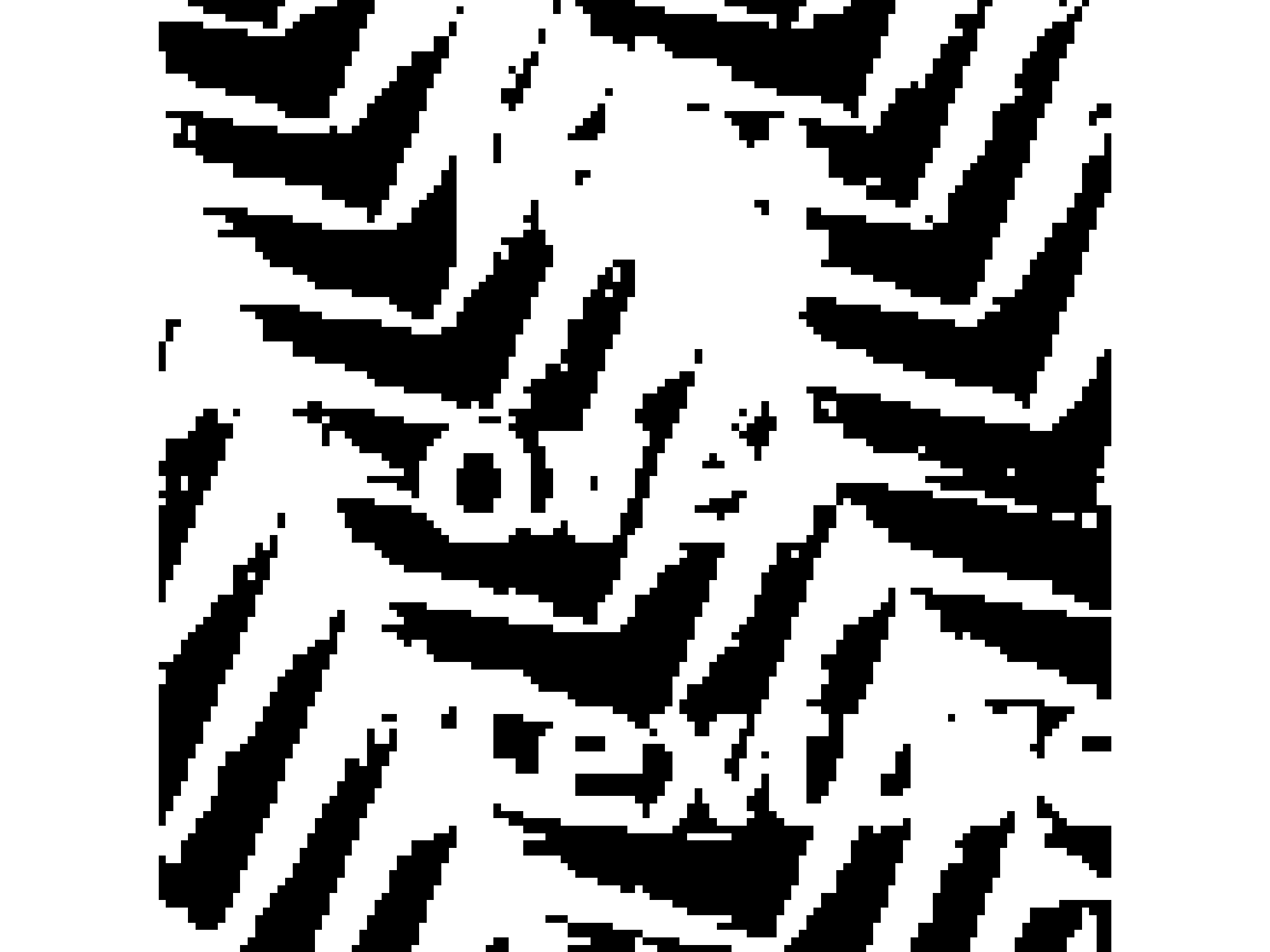}
                \vspace{-0.04cm}
            \hspace{-6cm} 
        \end{subfigure}%
        \begin{subfigure}[b]{0.25\textwidth}
			~ 
                \includegraphics[width=\textwidth]{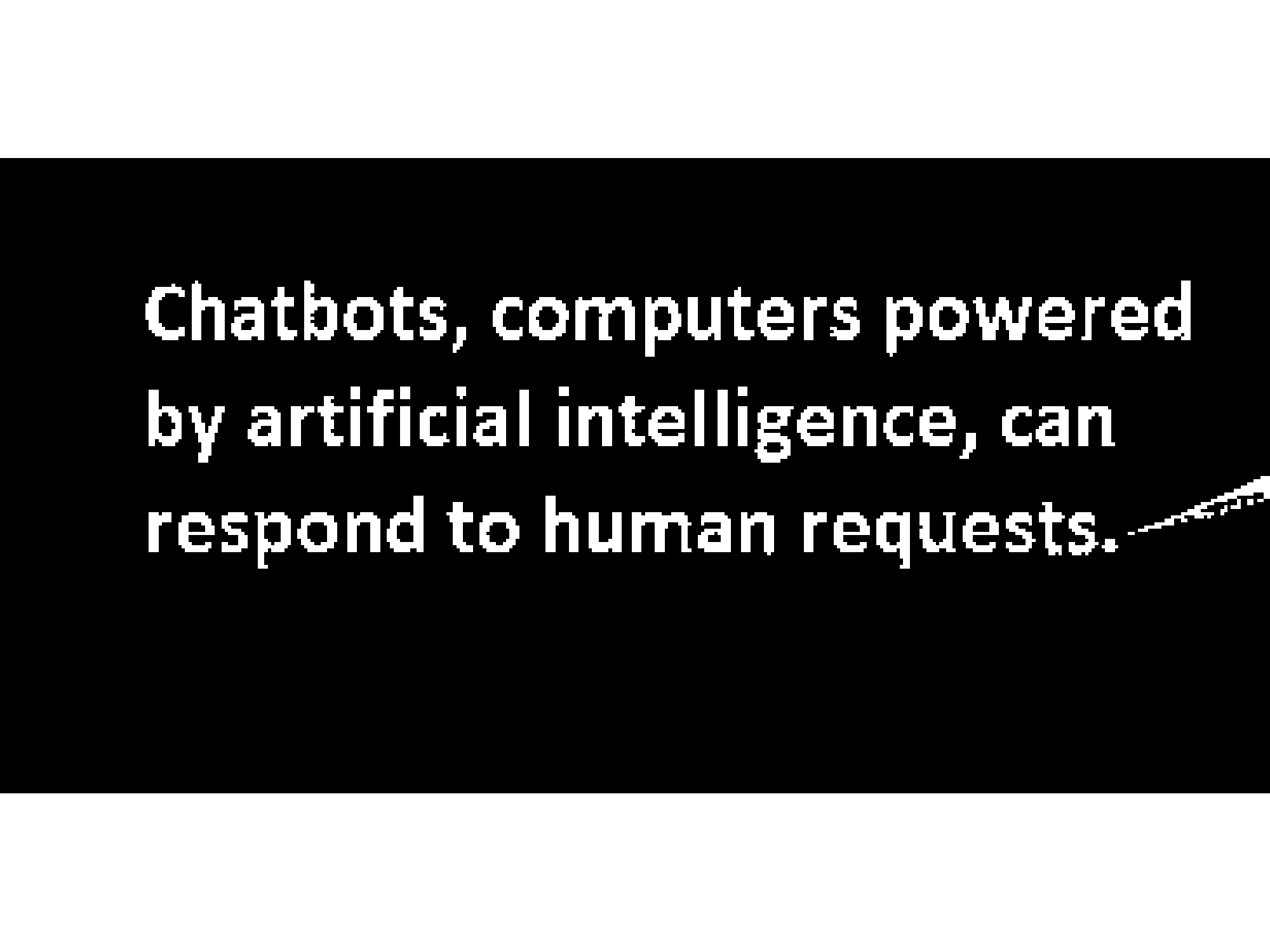}
                \vspace{-0.4cm}
            \hspace{-2cm} 
        \end{subfigure}%
        \begin{subfigure}[b]{0.30\textwidth}
                \includegraphics[width=\textwidth]{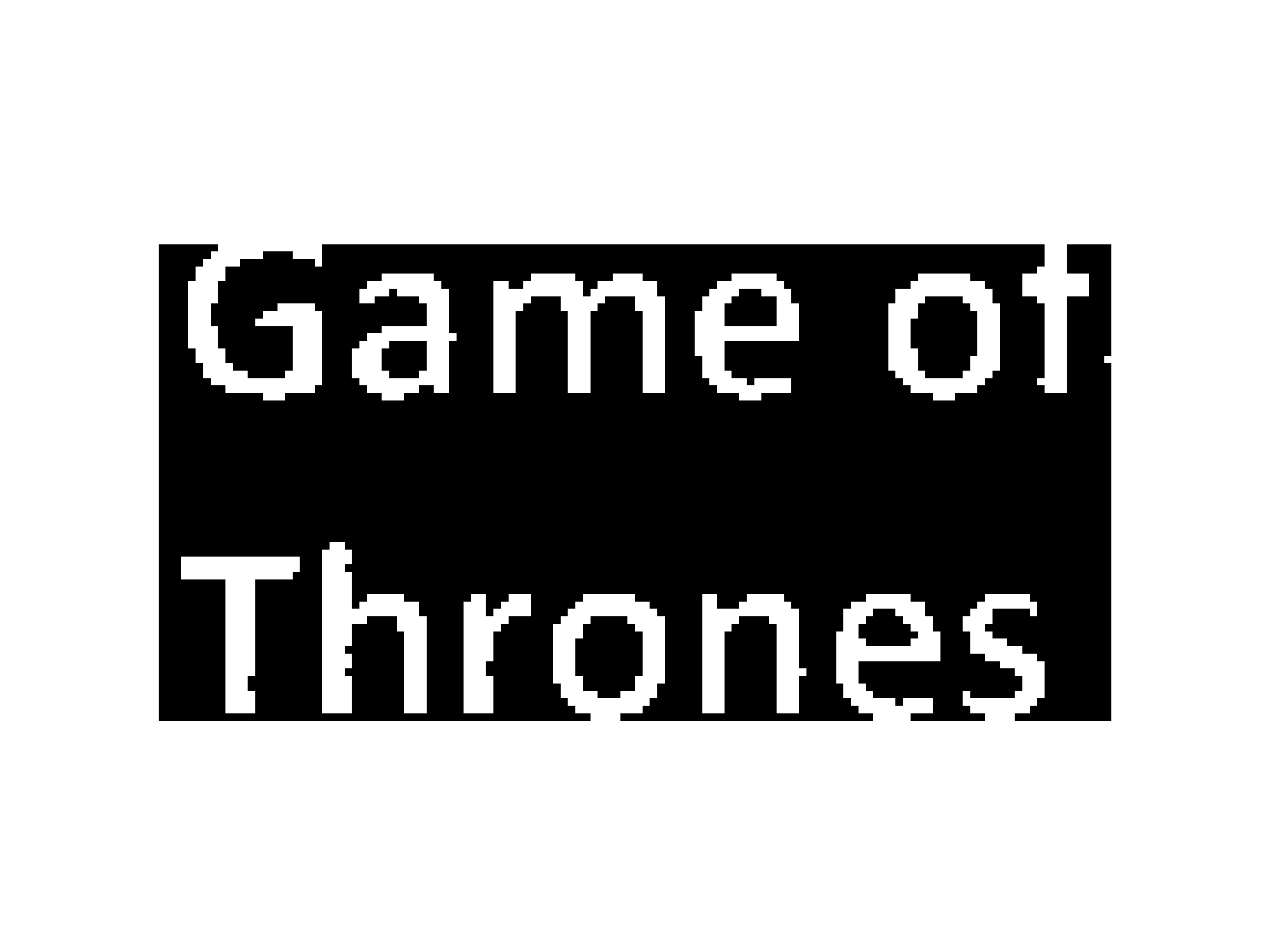}
                 \vspace{-0.61cm}
              \hspace{-4.8cm}
        \end{subfigure}
         \\[1ex]\vspace{-0.9cm}
                 \begin{subfigure}[b]{0.22\textwidth}
                \includegraphics[width=\textwidth]{mask_test-eps-converted-to.pdf}
                                \vspace{-0.3cm}
          \hspace{-1.5cm}    
        \end{subfigure}%
        ~ 
		\vspace{-0.02cm}        
        \begin{subfigure}[b]{0.18\textwidth}
                \includegraphics[width=\textwidth]{mask_texture8-eps-converted-to.pdf}
                \vspace{-0.04cm}
            \hspace{-6cm} 
        \end{subfigure}%
        \begin{subfigure}[b]{0.25\textwidth}
			~ 
                \includegraphics[width=\textwidth]{mask_test2-eps-converted-to.pdf}
                \vspace{-0.4cm}
            \hspace{-2cm} 
        \end{subfigure}%
        \begin{subfigure}[b]{0.30\textwidth}
                \includegraphics[width=\textwidth]{mask_texture10-eps-converted-to.pdf}
                 \vspace{-0.61cm}
              \hspace{-4.8cm}
        \end{subfigure}
        \caption{Segmentation result of the proposed method with different binarization methods. The images in the first row denotes the original images. And the images in the second and third rows show the foreground maps by binarization at the end of each iteration, and at the the very end respectively.}
\end{figure*}

\subsection{Robustness to Initialization}
In this section we present the stability of the algorithm with respect to the initialization of $w$.
One way to evaluate the stability of the optimization algorithm and its convergence, is to evaluate the effect of initialization in the final results.
If the final result does not depend much on the initialized values, it shows the robustness of the algorithm. To make sure the proposed algorithm is robust to the initialization, we provide the segmentation results for a test images, with 5 different initializations in Figure 10. 
The first one is to initialize the $w$ values with all zeros. The second one is to initialize them with the constant value of 0.5. The third one is to initialize them with Gaussian random variable with mean and variance equal to 0.5 and 0.1 respectively (and clipping the values to between 0 and 1).
The fourth one is to initialize them with uniform distribution in [0,1]. 
And the last scheme is to perform least squares fitting using $P_1$ only as the basis, and consider the pixels with large fitting error as foreground.
It is worth mentioning that the number of iterations in our optimization is set to 10, which is not very large to make the effect of initialization disappear.
As we can see the segmentation results with different initialization schemes are roughly similar, showing the robustness of this algorithm.

\begin{figure}
        \centering
        \hspace{-0.02cm}
        \begin{subfigure}[b]{0.22\textwidth}
                \includegraphics[width=\textwidth]{texture8_orig-eps-converted-to.pdf}
            \hspace{-3.5cm} 
        \end{subfigure}%
        ~ 
        \begin{subfigure}[b]{0.22\textwidth}
                \includegraphics[width=\textwidth]{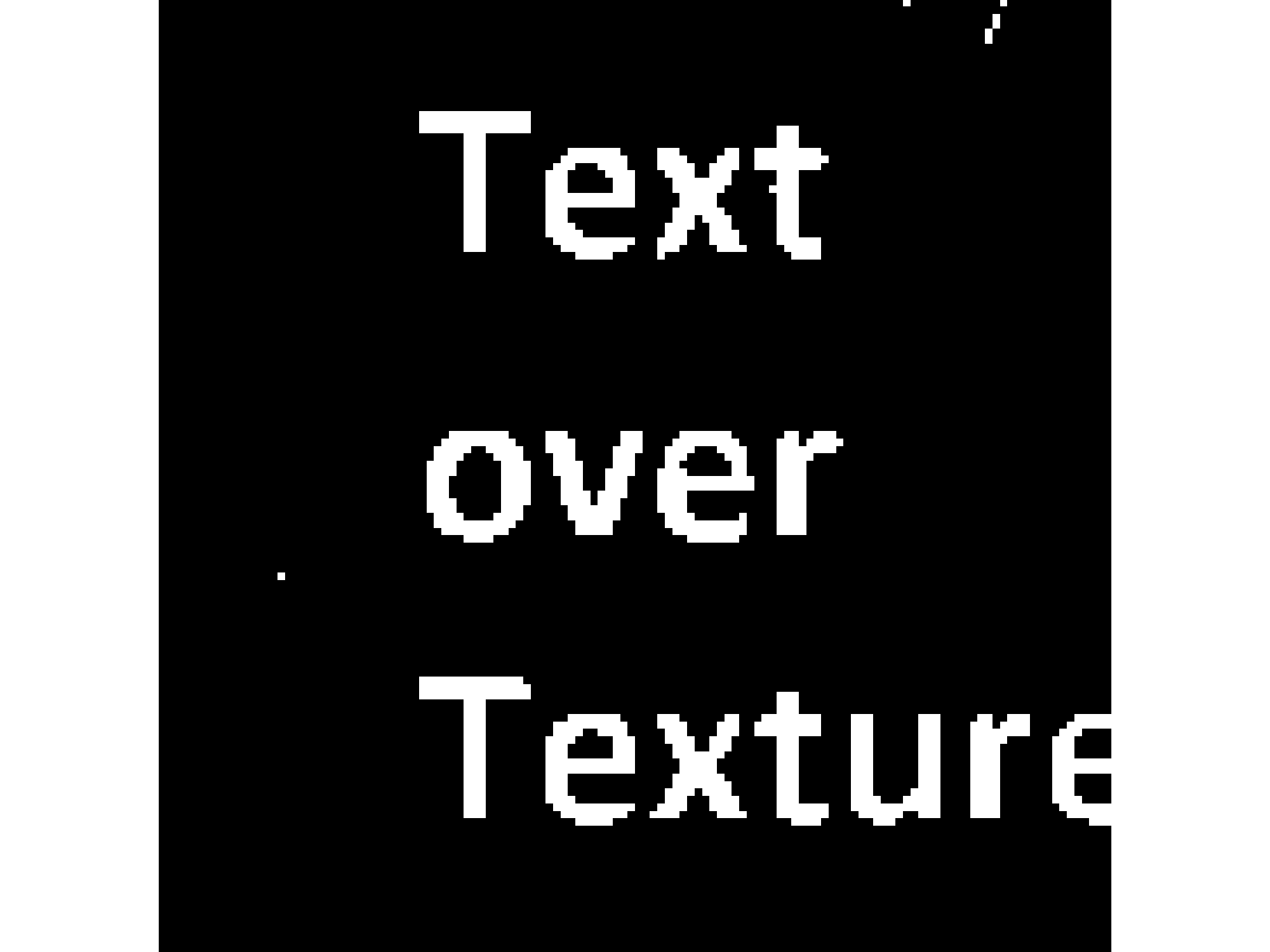}
              \hspace{-4.8cm}
        \end{subfigure}
         \\[1ex]
                 \centering
        \hspace{-1.4cm}

        \begin{subfigure}[b]{0.22\textwidth}
                \includegraphics[width=\textwidth]{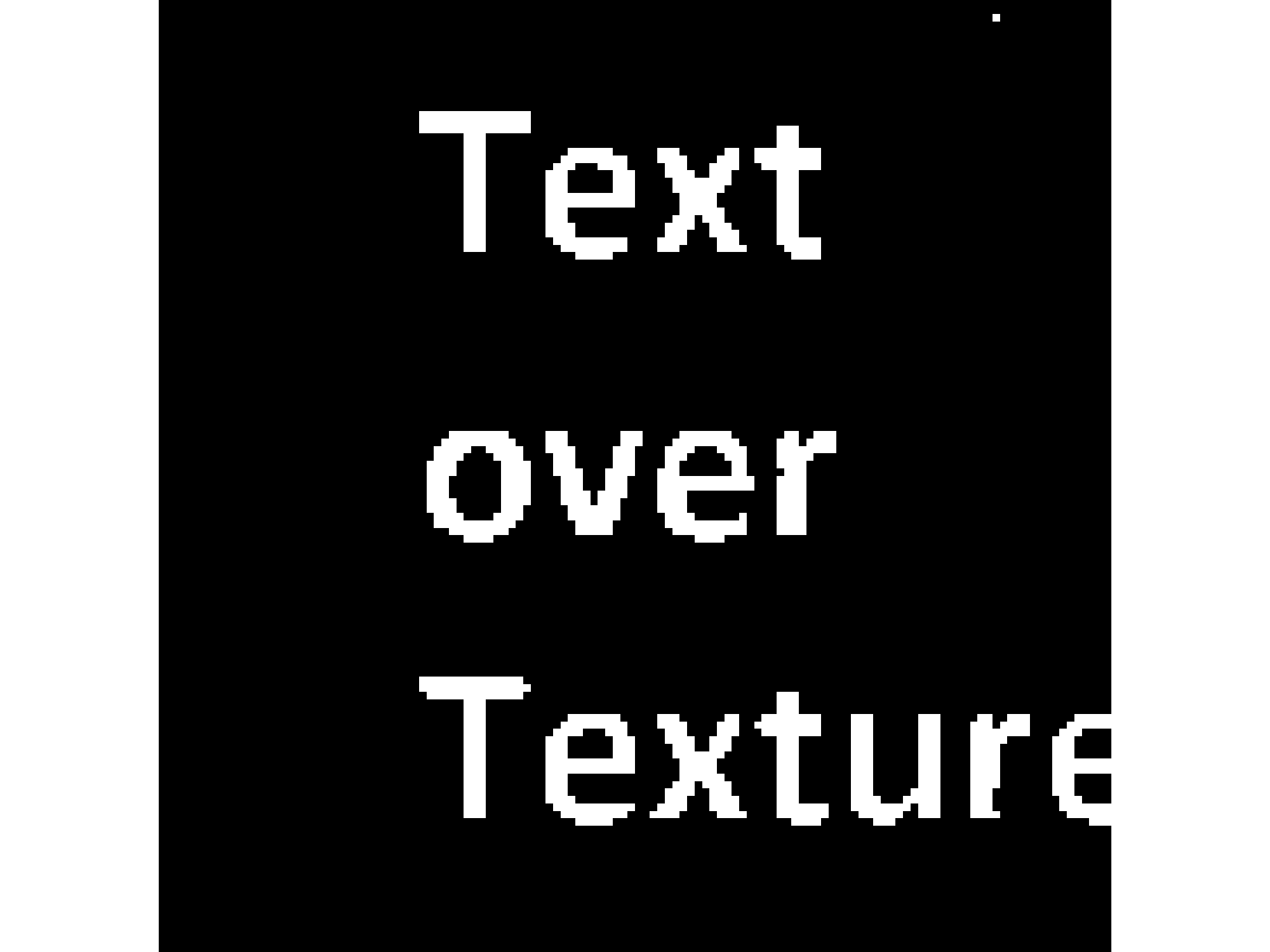}
            \hspace{-3cm} 
        \end{subfigure}%
        ~ 
        \begin{subfigure}[b]{0.22\textwidth}
                \includegraphics[width=\textwidth]{mask_texture8_gauss-eps-converted-to.pdf}
              \hspace{-4.8cm}
        \end{subfigure}
         \\[1ex]
                 \centering
        \hspace{-1.4cm}

        \begin{subfigure}[b]{0.22\textwidth}
                \includegraphics[width=\textwidth]{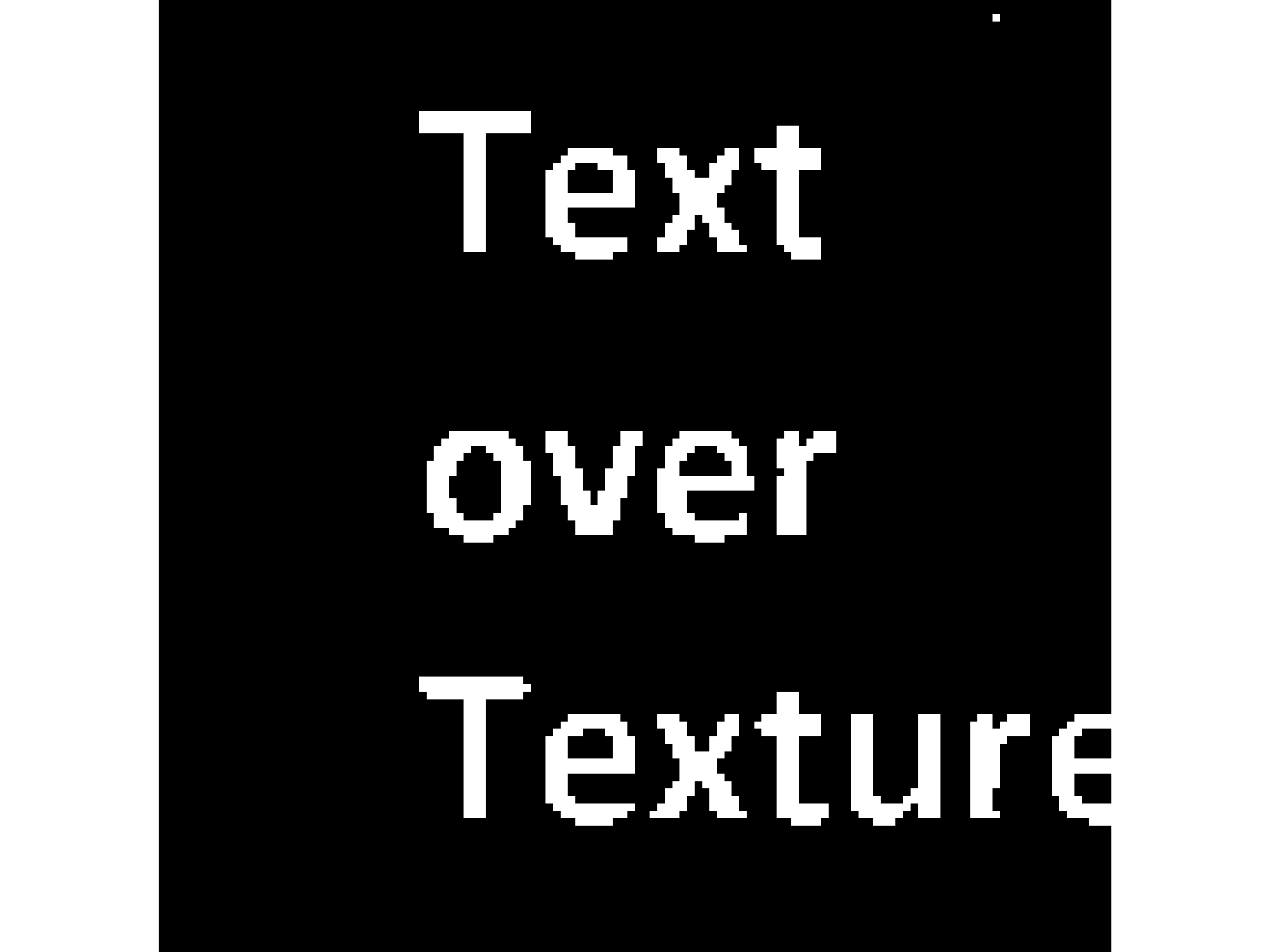}
            \hspace{-3cm} 
        \end{subfigure}%
        ~ 
        \begin{subfigure}[b]{0.22\textwidth}
                \includegraphics[width=\textwidth]{mask_texture8-eps-converted-to.pdf}
              \hspace{-4.8cm}
        \end{subfigure}   \\[1ex]
        \caption{ Segmentation result for different initialization schemes. The second, third, fourth, fifth and sixth images denote the segmentation results by all-zeros initialization, constant value of 0.5, zero-mean unit variance Gaussian, uniform distribution in [0,1], and error based initialization respectively.}
\end{figure}

\subsection{Convergence Analysis}
The optimization problem in Eq. (7) is a mixed integer programming problem, and is very difficult to solve directly.
In this work, we solve a relaxed constrained optimization problem defined in (12), and then binarize the resulting mask image. 
The relaxed problem is still a bi-convex problem as it involves bi-linear terms of the unknown variables (product of $w$ and $\alpha$).
We solve this problem iteratively using the ADMM method. 
Theoretical convergence analysis for this algorithm is very challenging and beyond the scope of this paper.  
Instead we provide experimental convergence analysis by looking at the reduction in the loss (Eq. (12)) at successive iterations. 
Specifically we look at the absolute relative loss reduction, calculated as $\frac{| L^{(k+1)}-L^{(k)} |}{L^{(k)}}$ over different iterations, where $L^{(k)}$ denotes the loss function value at $k$-th iteration.
The experimental convergence analysis for 4 sample images are shown in Figure 11.
As we can see from this figure, the loss reduction keeps decreasing until it converges to zero typically under 10 iterations. This is why we set the maximum iteration number to 10 for the experimental results shown earlier.
In terms of computational time, it takes around 2 seconds to solve this optimization for an image block of size 64x64, using MATLAB 2015 on a Laptop with core i-5 CPU running at 2.2 GHz.
This can be order of magnitudes faster by running it on a more powerful machine and possibly on GPU.
\begin{figure}[h]
\begin{center}
    \includegraphics [scale=0.32] {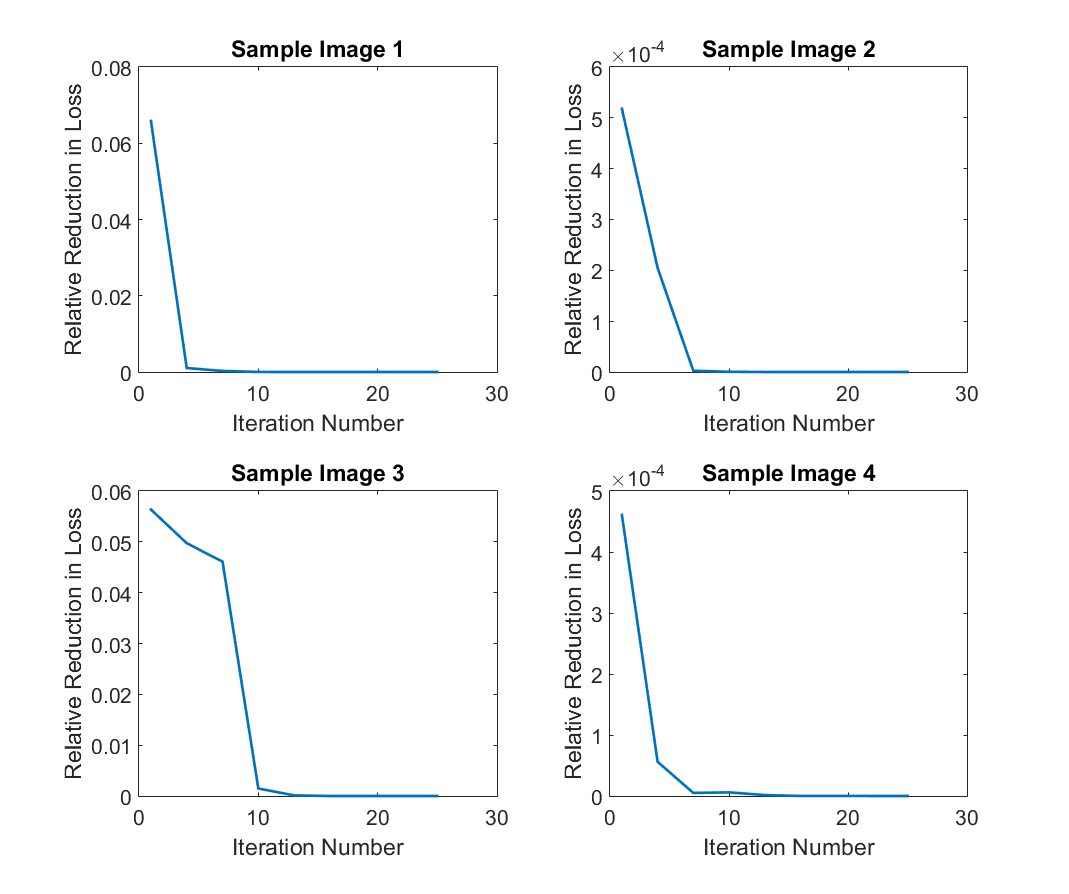}
  \caption{The relative loss reduction for four images.}
\end{center}
\end{figure}

\subsection{Choice of Subspaces}
As we can see from the optimization problem in Eq. (12), we assume that the subspaces/dictionaries are known beforehand.
It is obvious that the choice of $P_1$ and $P_2$ significantly affects the overall performance of the proposed signal decomposition framework.
Choice of $P_1$ and $P_2$ largely depends on the applications.
One could choose these subspaces by the prior knowledge in the underlying applications. For example as shown in the experimental result section, for separation of smooth background from foreground text and graphics, DCT \cite{DCT} and Hadamard subspaces \cite{had_tran} are suitable for background and foreground components respectively.
Figure 12 shows a comparison between segmentation results using low-frequency DCT subspace for both background and foreground, and DCT subspace for background and Hadamard subspace for foreground.
As we can see using Hadamard bases for foreground yields better results.

\begin{figure}[h]
\begin{center}
\hspace{-0.1cm}
    \includegraphics [scale=0.14] {texture8_orig-eps-converted-to.pdf}
  \hspace{-0.1cm} 
  \includegraphics [scale=0.14] {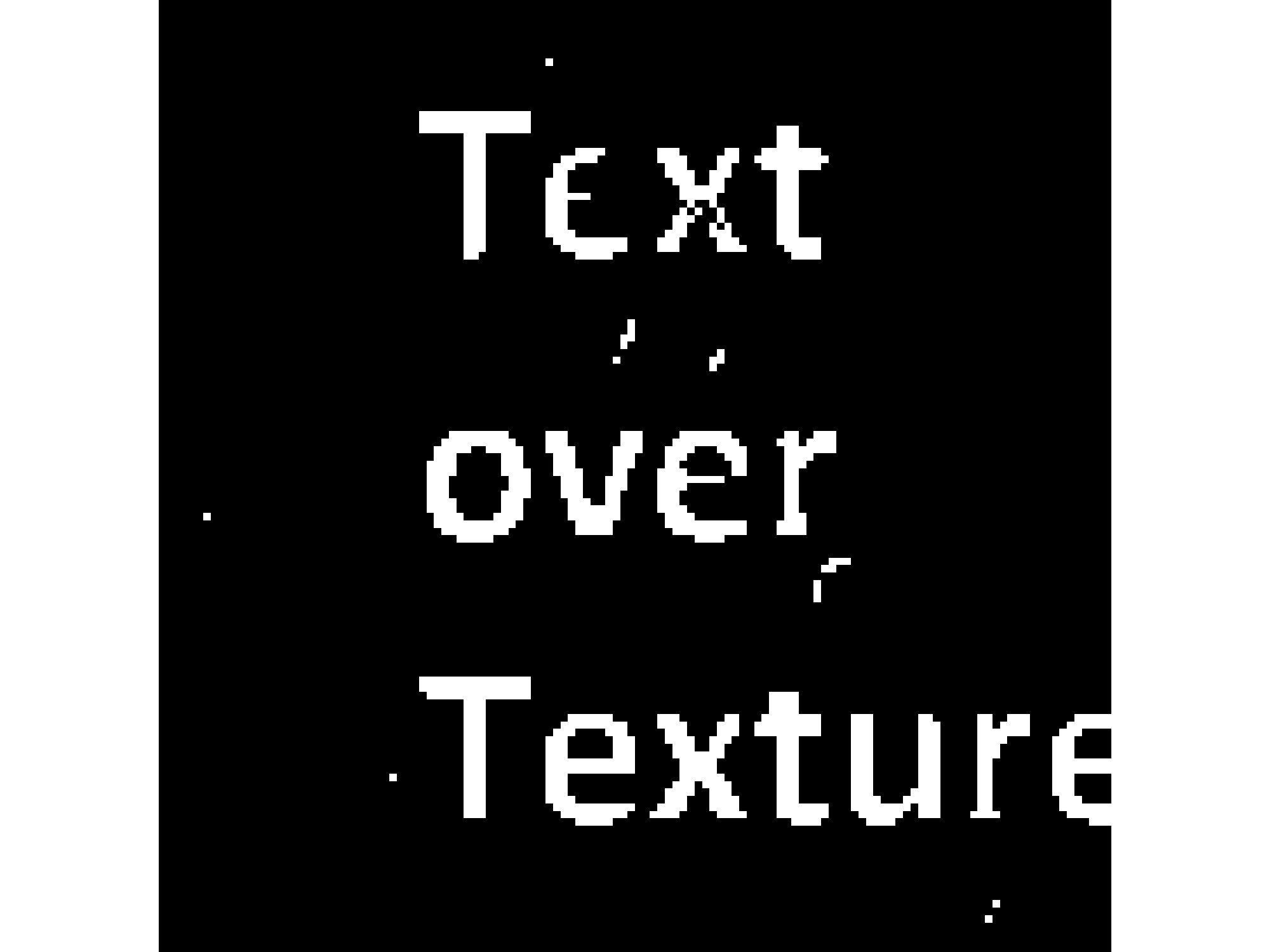}
    \hspace{-0.1cm}
  \includegraphics [scale=0.14] {mask_texture8-eps-converted-to.pdf}
  \caption{The left, middle and right images denote the original image, the foreground maps by using DCT bases for both background and foreground, and using DCT bases for background and Hadamard for foreground respectively.}
\end{center}
\end{figure}

\subsection{Comparison With a Text Detection Algorithm}
To evaluate the performance of one of the promising text detection algorithms on images used in our work, as well as the ones from ICDAR and VOC, we applied the Connectionist Text Proposal Network proposed in \cite{Connectionist} on our test images. 
We used the pretrained model and implementation available in \cite{Connectionist_url}, to derive the text boxes.
Fig 13 provides a comparison of the results by the Connectionist algorithm and our proposed model on some of our test images and also ICDAR 2017 text detection benchmark.
As we can see the Connectionist algorithm's results are descent, but it misses some parts of the texts in most of the following the images.
Furthermore, this algorithm can only provide a bounding box, not the segmented text pixels.
\begin{figure}
        \centering
        \hspace{-0.12cm}
        \begin{subfigure}[b]{0.22\textwidth}
                \includegraphics[width=\textwidth]{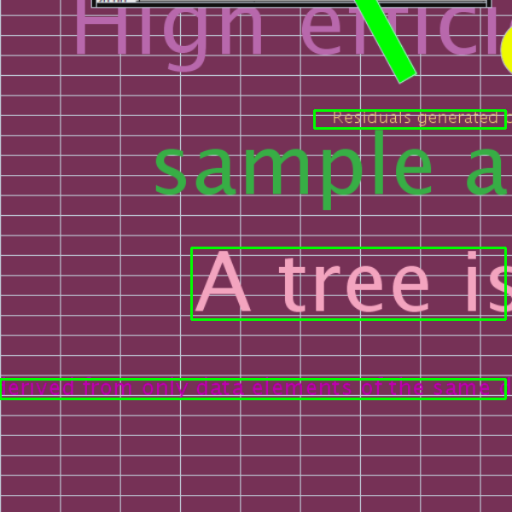}
            \hspace{-3.5cm} 
        \end{subfigure}%
        ~ 
        \begin{subfigure}[b]{0.22\textwidth}
                \includegraphics[width=\textwidth]{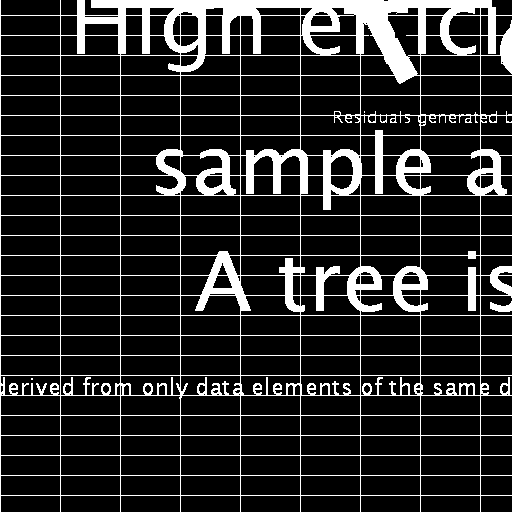}
              \hspace{-4.8cm}
        \end{subfigure}
         \\[1ex]
                 \centering
        \hspace{-1.4cm}

        \begin{subfigure}[b]{0.22\textwidth}
                \includegraphics[width=\textwidth]{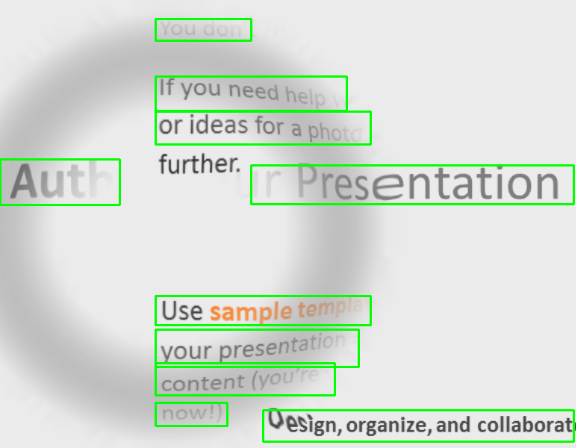}
            \hspace{-3cm} 
        \end{subfigure}%
        ~ 
        \begin{subfigure}[b]{0.22\textwidth}
                \includegraphics[width=\textwidth]{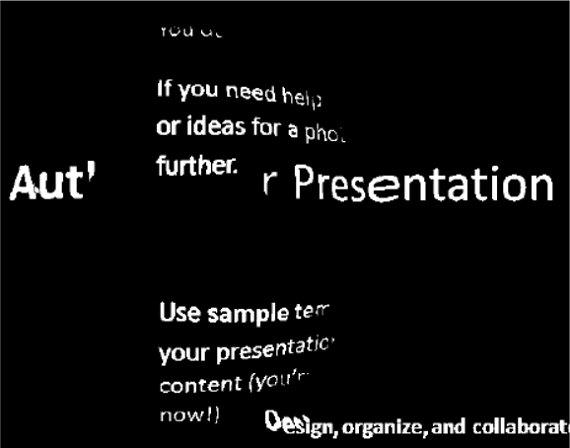}
              \hspace{-4.8cm}
        \end{subfigure}
         \\[1ex]
                 \centering
        \hspace{-1.4cm}

        \begin{subfigure}[b]{0.22\textwidth}
                \includegraphics[width=\textwidth]{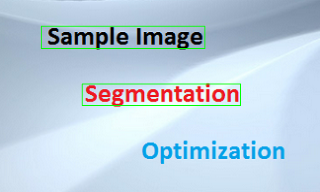}
            \hspace{-3cm} 
        \end{subfigure}%
        ~ 
        \begin{subfigure}[b]{0.22\textwidth}
                \includegraphics[width=\textwidth]{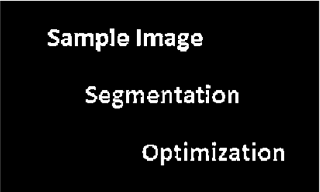}
              \hspace{-4.8cm}
        \end{subfigure}
        \\[1ex]
                 \centering
        \hspace{-1.4cm}

        \begin{subfigure}[b]{0.22\textwidth}
                \includegraphics[width=\textwidth]{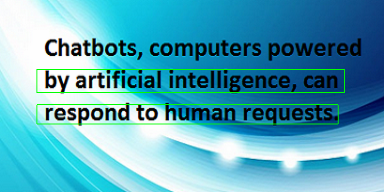}
            \hspace{-3cm} 
        \end{subfigure}%
        ~ 
        \begin{subfigure}[b]{0.22\textwidth}
                \includegraphics[width=\textwidth]{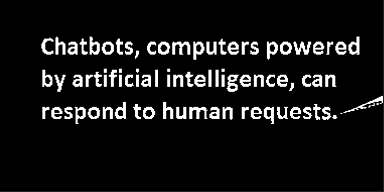}
              \hspace{-4.8cm}
        \end{subfigure} \\
        [1ex]
                 \centering
        \hspace{-1.4cm}

        \begin{subfigure}[b]{0.22\textwidth}
                \includegraphics[width=\textwidth]{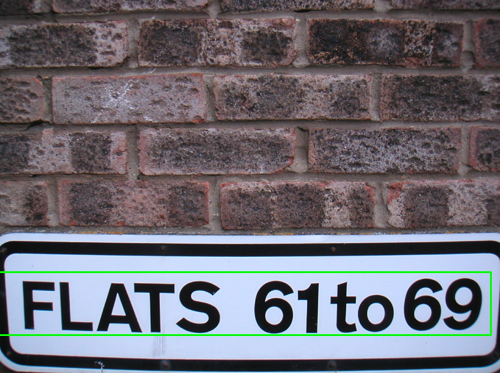}
            \hspace{-3cm} 
        \end{subfigure}%
        ~ 
        \begin{subfigure}[b]{0.22\textwidth}
                \includegraphics[width=\textwidth]{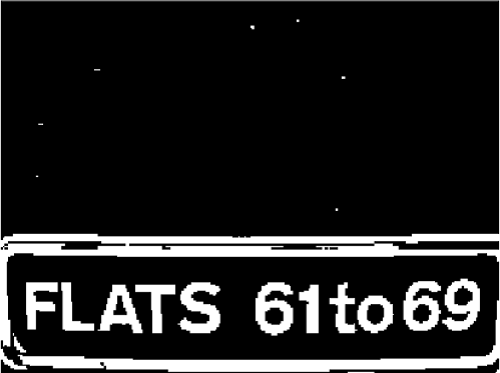}
              \hspace{-4.8cm}
        \end{subfigure} \\
        [1ex]
                 \centering
        \hspace{-1.4cm}

        \begin{subfigure}[b]{0.22\textwidth}
                \includegraphics[width=\textwidth]{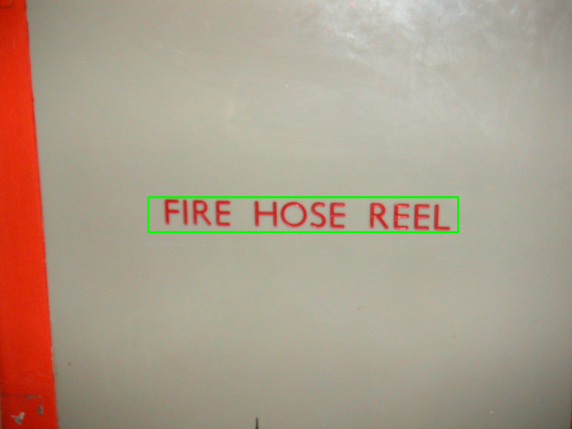}
            \hspace{-3cm} 
        \end{subfigure}%
        ~ 
        \begin{subfigure}[b]{0.22\textwidth}
                \includegraphics[width=\textwidth]{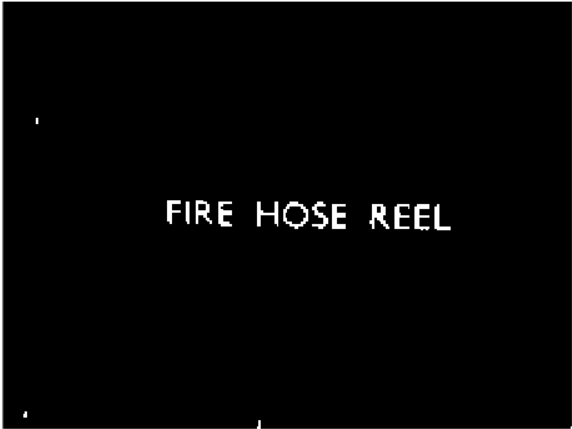}
              \hspace{-4.8cm}
        \end{subfigure}
        \caption{Comparison of the detected text region bounding boxes by the connectionist algorithm (on the left), and the segmented foreground mask by our algorithm (on the right).}
\end{figure}

\subsection{The Segmentation Results of The Proposed Model on a Few Images From VOC Segmentation Dataset}
We also explored the application of our method for the VOC dataset \cite{voc_dataset}. Many images in this dataset do not contain obvious foreground and hence is not suitable for testing the performance of our algorithm.  For some images with obvious foreground, the foreground and background do not have distinguished subspace representation, and hence also cannot be separated effectively by the proposed method.  
We selected some images where the background is smooth and the foreground follows a distinct pattern.
Fig. 14 presents the segmentation results of our algorithm on 4 images.
\begin{figure}[h]
\begin{center}
    \includegraphics [scale=0.28] {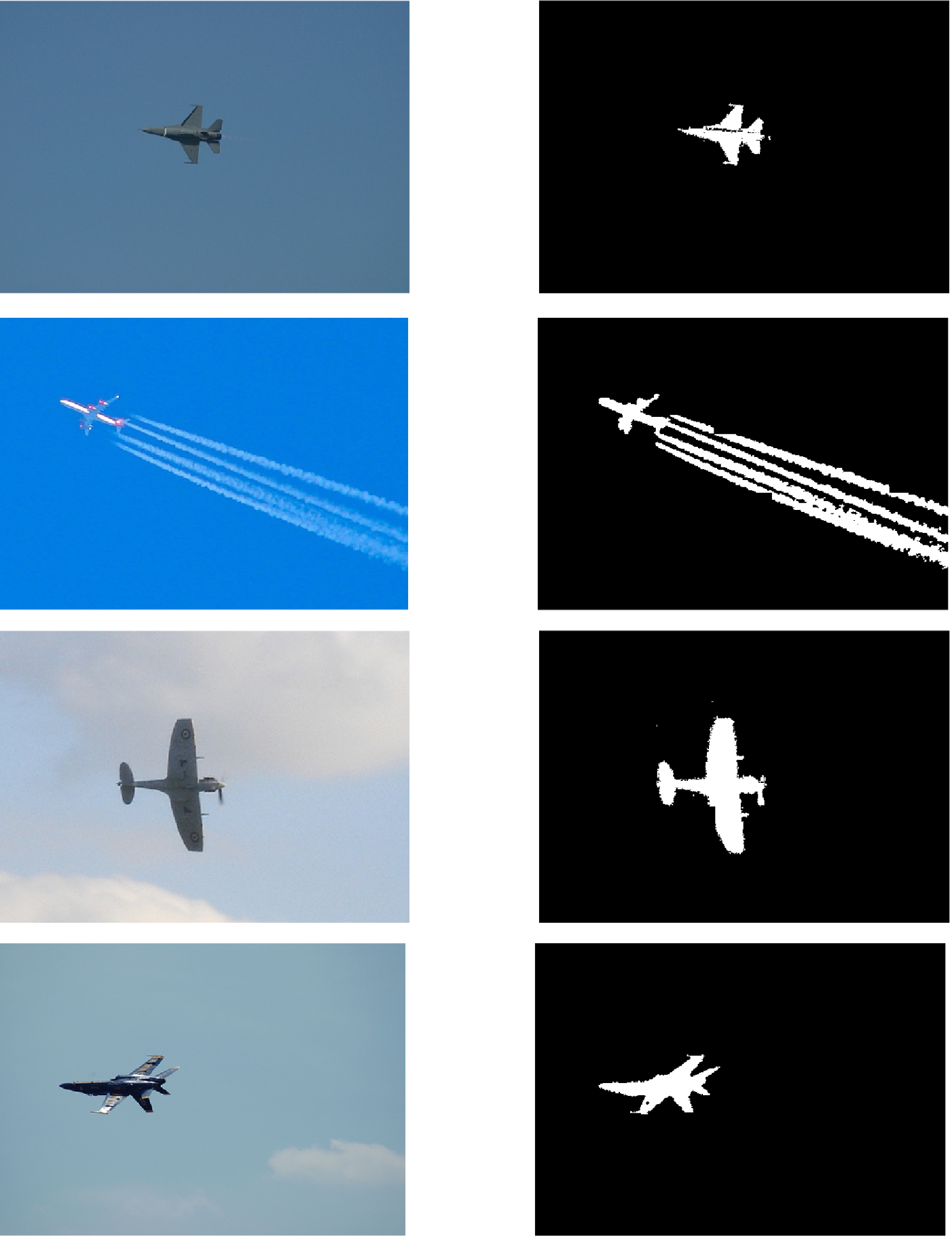}
  \caption{The segmentation results of the proposed algorithm on 4 samples images from VOC benchmark}
\end{center}
\end{figure}

\section{conclusion}
This paper looks at the signal decomposition problem under overlaid addition, where the signal values at each point comes from one and only one of the components (in contrast with the traditional signal decomposition case, which assumes a given signal is the sum of all signal components).
This problem is formulated in an optimization framework, and an algorithm based on the augmented Lagrangian method is proposed to solve it.
Suitable regularization terms are added to the cost function to promote desired structure of each component.
We evaluate the performance of this scheme for different applications, including 1D signal decomposition, text extraction from images, and moving object detection in video.
We also provide a comparison of this algorithm with some of the previous signal decomposition techniques on image segmentation task.
As the future work, we want to use subspace/dictionary learning algorithms to learn the subspaces for different components. 
There are many algorithms available for subspace/dictionary learning \cite{dic1}-\cite{dic5}.
In the case where it is possible to access training data that only consist of individual components, one could use transform learning methods (such as the KLT \cite{KLT} or K-SVD algorithm \cite{dic3}) on a large training set, to derive $P_1$ and $P_2$ separately.
The more challenging problem is when we do not have access to the training data of one component only.
In that case, one could use a training set of super-imposed signals, and use an optimization framework that simultaneously performs masked decomposition and subspace learning.

\section*{Acknowledgment}
The authors would like to thank Ivan Selesnick, Arian Maleki, and Pablo Sprechmann for their valuable comments and feedback regarding this work. We also thank Michael Black's group and Stephen Becker for providing us the source code for optical flow extraction, and Fast-RPCA respectively.

\ifCLASSOPTIONcaptionsoff
  \newpage
\fi


\begin{thebibliography}{1}
\begin{small}
\bibitem{fourier}
HJ Nussbaumer, ``Fast Fourier transform and convolution algorithms'',  Springer Science and Business Media, Vol. 2, 2012.
\bibitem{wave1}
SG Mallat, ``A theory for multiresolution signal decomposition: the wavelet representation'', IEEE transactions on pattern analysis and machine intelligence 11.7: 674-693, 1989.
\bibitem{wave2}
I Daubechies, ``Ten lectures on wavelets'', Society for industrial and applied mathematics, 1992.
\bibitem{bofili}
P Bofill, M Zibulevsky, ``Underdetermined blind source separation using sparse representations'', Signal processing, 2001.
\bibitem{starck}
JL Starck, M Elad, DL Donoho, ``Image decomposition via the combination of sparse representations and a variational approach'', IEEE transactions on image processing, 2005.
\bibitem{elad}
M Elad, JL Starck, P Querre, DL Donoho, ``Simultaneous cartoon and texture image inpainting using morphological component analysis (MCA)'', Applied and Computational Harmonic Analysis, 340-358, 2005.
\bibitem{fazel}
B Recht, M Fazel, PA Parrilo, ``Guaranteed minimum-rank solutions of linear matrix equations via nuclear norm minimization'', SIAM review, 471-501, 2010.
\bibitem{candes}
EJ Candes, X Li, Y Ma, J Wright, ``Robust principal component analysis?'', Journal of the ACM, 58(3), 2011.
\bibitem{peng}
Y Peng, A Ganesh, J Wright, W Xu, Y Ma, ``RASL: Robust alignment by sparse and low-rank decomposition for linearly correlated images'', IEEE Transactions on Pattern Analysis and Machine Intelligence, 2233-2246, 2012.
\bibitem{tilt}
Z Zhang, A Ganesh, X Liang, Y Ma, ``Tilt: Transform invariant low-rank textures'', IJCV, 2012.
\bibitem{back_sub}
T Bouwmans, S Javed, H Zhang, Z Lin, R Otazo, ``On the applications of robust PCA in image and video processing'', Proceedings of the IEEE, 106(8), pp.1427-1457, 2018
\bibitem{hadamard}
RA Horn, ``The hadamard product'', Proceedings of Symposia in Applied Mathematics, Vol. 40, 1990.
\bibitem{cycling}
CH Yeh, W Shi , ``Identifying Phase-Amplitude Coupling in Cyclic Alternating Pattern using Masking Signals'', Scientific reports, 8(1), 2649, 2018.
\bibitem{clus}
T Kanungo, DM Mount, NS Netanyahu, C, Piatko, R Silverman, A Wu, ``An efficient k-means clustering algorithm: Analysis and implementation'', IEEE transactions on pattern analysis and machine intelligence, 881-892, 2002.
\bibitem{clus2}
P Haffner,  P.G.  Howard,  P.  Simard,  Y.  Bengio  and  Y. Lecun, ``High quality document image compression with DjVu'', Journal of Electronic Imaging, 7(3), 410-425, 1998.
\bibitem{MRC}
R.L. DeQueiroz, R.R. Buckley and M. Xu, ``Mixed raster content (MRC) model for compound image compression'', Electronic Imaging, International Society for Optics and Photonics, 1998.
\bibitem{spec}
T. Lin and P. Hao, ``Compound image compression for real-time computer screen image transmission'', IEEE Transactions on Image Processing, 14(8), 993-1005, 2005.
\bibitem{mytv}
S Minaee, Y Wang, ``Screen content image segmentation using robust regression and sparse decomposition'', IEEE Journal on Emerging and Selected Topics in Circuits and Systems, 2016.
\bibitem{lad}
S Minaee and Y Wang, ``Screen Content Image Segmentation Using Least Absolute Deviation Fitting'', IEEE International Conference on Image Processing, 2015. 
\bibitem{sp_text}
TV Hoang, S Tabbone, ``Text extraction from graphical document images using sparse representation'', International Workshop on Document Analysis Systems, ACM, 2010.
\bibitem{TV}
S  Osher,  M  Burger,  D  Goldfarb,  J  Xu  and  W  Yin,  ``An  iterative  regularization method for total variation-based image restoration'', Multiscale Modeling and Simulation, 460-489, 2005.
\bibitem{lag1}
S. Boyd, N. Parikh, E. Chu, B. Peleato and J. Eckstein, ``Distributed optimization and statistical learning via the alternating direction method of multipliers'', Foundations and Trends in Machine Learning, 1-122, 2011.
\bibitem{lag2}
DP Bertsekas, ``Nonlinear programming'', Belmont: Athena scientific, 1999.
\bibitem{bach1}
F Bach, R Jenatton, J Mairal, G. Obozinski, ``Convex optimization with sparsity-inducing norms'', Optimization for Machine Learning, 2011.
\bibitem{patrick}
PL Combettes and VR Wajs, ``Signal recovery by proximal forward-backward splitting,'' Multiscale Modeling and Simulation, pp. 1168-1200, 2005.
\bibitem{ransac2}
M Zuliani, CS Kenney,  BS Manjunath, ``The multiransac algorithm and its application to detect planar homographies'', IEEE International Conference on Image Processing, 2005.
\bibitem{soft}
D. Donoho, ``De-noising by soft-thresholdingm'' IEEE Transactions on Information Theory, 613-627, 1995.



\bibitem{becker1}
A Aravkin, S Becker, V Cevher, P Olsen, ``A variational approach to stable principal component pursuit'', Conference on Uncertainty in Artificial Intelligence, 2014.
\bibitem{becker2}
https://github.com/stephenbeckr/fastRPCA


\bibitem{SCC_data}
ISO/IEC JTC 1/SC 29/WG 11 Requirements subgroup, ``Requirements for an extension of HEVC for coding of screen content,'' in MPEG 109 meeting, 2014.
\bibitem{SCC_tran}
T. Zhang, B. Ghanem, S. Liu, C. Xum and B. Yin, ``Screen content coding based on HEVC framework,'' IEEE Transactions on Multimedia, 1316-1326, 2014.

\bibitem{metrics}
DM Powers, ``Evaluation: from precision, recall and F-measure to ROC, informedness, markedness and correlation'', International Journal of Machine Learning Technology, 37-63, 2011.
\bibitem{of1}
D Sun, S Roth, MJ Black , ``Secrets of optical flow estimation and their principles'', IEEE Conference on Computer Vision and Pattern Recognition, 2010.
\bibitem{of2}
http://cs.brown.edu/~black/code.html

\bibitem{dic1}
X Shu, F Porikli, N Ahuja, ``Robust orthonormal subspace learning: Efficient recovery of corrupted low-rank matrices'', CVPR, IEEE, 2014.
\bibitem{dic2}
F De La Torre, MJ Black, ``A framework for robust subspace learning'', International Journal of Computer Vision, 2003.
\bibitem{dic3}
M Aharon, M Elad, A Bruckstein, ``K-SVD: An Algorithm for Designing Overcomplete Dictionaries for Sparse Representation'', IEEE Transactions on signal processing, 4311-4322, 2006.
\bibitem{dic4}
Q Pan, D Kong, CHQ Ding, B Luo, ``Robust Non-Negative Dictionary Learning'', In AAAI, pp, 2027-2033, 2014.
\bibitem{dic5}
Z Jiang, Z Lin, LS Davis, ``Label consistent K-SVD: Learning a discriminative dictionary for recognition'', IEEE Transactions on Pattern Analysis and Machine Intelligence, 2651-2664, 2013.
\bibitem{KLT}
Levey, A., and M. Lindenbaum, ``Sequential Karhunen-Loeve basis extraction and its application to images,'' IEEE Transactions on Image Processing, 1371-1374, 2000.
\bibitem{DCT}
A.B. Watson, ``Image compression using the discrete cosine transform'', Mathematica journal, 1994.
\bibitem{had_tran}
WK Pratt, J Kane, HC Andrews, ``Hadamard transform image coding'', Proceedings of the IEEE, 58-68, 1969.
\bibitem{Connectionist}
Z Tian, W Huang, T He, P He, Y Qiao, ``Detecting text in natural image with connectionist text proposal network'', In European conference on computer vision, Springer, 2016.
\bibitem{Connectionist_url}
https://github.com/eragonruan/text-detection-ctpn
\bibitem{voc_dataset}
http://host.robots.ox.ac.uk/pascal/VOC/







\end{small}
\end{thebibliography}
\end{document}